\documentclass[preprint,12pt]{elsarticle}
\usepackage{graphicx}
\usepackage{exscale}
\usepackage{times}
\usepackage{color}
\usepackage{amssymb}
\usepackage{setspace}
\usepackage{amsmath}

\font\pppppcarac=ptmr8y at 5pt
\font\ppppcarac=ptmr8y at 6pt
\font\pppcarac=ptmr8y at 7pt
\font\ppcarac=ptmr8y at 8pt

\font\Pcarac=ptmr8y at 9pt

\font\PPcarac=ptmr8y at 12pt
\font\Ppcarac=ptmr8y at 10pt
\font\bf=ptmb8y at 10pt

\newcommand{\bfB}{{\textbf{B}}}
\newcommand{\bfC}{{\textbf{C}}}
\newcommand{\bfD}{{\textbf{D}}}

\newcommand{\bfH}{{\textbf{H}}}

\newcommand{\bfL}{{\textbf{L}}}

\newcommand{\bfQ}{{\textbf{Q}}}
\newcommand{\bfR}{{\textbf{R}}}
\newcommand{\bfS}{{\textbf{S}}}

\newcommand{\bfU}{{\textbf{U}}}
\newcommand{\bfV}{{\textbf{V}}}
\newcommand{\bfW}{{\textbf{W}}}
\newcommand{\bfX}{{\textbf{X}}}
\newcommand{\bfY}{{\textbf{Y}}}
\newcommand{\bfZ}{{\textbf{Z}}}

\newcommand{\bfun}{{\boldsymbol{1}}}
\newcommand{\bfzero}{{ \hbox{\bf 0} }}

\newcommand{\bfa}{{\textbf{a}}}
\newcommand{\bfb}{{\textbf{b}}}

\newcommand{\bfe}{{\textbf{e}}}
\newcommand{\bff}{{\textbf{f}}}
\newcommand{\bfg}{{\textbf{g}}}
\newcommand{\bfh}{{\textbf{h}}}

\newcommand{\bfp}{{\textbf{p}}}
\newcommand{\bfq}{{\textbf{q}}}

\newcommand{\bfs}{{\textbf{s}}}

\newcommand{\bfu}{{\textbf{u}}}
\newcommand{\bfv}{{\textbf{v}}}
\newcommand{\bfw}{{\textbf{w}}}
\newcommand{\bfx}{{\textbf{x}}}
\newcommand{\bfy}{{\textbf{y}}}

\newcommand{\bfPhi}{{\boldsymbol{\Phi}}}

\newcommand{\bfbeta}{{\boldsymbol{\beta}}}

\newcommand{\bfeta}{{\boldsymbol{\eta}}}

\newcommand{\bfphi}{{\boldsymbol{\phi}}}
\newcommand{\bfsigma}{{\boldsymbol{\sigma}}}

\newcommand{\BB}{{\mathbb{B}}}

\newcommand{\DD}{{\mathbb{D}}}
\newcommand{\FF}{{\mathbb{F}}}

\newcommand{\HH}{{\mathbb{H}}}

\newcommand{\MM}{{\mathbb{M}}}
\newcommand{\NN}{{\mathbb{N}}}
\newcommand{\PP}{{\mathbb{P}}}
\newcommand{\QQ}{{\mathbb{Q}}}
\newcommand{\RR}{{\mathbb{R}}}

\newcommand{\UU}{{\mathbb{U}}}

\newcommand{\WW}{{\mathbb{W}}}
\newcommand{\XX}{{\mathbb{X}}}

\DeclareMathAlphabet{\mathonebb}{U}{bbold}{m}{n}
\def\11{{\ensuremath{\mathonebb{1}}}}

\def\bb{{\ensuremath{\mathonebb{b}}}}

\def\ff{{\ensuremath{\mathonebb{f}}}}

\def\qq{{\ensuremath{\mathonebb{q}}}}
\def\uu{{\ensuremath{\mathonebb{u}}}}

\def\ww{{\ensuremath{\mathonebb{w}}}}
\def\xx{{\ensuremath{\mathonebb{x}}}}

\newcommand{\curB}{{\mathcal{B}}}
\newcommand{\curC}{{\mathcal{C}}}
\newcommand{\curE}{{\mathcal{E}}}

\newcommand{\curL}{{\mathcal{L}}}

\newcommand{\curP}{{\mathcal{P}}}

\newcommand{\curR}{{\mathcal{R}}}
\newcommand{\curS}{{\mathcal{S}}}
\newcommand{\curT}{{\mathcal{T}}}
\newcommand{\curU}{{\mathcal{U}}}
\newcommand{\curV}{{\mathcal{V}}}

\newcommand{\bfcurL}{{\boldsymbol{\mathcal{L}}}}
\newcommand{\bfcurN}{{\boldsymbol{\mathcal{N}}}}

\newcommand{\bfcurW}{{\boldsymbol{\mathcal{W}}}}
\newcommand{\bfcurY}{{\boldsymbol{\mathcal{Y}}}}
\newcommand{\bfcurZ}{{\boldsymbol{\mathcal{Z}}}}

\newcommand{\tr}{\hbox{{\PPcarac tr}}\,}
\newcommand{\err}{\hbox{{\PPcarac err}}}
\newcommand{\bfnabla}{{{\boldsymbol{\nabla}}}}
\newcommand{\OVL}{{\hbox{{\Pcarac OVL}}}}
\newcommand{\conv}{\hbox{{\Ppcarac conv}}}
\newcommand{\std}{{\hbox{{\pppcarac std}}}}

\newcommand{\MC}{{\hbox{{\ppppcarac MC}}}}
\newcommand{\pMC}{{\hbox{{\pppppcarac MC}}}}

\newcommand{\ar}{{\hbox{{\ppcarac ar}}}}
\newcommand{\arp}{{\hbox{{\pppcarac ar}}}}
\newcommand{\arpp}{{\hbox{{\ppppcarac ar}}}}

\newcommand{\exper}{{\hbox{{\pppcarac exp}}}}

\newcommand{\predictor}{{\hbox{{\pppcarac pred}}}}
\newcommand{\linear}{{\hbox{{\pppcarac lin}}}}
\newcommand{\opt}{{\hbox{{\pppcarac opt}}}}
\newcommand{\popt}{{\hbox{{\ppppcarac opt}}}}
\newcommand{\diff}{{\hbox{{\ppppcarac diff}}}}
\newcommand{\pdiff}{{\hbox{{\pppppcarac diff}}}}
\newcommand{\wien}{{\hbox{{\pppcarac wien}}}}
\newcommand{\pwien}{{\hbox{{\ppppcarac wien}}}}
\newcommand{\post}{{\hbox{{\pppcarac post}}}}
\newcommand{\ppost}{{\hbox{{\pppppcarac post}}}}
\newcommand{\pmax}{{\hbox{{\pppcarac max}}}}
\newcommand{\pmin}{{\hbox{{\pppcarac min}}}}

\newcommand{\cond}{{\hbox{{\PPcarac cond}}}}
\newcommand{\st}{{\hbox{{\pppcarac st}}}}

\journal{Template of Elsevier Journal}

\begin{document}

\begin{frontmatter}

\title{Sampling of Bayesian posteriors with a non-Gaussian probabilistic learning on manifolds from a small dataset}


\author[1]{C. Soize \corref{cor1}}
\ead{christian.soize@u-pem.fr}
\author[2]{R. Ghanem}
\ead{ghanem@usc.edu}
\author[1]{C. Desceliers}
\ead{christophe.desceliers@u-pem.fr}

\cortext[cor1]{Corresponding author: C. Soize, christian.soize@u-pem.fr}
\address[1]{Universit\'e Paris-Est Marne-la-Vall\'ee, Laboratoire Mod\'elisation et Simulation Multi-Echelle, MSME UMR 8208, 5 bd Descartes, 77454 Marne-la-Vall\'ee, France}
\address[2]{University of Southern California, 210 KAP Hall, Los Angeles, CA 90089, United States}

\begin{abstract}
This paper tackles the challenge presented by small-data to the
task of Bayesian inference. A novel methodology, based on manifold
learning and manifold sampling, is proposed for
solving this computational statistics problem under the following
assumptions: {\bf 1)} neither the prior model nor the likelihood function are
Gaussian and neither can be approximated by a Gaussian measure; {\bf
  2)} the number of functional input (system parameters) and
functional output (quantity of interest) can be large; {\bf 3)} the
number of available realizations of the prior model is small, leading
to the small-data challenge typically associated with expensive
numerical simulations; the number of experimental realizations is also
small; {\bf 4)} the number of the posterior realizations required for
decision is much larger than the available initial dataset. The method
and its mathematical aspects are detailed. Three applications are
presented for validation: The first two involve mathematical constructions
aimed to develop intuition around the method and to explore its
performance. The third example aims to demonstrate the operational
value of the method using a more complex application related to the
statistical inverse identification of the non-Gaussian matrix-valued
random elasticity field of a damaged  biological tissue (osteoporosis
in a cortical bone) using ultrasonic waves.
\end{abstract}


\end{frontmatter}

\section{Introduction}
\label{Section1}
\subsection{Overview of the Bayesian approach}
\label{Section1.1}
The Bayesian approach is a very powerful statistical tool that provides a rigorous formulation for statistical inverse problems and about which numerous papers and treatises have been published \cite{Spall2005,Kaipio2005,Serfling1980,Lunn2000,Tarantola2005,Congdon2007,Carlin2009,Stuart2010,Robert2013,Dashti2017,Ghanem2017c,Soize2017b}. In general, this approach requires the use of variants of the Markov Chain Monte Carlo (MCMC) methods \cite{Andrieu2008} for generating realizations (samples) of the posterior model given a prior model and data typically derived either from numerical simulations or from experimental measurements. This probabilistic approach is extensively used in many fields of physical and life sciences, computational and engineering sciences, and also in machine learning \cite{Murphy2012,Scott2016,Witten2017} and in algorithms devoted to artificial intelligence \cite{Korb2010,Ghahramani2015}.

In the supervised case, the  most popular Bayesian approach consists in constructing the likelihood function using a Gaussian model. For instance, using the output predictive error, the conditional probability density function (pdf) of the random quantity of interest, $\QQ$, given a value $\ww$ of the random parameter $\WW$, is constructed using the equation $\QQ = \ff(\WW) + \BB$ in which $\BB$ is a Gaussian random vector that accounts for modeling errors introduced during the construction of the mathematical/computational model  of the system (represented by the deterministic mapping $\ff$) and/or the experimental measurements errors. Although generally more efficient than their alternatives, MCMC generators for sampling from the posterior distribution \cite{Givens2013,Soize2017b}, still require a large number of calls to the computational model, which can present insurmountable difficulties for expensive models, specially when dealing with high-dimensional problems (functional inputs/outputs). Generally, this situation requires the introduction of a surrogate model for $\ff$ in order to decrease the numerical cost such as the Gaussian-process surrogate model including Gaussian-process regression and linearization techniques (see for instance \cite{Kennedy2000,Kennedy2001,Santner2003} for calibration of computer models, \cite{Bilionis2017,Parussini2017} for formulations using Gaussian processes, and \cite{Flath2011,Pratola2013,Isaac2015,Spantini2017,Zhou2018} for algorithms adapted to large-scale inverse problems in the  Gaussian likelihood framework).

Nevertheless, the additive Gaussian noise model for the likelihood is not always sufficient and embedded models have to be considered for the likelihood. Consequently, the Bayesian approach becomes much more computationally taxing, in particular for high-dimension where it can become outright prohibitive. This is the case if $\QQ = \ff(\WW)$ is replaced by $\QQ = \ff(\WW,\UU)$ in which $\UU$ is a random vector. For instance, $\UU$ corresponds to the spatial discretization of a non-Gaussian tensor-valued random field appearing as a coefficient in a partial differential operator. In such a case, the conditional probability density function of $\QQ = \ff(\ww,\UU)$, given $\WW=\ww$, involves solving the forward problem for a several realizations of $\UU$. A number of procedures have been proposed in recent to tackle this challenge, ranging from adapted representations \cite{Tipireddy:2014,Tsilifis2018}, to reduced-order models, and surrogate models (see for instance, \cite{Grepl2007,Chaturantabut2010,Carlberg2011,Farhat2015,Ryckelynck2005} for reduced-order models and \cite{Meyer2003,Nouy2017,Capiez2014,Soize2019a} for stochastic reduced-order models). Many methods based on the use of polynomial chaos expansions have also been developed (see for instance, \cite{Ghanem2006,Arnst2010,Dolgov2015,Soize2017b} for the identification of stochastic system parameters and random fields in stochastic boundary value problems, \cite{Marzouk2007,Marzouk2009,Rosic2012,Tsilifis2018} for Bayesian inference in inverse problems, and \cite{Nagel2016,Giraldi2017} for explicit construction of surrogate models).

The Bayesian approach for parameter estimation in the non-Gaussian embedded likelihood case has significantly been developed for low dimension \cite{Palacios2006,Arnst2017} and using filtering techniques and functional approximations \cite{Gordon1993,Pajonk2012,Matthies2016,Afshari2017}. Recently, a
nonparametric Bayesian approach for non-Gaussian cases has been proposed \cite{Perrin2018a} for which the invertible covariance matrix of the Gaussian kernel-density estimation is optimized by taking into account the unknown block dependence structure.
\subsection{Framework of the developments and difficulties involved}
\label{Section1.2}
This paper is devoted to the Bayesian inference for the small-data challenge using probabilistic learning on manifolds. We consider the case  $\QQ = \ff(\WW,\UU)$ in which $\WW$, $\UU$, $\QQ$ are random variables with values in $\RR^{n_w}$, $\RR^{n_u}$, $\RR^{n_q}$, and where $(\ww,\uu)\mapsto \ff(\ww,\uu)$ is a nonlinear mapping. In addition to the mapping $\ff$, only two pieces of information are available. The first one consists of an \textit{initial dataset}
(the training set), $\DD_{N_d}$, made up of $N_d$ independent realizations (samples) $\{(\qq^j,\ww^j),j=1,\ldots, N_d \}$ of random variables $(\QQ,\WW)$. The second piece of information consists of an \textit{experimental dataset}, $\DD^\exper_{n_r}$, used for updating, and consisting of $n_r$ given independent experimental realizations (measures or simulations) $\{(\qq^{\exper,r}, r=1,\ldots,n_r \}$ of $\QQ$. The objective then is to construct, using the Bayesian approach, a set of $\nu_\post$ realizations, $\{\ww^{\post,\ell},\ell=1,\ldots,\nu_\post\}$ of the posterior random variable, denoted by $\WW^\post$.  The following requirements have guided the development of the proposed methodology.
\begin{enumerate}
\item The non-Gaussian case is considered. The conditional probability distribution of $\QQ$ given $\WW=\ww$ is not Gaussian. For instance, mapping $\ff$ is not additive with respect to the Gaussian random vector $\UU$, contrarily to the  case for which the output-predictive-error formulation is used, which consists in adding to $\QQ =\ff(\WW)$  a Gaussian noise $\UU$.
\item The problem is in high dimension, either of $n_q$ and $n_w$ can be large.
\item The number $N_d$ of realizations of the prior model is small, which means that we are in the context of the small-data challenge. This situation can be induced, for instance, by the use of an expensive computer code for generating the set $\DD_{N_d}$ of realizations.
\item The number $n_r$ of experimental realizations is small.
\item The number $\nu_\post$ of the posterior realizations required for decision is large.
\end{enumerate}
\subsection{Outline of the proposed method}
\label{Section1.3}
In order to improve numerical conditioning, the initial dataset
$\DD_{N_d}$ is scaled using an adapted affine transformation into a
dataset $D_{N_d}$ made up of the $N_d$ independent realizations
$\{(\bfq^j,\bfw^j),j=1,\ldots, N_d \}$ of the scaled random variables
$(\bfQ,\bfW)$ with values in $\RR^{n_q}\times \RR^{n_w}$. Using this
same affine transformation, experimental dataset $\DD^\exper_{n_r}$ is
transformed into a scaled experimental dataset $D^\exper_{n_r}$ made
up of the $n_r$ independent realizations
$\{\bfq^{\exper,r},r=1,\ldots, n_r \}$.\\

Each of the requirements listed in Section~\ref{Section1.2} presents
its own significant challenges which are addressed throughout the paper.\\

\noindent (i)- For addressing the small-data challenge, the
probabilistic learning on manifolds (PLoM), which has been introduced
in \cite{Soize2016}, is used. This PLoM allows for generating a
\textit{learned dataset} (big dataset) $D_{\nu_\ar}$ of $\nu_\ar$
additional realizations of the prior model of the scaled random vector
$(\bfQ,\bfW)$ in which the number $\nu_\ar$ can be arbitrarily large
($\nu_\ar \gg N_d$), using only information defined by the scaled
initial dataset $D_{N_d}$. The convergence of the learning with
respect to $N_d$ is investigated. This learned dataset $D_{\nu_\ar}$
allows for constructing an accurate estimate of the posterior
distribution.\\

\noindent (ii)- For addressing the high-dimension data challenge, two
reduced-order representations are separately constructed, one for
random vector $\bfQ$ and another one for random vector $\bfW$, using
for each one a principal component analysis (PCA) based on their
covariance matrix estimated with the $\nu_\ar$ additional realizations
that are extracted from the learned dataset $D_{\nu_\ar}$. Random
vector $\bfQ$ (resp. $\bfW$) is then transformed into a random vector
$\widehat\bfQ$ (resp. $\widehat\bfW$) with values in $\RR^{\nu_q}$
(resp. in $\RR^{\nu_w}$). In general, but depending on the
application, the reduced dimensions are such that
$\nu_q \ll n_q$ and $\nu_w \ll n_w$. It should be noted that a direct
construction by PCA of a reduced-order representation of random vector
$\bfX = (\bfQ,\bfW)$ cannot be done because we need to have a separate
representation for the projected random variable  $\widehat\bfQ$ and
for its counterpart $\widehat\bfW$ in order to be able to write Bayes
formula. Consequently, the random vector $\widehat\bfQ$
(resp. $\widehat\bfW$) is centered with an empirical-estimated
covariance matrix that is the identity matrix $[I_{\nu_q}]$
(reps. $[I_{\nu_w}]$). The centered random variables $\widehat\bfQ$
and $\widehat\bfW$, which are statistically dependent, are then
correlated. This means that the empirical-estimated covariance matrix
$[C_{\widehat X}]$ of random vector $\widehat\bfX =
(\widehat\bfQ,\widehat\bfW)$ is not a diagonal matrix. The $(2\times
2)$ block writing of $[C_{\widehat X}]$ (with respect to
$\widehat\bfQ$ and $\widehat\bfW$) exhibits two block diagonal
identity matrices, namely $[I_{\nu_q}]$ and $[I_{\nu_w}]$, but there
are extradiagonal block matrices that, in general, are not equal to
zero. At this stage, there is an additional difficulty that is related
to the fact that, in general, matrix $ [C_{\widehat X}] $ is not
invertible or is not sufficiently well conditioned to carry out the
algebraic manipulations necessary for the construction of the
posterior pdf based on the use of the Gaussian kernel-density
estimation method, using the learned dataset $D_{\nu_\ar} $. Most
often, in the literature, either the rank of $[C_{\widehat X}]$ is
assumed to be less than $\nu=\nu_q+\nu_w$ (in this case, adapted
algebraic methods have been proposed) or matrix $[C_{\widehat X}]$ is
assumed to be invertible (in that case, there is no
difficulty). However, no adapted method seems to have been proposed
for the "intermediate" case. Therefore, we had to develop a novel
regularization $[\widehat C_{\varepsilon}]$ of $[C_{\widehat X}]$ in
order to achieve the required robustness.\\

\noindent (iii)- To ensure the robustness of proposed methodology,
several ingredients have been analyzed, tested, and validated.

- The first one (as explained above) is related, if necessary, to the
construction of a regularization $[\widehat C_{\varepsilon}]$ in
$\MM^+_{\nu}$ of $[C_{\widehat X}]$ in order to obtain a
positive-definite inverse matrix $[\widehat C_{\varepsilon}]^{-1}$
whose condition number is of order $1$ and for which the value of the
hyperparameter $\varepsilon$ can be set, independently of
applications.

- The second one is related  to the construction of the MCMC generator
for obtaining a robust algorithm for the computation of the
$\nu_\post$ realizations of  $\WW^\post$ whose posterior pdf is
$p^\post_{\widehat\bfW}$. This pdf is explicitly deduced from the
Gaussian kernel-density representation of the joint pdf
$p_{\widehat\bfQ,\widehat\bfW}$ using the $\nu_\ar$ additional
realizations of $(\widehat\bfQ,\widehat\bfW)$ and the $n_r$
experimental realizations of $\widehat\bfQ$. This MCMC generator is
the one (but adapted to the posterior model)  used for the
PLoM. However, it has been seen through many numerical experiments
that a normalization with respect to the covariance matrix of the
posterior model  $\widehat\bfW^\post$ of $\widehat\bfW$ had to be made
in order to improve the robustness of the algorithm. Unfortunately,
although the expression of $p^\post_{\widehat\bfW}$  is explicitly
known, the algebraic calculation of this covariance matrix is not
possible and, as it will be explained in the following, an
approximation has to be constructed. Finally, a statistical reduction
along the data axis is performed using  a diffusion maps basis in
order to avoid a possible scattering of the posterior realizations
generated, which then allows for preserving the concentration of the
posterior probability measure (when such a concentration exists).
\subsection{Organization of the paper}
\label{Section1.4}
\noindent In order to discuss and motivate the intricate interplay
between the requirements presented in Section~\ref{Section1.2}
necessary details concerning PMoL and the various modeling
choices are included in the paper, which is organized as follows.

\noindent Section~\ref{Section2} is devoted to the mathematical
statement of the problem.
In Section~\ref{Section3}, we introduce the scaling of the initial dataset.
Section~\ref{Section4} deals with the generation of additional
realizations for the prior probability model using the probabilistic
learning on manifolds while the reduced-order representations for
$\bfQ$ and  $\bfW$  are constructed in Section~\ref{Section5} using
the learned dataset.
Section~\ref{Section6} is devoted to the Bayesian formulation for the
posterior model and Section~\ref{Section7} deals with the
nonparametric statistical estimation of the posterior pdf using the
learned dataset, for which a regularization model is proposed. The
dissipative Hamiltonian MCMC generator is detailed in
Section~\ref{Section8} for the posterior pdf. The question relative to
the choice of a value of the regularization parameter is analyzed in
Section~\ref{Section9}. Three applications are presented in
Sections~\ref{Section10} and \ref{Section11}. The first two are
relatively simple and can easily be reproduced. The third application
is devoted to the ultrasonic wave propagation in biological tissues
for which $\WW$ is the random vector corresponding to the spatial
discretization of a non-Gaussian tensor-valued random elasticity field
of a cortical bone exhibiting osteoporosis.
In order to retain clarity throughout the paper, several of the
mathematical and algorithmic details have been relegated to 6
appendices. Appendix~A is a summary of the algorithm of the
probabilistic learning on manifolds. In Appendix~B, we give the proof
of the convergence of the random sequence $\bfX^{(\nu_q,\nu_w)}$
related to the introduction of the reduced-order representations of
$\bfQ$ and  $\bfW$. The proof of the range of the values of the
covariance matrix of $\widehat\bfX = (\widehat\bfQ,\widehat\bfW)$ is
detailed in Appendix~C. In Appendix~D, we give the proof of the
consistency of the estimator of the regularized pdf of $\widehat\bfX$,
for which an upper bound is constructed as a function of
$\varepsilon$, $\nu_\ar$, and $\nu$. The construction of the
diffusion-maps basis for the posterior model is detailed in
Appendix~E. Finally, the St\"{o}rmer-Verlet scheme for solving the
reduced-order ISDE is given in Appendix~F.

\section*{Notations}

\noindent Lower-case letters such as $q$ or $\eta$ are deterministic real variables.\\
Boldface lower-case letters such as $\bfq$ or $\bfeta$ are deterministic vectors.\\
Lower-case letters $\qq$, $\ww$, and $\xx$ are deterministic vectors.\\
Upper-case letters such as $Y$ or $H$ are real-valued random variables.\\
Boldface upper-case letters such as $\bfY$ or  $\bfH$ are vector-valued random variables.\\
Upper-case letters $\QQ$, $\UU$, $\WW$, and $\XX$ are vector-valued random variables.\\
Lower-case letters between brackets such as  $[y]$ or $[\eta]$ are deterministic matrices.\\
Boldface upper-case letters between brackets such as $[\bfY]$ or $[\bfH]$ are matrix-valued random variables.\\

\noindent
$n$: dimension ($n=n_q+n_w$) of vector $\bfx$ or $\XX$.\\
$n_q$: dimension of vectors $\bfq$, $\qq$, $\bfQ$, and $\QQ$.\\
$n_r$: number of independent experimental realizations.\\
$n_w$: dimension of vectors $\bfw$, $\ww$, $\bfW$, and $\WW$.\\

\noindent
$\nu$: dimension ($\nu=\nu_q+\nu_w$) of vectors $\widehat\bfx$ and $\widehat\bfX$.\\
$\nu_\ar$: number of additional realizations computed with the PLoM.\\
$\nu_\post$: number of independent realizations for the posterior model.\\
$\nu_q$: dimension of vectors $\widehat\bfq$ and $\widehat\bfQ$ coming from the PCA of $\bfQ$.\\
$\nu_w$: dimension of vectors $\widehat\bfw$ and $\widehat\bfW$ coming from the PCA of $\bfW$.\\
$\nu_x$: dimension of vector $\bfH$ coming from the PCA of $\bfX$ (Appendix~A).\\

\noindent
$[I_{n}]$: identity matrix in $\MM_n$.\\
$\MM_{n,N}$: set of all the $(n\times N)$ real matrices.\\
$\MM_n$: set of all the square $(n\times n)$ real matrices.\\
$\MM_n^+$: set of all the positive-definite symmetric $(n\times n)$ real matrices.\\
$\MM_n^{+0}$: set of all the positive-semidefinite symmetric $(n\times n)$ real matrices.\\
$\RR$: set of all the real numbers.\\
$\RR^n$: Euclidean vector space on $\RR$ of dimension $n$.\\
$[y]_{kj}$: entry of matrix $[y]$.\\
$[y]^T$: transpose of matrix $[y]$.\\
$\delta_{kk'}$: Kronecker's symbol such that $\delta_{kk'} =0$ if $k\not= k'$ and $=1$ if $k=k'$.\\
$E$: Mathematical expectation.\\
$\Vert\bfx\Vert$: usual Euclidean norm in $\RR^n$.\\
$<\! \bfx,\bfy \!> $: usual Euclidean inner product in $\RR^n$.\\
$\Vert [A]\Vert_F$: Frobenius norm of a real matrix $[A]$.\\
$\delta_{kk'}$: Kronecker's symbol.
%
%
\section{Formulation}
\label{Section2}
In this paper, any Euclidean space $\curE$ (such as $\RR^{n_w}$) is
equipped with its Borel field $\curB_\curE$, which means that
$(\curE,\curB_\curE)$ is a measurable space on which a probability
measure can be defined. In this section, we first detail the
mathematical formulation of the problem introduced in
Section~\ref{Section1} and we state the objective.\\

\noindent\textit{Defining the stochastic mapping $\FF$ and the initial
  dataset $\DD_{N_d}$}.
Let $\ww\mapsto\FF(\ww)$ be a stochastic mapping from $\RR^{n_w}$ into
the space $L^2(\Theta,\RR^{n_q})$ of all the second-order random
variables defined on a probability space $(\Theta,\curT, \curP)$ with
values in $\RR^{n_q}$. The
vector $\ww$ (the input) belongs to an admissible set
$\curC_\ww\subset \RR^{n_w}$ and is modeled by a second-order random
variable $\WW = (\WW_1,\ldots ,\WW_{n_w})$ defined on $(\Theta,\curT,
\curP)$ with values in $\RR^{n_w}$, for which the support of its probability distribution
$P_\WW(d\ww)$ is $\curC_\ww$, and which is assumed to be statistically
independent of $\FF$. The quantity of interest (the output) is a
random variable $\QQ=(\QQ_1,\ldots,\QQ_{n_q})$ defined on
$(\Theta,\curT, \curP)$ with values in $\RR^{n_q}$, which is written
as $\QQ = \FF(\WW)$, which is statistically dependent of $\FF$ and
$\WW$, and which is assumed to be of second order. For the problem
considered, the only available information consists of a given
\textit{initial dataset} (training set) constituted of $N_d$
independent realizations $\{ (\qq_q^j,\ww_d^j) , j=1,\ldots , N_d\}$
of random variable $(\QQ,\WW)$ with values in
$\RR^{n_q}\times\RR^{n_w}$.\\

\noindent\textit{Example of stochastic mapping $\FF$ and origin of the given initial dataset $\DD_{N_d}$}.
The stochastic nature of the mapping $\FF$ deserves a
clarification. It is induced by the division of the input random
parameters of a computational model into two separate subsets only one
of which is initially observed, and the influence of the other subset
is manifested as uncertainty about the mapping.
Thus consider, for instance, a large-scale stochastic computational model
of a discretized stochastic physical system for which the random
quantity of interest is written as $\QQ = \ff(\WW,\UU)$. The random
variable $\UU=(\UU_1,\ldots , \UU_{n_u})$ is construed as a hidden
variable defined on $(\Theta,\curT, \curP)$, with values in
$\RR^{n_u}$, with probability distribution $P_\UU(d\uu)$, and which is
statistically independent of $\WW$.  The function $(\ww,\uu)\mapsto
\ff (\ww,\uu)$ is a measurable mapping from $\RR^{n_w}\times\RR^{n_u}$
into $\RR^{n_q}$, which is a representation of the solution of the
stochastic computational model. Consequently, the joint probability
distribution $P_{\WW,\UU}(d\ww,d\uu)$ of $\WW$ and $\UU$ is
$P_{\WW}(d\ww)\otimes P_{\UU}(d\uu)$. For all $\ww$ in $\RR^{n_w}$,
stochastic mapping $\FF$ is such that $\FF(\ww) = \ff(\ww,\UU)$. The
origin of the initial dataset $\DD_{N_d}$ can come from the
computation of $N_d$ independent realizations $\{\qq_d^j ,
j=1,\ldots,N_d\}$ such that $\qq_d^j = \ff(\ww_d^j,\UU(\theta_j))$, in
which $\{\ww_d^j =\WW(\theta_j)\}_j$ are $N_d$ independent
realizations of $\WW$ generated with $P_\WW(d\ww)$,  and where
$\{\UU(\theta_j)\}_j$ are $N_d$ independent realizations of $\UU$
generated with $P_\UU(d\uu)$. It should be noted that realizations
$\{\UU(\theta_j)\}_j$ are not explicitly included in the initial
dataset.\\

\noindent\textit{Introducing the random variable $\XX$ and its realizations}.
We then introduce the random variable $\XX = (\QQ,\WW)$ defined on
$(\Theta,\curT, \curP)$, with values in $\RR^{n}$  ($n=n_q+n_w$), and
for which the probability distribution, $P_\XX(d\xx)$, on $\RR^{n}$ is
unknown, and for which the initial dataset defined by
\begin{equation}
\DD_{N_d} =\{ \xx_d^j = (\qq_d^j,\ww_d^j) , j=1,\ldots , N_d\} \, ,                                                                  \label{EQO1}
\end{equation}
is the only available information.\\

\noindent\textit{Existence hypothesis of probability density function for $\XX$}.
It is assumed that the unknown probability distribution $P_\XX(d\xx)$
admits a density $p_\XX(\xx)$ with respect to the Lebesgue measure
$d\xx$ on $\RR^{n}$. Therefore, the joint probability distribution
$P_{\QQ,\WW}(d\qq,d\ww)$ on $\RR^{n}$ of $\QQ$ and $\WW$ admits a
density $p_{\QQ,\WW}(\qq,\ww)$ with respect to the Lebesgue measure
$d\qq\, d\ww$ on $\RR^{n}$. The probability distributions
$P_\QQ(d\qq)$ and $P_\WW(d\ww)$ of $\QQ$ and $\WW$ admit the densities
$p_\QQ(\qq)=\int p_{\QQ,\WW}(\qq,\ww) \, d\ww$ and $p_\WW(\ww)=\int
p_{\QQ,\WW}(\qq,\ww) \, d\qq$ with respect to the Lebesgue measures
$d\qq$ on $\RR^{n_q}$ and $d\ww$ on $\RR^{n_w}$, respectively. The
conditional pdf $\qq\mapsto p_{\QQ\vert\WW}(\qq\vert \ww)$ on
$\RR^{n_q}$ of $\QQ$ given $\WW=\ww$ in $\curC_\ww \subset \RR^{n_w}$
is such that
$p_{\QQ,\WW}(\qq,\ww) = p_{\QQ\vert\WW}(\qq\vert \ww)\,
p_\WW(\ww)$. Since the support of $p_\WW$ is $\curC_\ww \subset
\RR^{n_w}$, if $\ww$ is given in $\RR^{n_w} \backslash \curC_\ww$,
then $p_\WW(\ww) = 0$, and consequently,
$\qq\mapsto p_{\QQ,\WW}(\qq,\ww)$ is the zero function. It should be noted that
hypothesis $P_\XX(d\xx) = p_\XX(\xx)\, d\xx$ would not be satisfied
if $\FF$ was a deterministic mapping, $\FF(\ww) = \ff(\ww)$
independent of $\UU$, because the support, $\curS_{n_w} = \{
(\ww,\ff(\ww)),\ww\in\curC_\ww\subset\RR^{n_w}\}$ of $P_\XX(d\xx)$ on
$\RR^{n}$, would be the manifold of dimension $n_w$ in $\RR^{n}$,
consisting of the graph of the deterministic mapping $\ff$.\\

\noindent\textit{Specifying the experimental dataset $\DD^\exper_{n_r}$}.
An \textit{experimental dataset} $\DD^\exper_{n_r}$ is given and is
constituted of $n_r$ independent experimental realizations of $\QQ$,
\begin{equation}
\DD^\exper_{n_r} = \{ \qq^{\exper,r}, r=1,\ldots , n_r\} \, ,                                                                  \label{EQO2}
\end{equation}
that are also assumed to be independent of realizations $\{\qq_d^j\}_j$.\\

\noindent\textit{Objective}.
As explained in Section~\ref{Section1}, the objective is to generate
realizations $\{\ww^{\post,\ell},\ell=1,\ldots ,\nu_\post\}$ of the
posterior model of $\WW$ for which the only available information consist
of the initial dataset $\DD_{N_d}$ associated with a prior model of
$\WW$ and of the experimental dataset $\DD^\exper_{n_r}$.
\section{Scaling the initial dataset}
\label{Section3}
Initial dataset $\DD_{N_d}$ can be made up of heterogeneous numerical
values and must be scaled for performing computational statistics.
Let $\xx^\pmax = \max_j\{\xx_d^j\}$, $\xx^\pmin = \min_j\{\xx_d^j\}$,
and $\bfbeta_x = \xx^\pmin$ be a vector in $\RR^{n}$. The diagonal
$(n\times n)$ real matrix $[\alpha_x]_{kk'} = (\xx^\pmax_k
-\xx^\pmin_k)\delta_{kk'}$ is invertible. The scaling of random vector
$\XX$ with values in $\RR^{n}$ is the random vector $\bfX$ with values
in $\RR^{n}$ such that
\begin{equation}
\XX = [\alpha_x]\, \bfX + \bfbeta_x \quad , \quad \bfX =  [\alpha_x]^{-1}(\XX-\bfbeta_x)\, .                           \label{EQ1}
\end{equation}
From Eq.~\eqref{EQ1}, the scaled random variables $\bfQ$ and $\bfW$ with values in $\RR^{n_q}$ and $\RR^{n_w}$ can directly be deduced,
\begin{align}
\QQ= [\alpha_q]\, \bfQ + \bfbeta_q \quad , & \quad \bfQ =  [\alpha_q]^{-1}(\QQ-\bfbeta_q) \, ,                       \label{EQ2} \\
\WW= [\alpha_w]\, \bfW + \bfbeta_w \quad , & \quad \bfW =  [\alpha_w]^{-1}(\WW-\bfbeta_w) \, .                       \label{EQ3}
\end{align}
The $N_d$ realizations of $\bfX$ are then $\{\bfx_d^j\}_j$ with  $\bfx_d^j=  [\alpha_x]^{-1}(\xx_d^j-\bfbeta_x)$. The scaled initial dataset is then defined by
\begin{equation}
D_{N_d} = \{ \bfx_d^j = (\bfq_d^j,\bfw_d^j), j = 1, \ldots , N_d\}\, ,                                                           \label{EQ3bis}
\end{equation}
in which  $\bfq_d^j=  [\alpha_q]^{-1}(\qq_d^j-\bfbeta_q)$ and
$\bfw_d^j=  [\alpha_w]^{-1}(\ww_d^j-\bfbeta_w)$. The collection of these $N_d$ vectors $\{\bfx_d^j\}_j$ in $\RR^{n}$ is represented by the matrix $[x_d]$ such that
\begin{equation}
[x_d] = [\bfx_d^1 \ldots \bfx_d^{N_d}]\in \MM_{n,N_d}\, .                                                                     \label{EQ4}
\end{equation}
In the following, we will use the scaled random variable $\bfX = (\bfQ,\bfW)$ with values in $\RR^{n} = \RR^{n_q}\times \RR^{n_w}$. The experimental dataset $\DD^\exper_{n_r}$ defined in Section~\ref{Section2} is scaled using Eq.~\eqref{EQ2}, yielding the scaled experimental dataset,
\begin{equation}
D^\exper_{n_r} = \{\bfq^{\exper,r},r=1,\ldots , n_r\} \quad , \quad \bfq^{\exper,r} = [\alpha_q]^{-1}(\qq^{\exper,r}-\bfbeta_q) \, .   \label{EQ5}
\end{equation}
If $\QQ = \ff(\WW,\UU)$ (see the example of stochastic function $\FF$ presented in Section~\ref{Section2}), then $\bfQ$ can be rewritten as
\begin{equation}
\bfQ = \bff(\bfW,\UU)\, ,                                                                                              \label{EQ6}
\end{equation}
in which $\bff$ corresponds to the induced transformation of mapping $\ff$.
\section{Generating additional realizations for the prior probability model using the probabilistic learning on manifolds}
\label{Section4}
As explained in Section~\ref{Section1.2}, the framework of this paper is the Bayesian approach for the small-data challenge because $N_d$ is assumed to be small. The Bayesian method allows for updating the prior pdf $p_\bfW$ on $\RR^{n_w}$ of $\bfW$ using experimental dataset $D^\exper_{n_r}$ relative to $\bfQ$ with values in $\RR^{n_q}$ in order to obtain the posterior pdf $p_\bfW^\post$ on $\RR^{n_w}$. Clearly, the posterior pdf strongly depends on the joint pdf $p_{\bfQ,\bfW}$ on $\RR^{n_q}\times\RR^{n_w}$. Consequently, a bigger dataset $D_{\nu_\ar}$ (that we have called  "learned dataset" in Section~\ref{Section1.3}),
\begin{equation}
D_{\nu_\ar} = \{\bfx_\ar^\ell = (\bfq_\ar^\ell,\bfw_\ar^\ell), \ell=1,\ldots ,\nu_\ar\}\, ,                                                \label{EQ7}
\end{equation}
which is made up of $\nu_\ar \gg N_d$ independent realizations of $\bfX = (\bfQ,\bfW)$, is required for the two following reasons:

\noindent - a better estimate of prior pdf $p_\bfW$ has to be constructed using $D_{\nu_\ar}$ instead of $D_{N_d}$.

\noindent - the non-Gaussian conditional pdf $\bfq\mapsto
p_{\bfQ\vert\bfW}(\bfq\vert\bfw)$ on $\RR^{n_q}$ of $\bfQ$ for given
$\bfW=\bfw$ in $\RR^{n_w}$ has to be correctly estimated thus
requiring a big dataset such as $D_{\nu_\ar}$. The use of $D_{N_d}$ for
such an estimation would not be sufficiently "good" because $N_d$ is
assumed to be small.

In this paper, only $D_{N_d}$ and $D^\exper_{n_r}$ are known. In
addition, $D_{N_d}$ is assumed to be constituted of numerical
simulations performed with a large-scale computational model
represented by $\bfQ= \bff(\bfW,\UU)$ (see Eq.~\eqref{EQ6}) in which
$\UU$ is not an "observation noise and model discrepancy", but is for
instance (as explained in Section~\ref{Section1.1}), the spatial
discretization of a non-Gaussian tensor-valued random field that
appears as a coefficient in a partial differential operator in a
stochastic boundary value problem. In this framework, it is important
to preserve the non-Gaussian character of the conditional pdf
$p_{\bfQ\vert\bfW}( \cdot \vert\bfw)$, which is the pdf of random
vector $\bff(\bfw,\UU)$. Since $\bff$ and $\UU$ are unknown (only
$D_{N_d}$ is assumed to be known), we propose to construct the big
dataset (learned dataset) $D_{\nu_\ar}$ of additional realizations
using the probabilistic learning on manifolds \cite{Soize2016}. In
order to facilitate the reading of this paper, a summary of this
algorithm is given in Appendix~A in which we propose numerical values
and identification methods for the parameters involved in the
algorithm.
\section{Reduced-order representations for $\bfQ$ and $\bfW$ using the learned dataset}
\label{Section5}
As explained in Section~\ref{Section1.2}, dimension $n=n_q+n_w$ of random vector $\bfX$ can be high. It is thus necessary to decrease the numerical cost
of the MCMC generator of $p^\post_\bfW$. For that and as explained in Section~\ref{Section1.3}-(ii), a statistical reduction of $\bfQ$ and $\bfW$ is performed using a PCA for which the learned dataset $D_{\nu_\ar}$ is used.
\subsection{PCA of random vector $\bfQ$}
\label{Section5.1}
Let $\underline\bfq_\ar\in\RR^{n_q}$ and $[C_{\bfQ,\ar}]\in\MM^{+0}_{n_q}$ be the empirical estimates of the mean vector and the covariance matrix of $\bfQ$, constructed using the additional realizations $\{\bfq_\ar^\ell, \ell=1,\ldots, \nu_\ar\}$. The PCA representation, $\bfQ^{(\nu_q)}$,  of $\bfQ$ at order $1\leq \nu_q \leq \nu_\ar$ is written as
\begin{equation}
\bfQ^{(\nu_q)} = \underline\bfq_\ar + [\varphi_q]\, [\mu_q]^{1/2}\, \widehat\bfQ\, ,                                        \label{EQ8}
\end{equation}
in which $[\varphi_q]\in \MM_{n_q,\nu_q}$ is the matrix of the eigenvectors of $[C_{\bfQ,\ar}]$ associated with its $\nu_q$ largest eigenvalues
$\mu_{q,1} \geq \mu_{q,2} \geq \ldots \geq \mu_{q,\nu_q} > 0$, represented by the diagonal matrix $[\mu_q]\in\MM_{\nu_q}$. The value of $\nu_q$ is classically calculated in order that the $L^2$-error function $\nu_q\mapsto\err_\bfQ(\nu_q)$ defined by
\begin{equation}
\err_\bfQ(\nu_q) = \frac{E\{\Vert\bfQ - \bfQ^{(\nu_q)}\Vert^2\}}{E\{\Vert\bfQ-\underline\bfq_\ar\Vert^2\}}
                 = 1- \frac{\sum_{\alpha=1}^{\nu_q} \mu_{q,\alpha}}{\tr[C_{\bfQ,\ar}]}\, ,                                    \label{EQ9}
\end{equation}
be smaller than $\varepsilon_q > 0$. In Eq.~\eqref{EQ9}, $\bfQ$ stands for $\bfQ^{(n_q)}$. Since $[\varphi_q]^T[\varphi_q] =[I_{\nu_q}]$, the random variable $\widehat\bfQ$ with values in $\RR^{\nu_q}$ and its $\nu_\ar$ independent realizations are written as
\begin{equation}
\widehat\bfQ = [\mu_q]^{-1/2}\, [\varphi_q]^T (\bfQ- \underline\bfq_\ar )\, ,                                                 \label{EQ10}
\end{equation}
\begin{equation}
\widehat\bfq^\ell = [\mu_q]^{-1/2}\, [\varphi_q]^T (\bfq_\ar^\ell- \underline\bfq_\ar ) \quad , \quad \ell=1,\ldots, \nu_\ar\, .   \label{EQ10bis}
\end{equation}
It can then be deduced that the empirical estimate $\underline{\widehat\bfq}\in\RR^{\nu_q}$ of the mean vector of $\widehat\bfQ$, and the empirical estimate $[C_{\widehat\bfQ}]\in\MM^{+}_{\nu_q}$  of its covariance matrix are such that
\begin{equation}
\underline{\widehat\bfq} = \bfzero \quad , \quad [C_{\widehat\bfQ}] = [I_{\nu_q}]\, .                                              \label{EQ11}
\end{equation}
Therefore, the components $\widehat Q_1,\ldots , \widehat Q_{\nu_q}$ of $\widehat\bfQ$ are centered and uncorrelated but they are
statistically dependent because, in general, $\widehat\bfQ$ is not a Gaussian vector.
\subsection{Projection of experimental dataset $D^\exper_{n_r}$}
\label{Section5.2}
Using the representation of $\bfQ$ (at convergence) defined by Eq.~\eqref{EQ8}, the experimental dataset $D^\exper_{n_r}$ is transformed into the data set
$\widehat D^\exper_{n_r}$ such that
\begin{equation}
\widehat D^\exper_{n_r} = \{\widehat\bfq^{\exper,r}, r=1,\ldots ,n_r\}\, ,                                                             \label{EQ12}
\end{equation}
in which $\widehat\bfq^{\exper,r} \in \RR^{\nu_q}$ is given by
\begin{equation}
\widehat\bfq^{\exper,r} = [\mu_q]^{-1/2}\, [\varphi_q]^T (\bfq^{\exper,r}- \underline\bfq_\ar )  .   \label{EQ13}
\end{equation}
\subsection{PCA of random vector $\bfW$}
\label{Section5.3}
Similarly to the PCA of $\bfQ$, let $\underline\bfw_\ar\in\RR^{n_w}$ and $[C_{\bfW,\ar}]\in\MM^{+0}_{n_w}$ be the empirical estimates of the mean vector and the covariance matrix of $\bfW$, which are constructed using the additional realizations $\{\bfw_\ar^\ell, \ell=1,\ldots, \nu_\ar\}$. The PCA representation, $\bfW^{(\nu_w)}$,  of $\bfW$ at order $1\leq \nu_w \leq \nu_\ar$ is written as
\begin{equation}
\bfW^{(\nu_w)} = \underline\bfw_\ar + [\varphi_w]\, [\mu_w]^{1/2}\, \widehat\bfW\, ,                                        \label{EQ14}
\end{equation}
in which $[\varphi_w]\in \MM_{n_w,\nu_w}$ is the matrix of the eigenvectors of $[C_{\bfW,\ar}]$ associated with its $\nu_w$ largest strictly positive eigenvalues $\mu_{w,1} \geq \mu_{w,2} \geq \ldots \geq \mu_{w,\nu_w} > 0$, represented by the diagonal matrix $[\mu_w]\in\MM_{\nu_w}$. The value of $\nu_w$ is calculated in order that the $L^2$-error function $\nu_w\mapsto\err_\bfW(\nu_w)$ defined by
\begin{equation}
\err_\bfW(\nu_w) = \frac{E\{\Vert\bfW - \bfW^{(\nu_w)}\Vert^2\}}{E\{\Vert\bfW -\underline\bfw_\ar\Vert^2\}}
                 = 1- \frac{\sum_{\alpha=1}^{\nu_w} \mu_{w,\alpha}}{\tr[C_{\bfW,\ar}]}\, ,                                    \label{EQ15}
\end{equation}
be smaller that $\varepsilon_w > 0$. As previously, in Eq.~\eqref{EQ15}, $\bfW$ stands for $\bfW^{(n_w)}$.
Since $[\varphi_w]^T[\varphi_w] =[I_{\nu_w}]$, the random variable $\widehat\bfW$ with values in $\RR^{\nu_w}$ and its $\nu_\ar$ independent realizations are written as
\begin{equation}
\widehat\bfW = [\mu_w]^{-1/2}\, [\varphi_w]^T (\bfW- \underline\bfw_\ar )\, ,                                                 \label{EQ15bis}
\end{equation}
\begin{equation}
\widehat\bfw^\ell = [\mu_w]^{-1/2}\, [\varphi_w]^T (\bfw_\ar^\ell- \underline\bfw_\ar ) \quad , \quad \ell=1,\ldots, \nu_\ar\, .   \label{EQ16}
\end{equation}
Therefore, as previously, the empirical estimate $\underline{\widehat\bfw}\in\RR^{\nu_w}$ of the mean vector of $\widehat\bfW$, and the empirical estimate $[C_{\widehat\bfW}]\in\MM^{+}_{\nu_w}$  of its covariance matrix are such that
\begin{equation}
\underline{\widehat\bfw} = \bfzero \quad , \quad [C_{\widehat\bfW}] = [I_{\nu_w}]\, .                                              \label{EQ17}
\end{equation}
As for $\widehat\bfQ$, the components $\widehat W_1,\ldots , \widehat W_{\nu_w}$ of $\widehat\bfW$ are centered, uncorrelated, and
statistically dependent (in the general case).
\subsection{Convergence of the sequence of random sequence $\{\bfX^{(\nu_q,\nu_w)}\}_{\nu_q,\nu_w}$}
\label{Section5.4}
Let $\bfX^{(\nu_q,\nu_w)}= (\bfQ^{(\nu_q)} ,\bfW^{(\nu_w)} )$ be the random variable with values in $\RR^{n}=\RR^{n_q}\times \RR^{n_w}$. Let
$\err_{\bfX}(\nu_q,\nu_w) = {E\{\Vert\bfX \!- \!\bfX^{(\nu_q,\nu_w)}\Vert^2\}}/{E\{\Vert\bfX\! -\!\underline\bfx_\ar\Vert^2\}}$ be the $L^2$-error function
in which $\underline\bfx_\ar=(\underline\bfq_\ar , \underline\bfw_\ar)\in\RR^{n}=\RR^{n_q}\times \RR^{n_w}$. Taking into account Eqs.~\eqref{EQ9} and \eqref{EQ15}, if $\nu_q$ and $\nu_w$ are such that $\err_{\bfQ}(\nu_q) \leq \varepsilon_q$ and $\err_{\bfW}(\nu_w) \leq \varepsilon_w$, then
\begin{equation}
\err_{\bfX}(\nu_q,\nu_w) \leq \varepsilon_q +\varepsilon_w\, .                                                                      \label{EQ18}
\end{equation}
The proof of this result is given in Appendix~B.
\subsection{Learned dataset for the random vector $\widehat\bfX =(\widehat\bfQ , \widehat\bfW)$ and methodology remark}
\label{Section5.5}
For a fixed level of convergence defined by $\varepsilon_q +\varepsilon_w$, we introduce the learned dataset $\widehat D_{\nu_\ar}$ constituted of the $\nu_\ar$ independent realizations defined by Eqs.~\eqref{EQ10bis} and \eqref{EQ16} for the random vector $\widehat\bfX = (\widehat\bfQ , \widehat\bfW)$ with values in $\RR^{\nu}$ ($\nu = \nu_q+\nu_w$),  such that
\begin{equation}
\widehat D_{\nu_\ar}  = \{ \widehat\bfx^\ell = (\widehat\bfq^\ell,\widehat\bfw^\ell) \,\, , \,\ell =1,\ldots , \nu_\ar\} \, .              \label{EQ19}
\end{equation}
The methodology proposed consists in constructing a MCMC generator of independent realizations
$\{\widehat\bfw^{\post,\ell}\, , \ell=1,\ldots ,\nu_\post\}$ (for a given $\nu_\post$ as big as we want) of the posterior model $\widehat\bfW^\post$ of
$\widehat\bfW$, for which the pdf is $p_{\widehat\bfW}^\post$, using the learned dataset $\widehat D_{\nu_\ar}$
defined by Eq.~\eqref{EQ19} and the experimental dataset $\widehat D^\exper_{n_r}$ defined by Eq.~\eqref{EQ12}.
As soon as these $\nu_\post$ realizations have been generated, the corresponding independent realizations
$\{\widehat\ww^{\post,\ell}\, , \ell=1,\ldots ,\nu_\post\}$ of  $\WW^\post$, given experimental dataset $\DD^\exper_{n_r}$ for $\QQ$, are calculated using Eq.~\eqref{EQ14} and \eqref{EQ3}, by
\begin{align}
\bfw^{\post,\ell} &  = \underline\bfw_\ar + [\varphi_w]\, [\mu_w]^{1/2}\, \widehat\bfw^{\post,\ell}\, ,                  \label{EQ19bis} \\
\ww^{\post,\ell} &= [\alpha_w]\, \bfw^{\post,\ell} + \bfbeta_w  \, .                                                          \label{EQ20}
\end{align}
\section{Bayesian formulation for the posterior model $\widehat\bfW^\post$ of  $\widehat\bfW$ given $\widehat D^\exper_{n_r}$}
\label{Section6}
The classical Bayes formula is used for constructing the pdf
$p_{\widehat\bfW}^\post$ of the posterior model $\widehat\bfW^\post$
of $\widehat\bfW$ with values in $\RR^{\nu_w}$ given the datasets
$\widehat D_{\nu_\ar}$ defined by Eq.~\eqref{EQ19} and $\widehat
D^\exper_{n_r}$ defined by Eq.~\eqref{EQ12}. It is assumed that the
convergence level of $\bfX^{(\nu_q,\nu_w)}$ is sufficient for
substituting $\bfX^{(\nu_q,\nu_w)}$ by $\bfX$ or equivalently, substituting
$\bfQ^{(\nu_q)}$ by $\bfQ$ and $\bfW^{(\nu_w)}$ by $\bfW$. The pdf $p_{\widehat\bfX}$ of $\widehat\bfX$ with respect to the Lebesgue measure $d\widehat\bfx$ on $\RR^{\nu}$ is replaced by its nonparametric estimate using the learned dataset $\widehat D_{\nu_\ar}$. The use of Eqs.~\eqref{EQ8} and \eqref{EQ14} allows for deducing the measurable mapping $\widehat\bff$ from $\RR^{\nu_w}\times\RR^{n_u}$ into $\RR^{\nu_q}$ such that
\begin{equation}
\widehat\bfQ =\widehat\bff(\widehat\bfW,\UU) \, ,                                                                               \label{EQ21}
\end{equation}
in which $\UU$ is the $\RR^{n_u}$-valued random variable defined in Section~\ref{Section2}, which is statistically independent of $\widehat\bfW$.
Let $\ww\mapsto\widehat\bfw=\bfh (\ww)$ be the continuous mapping from $\RR^{n_w}$ into $\RR^{\nu_w}$ defined by Eqs.~\eqref{EQ3} and \eqref{EQ15bis},
that is to say, $\bfh(\ww) = [\mu_w]^{-1/2}\, [\varphi_w]^T (\bfw- \underline\bfw_\ar )$ with
$\bfw =  [\alpha_w]^{-1}(\ww-\bfbeta_w)$. Let $\curC_{\widehat\bfw} =\bfh(\curC_\ww)$ be the subset of $\RR^{\nu_w}$ such that
\begin{equation}
\curC_{\widehat\bfw} = \{\widehat\bfw\in\RR^{\nu_w}\, ; \, \widehat\bfw = \bfh(\ww) \, , \, \ww\in\curC_\ww\subset\RR^{n_w}\}\, .           \label{EQ22}
\end{equation}
Consequently, the support of the prior pdf $\widehat\bfw\mapsto p_{\widehat\bfW}(\widehat\bfw)$ on $\RR^{\nu_w}$ of random variable $\widehat\bfW$ is
$\curC_{\widehat\bfw}\subset\RR^{\nu_w}$. The conditional pdf $\widehat\bfq\mapsto p_{\widehat\bfQ\vert\widehat\bfW}(\widehat\bfq\vert\widehat\bfw)$ of $\widehat\bfQ$ given $\widehat\bfW=\widehat\bfw$ is defined for $\widehat\bfw \in \curC_{\widehat\bfw}$. Taking into account all the hypotheses previously introduced, pdf $p_{\widehat\bfW}^\post$ is given by the Bayes formula that is written, for all $\widehat\bfw$ in $\curC_{\widehat\bfw}$, as
\begin{equation}
p_{\widehat\bfW}^\post(\widehat\bfw) = c_0\, \{ \prod_{r=1}^{n_r} p_{\widehat\bfQ\vert\widehat\bfW}(\widehat\bfq^{\exper,r}\vert\widehat\bfw)\} \,
                     p_{\widehat\bfW}(\widehat\bfw) \, ,                                                                                    \label{EQ23}
\end{equation}
in which $c_0$ is a positive constant of normalization. Let $p_{\widehat\bfQ,\widehat\bfW}$ be the joint pdf of $\widehat\bfQ$ and $\widehat\bfW$ with respect to the Lebesgue measure $d\widehat\bfq\, d\widehat\bfw$ on $\RR^{\nu_q}\times\RR^{\nu_w}$. Then, for all $\widehat\bfw$ in $\curC_{\widehat\bfw}$,
Eq.~\eqref{EQ23} can be rewritten as
\begin{equation}
p_{\widehat\bfW}^\post(\widehat\bfw) = c_0\, \{ \prod_{r=1}^{n_r} p_{\widehat\bfQ,\widehat\bfW}(\widehat\bfq^{\exper,r},\widehat\bfw) \}\,
                     p_{\widehat\bfW}(\widehat\bfw)^{1-n_r} \, .                                                                       \label{EQ24}
\end{equation}
\section{Nonparametric statistical estimation of the posterior pdf of $\widehat\bfW$ using the learned dataset $\widehat D_{\nu_\ar}$}
\label{Section7}
Many works have been published concerning the multidimensional
Gaussian kernel-density estimation method
\cite{Duong2005,Duong2008,Filippone2011,Zougab2014}. However, for the
high dimensional case, we propose to use a constant covariance matrix
that is parameterized by the Silverman bandwidth.
\subsection{Formulation proposed and its difficulties}
\label{Section7.1}
Taking into account Eq.~\eqref{EQ24}, we have to characterize the joint
pdf $p_{\widehat\bfQ,\widehat\bfW}$ that can be deduced from an
estimation of the pdf $p_{\widehat\bfX}$ of $\widehat\bfX =
(\widehat\bfQ,\widehat\bfW)$. The estimate of $p_{\widehat\bfX}$
is constructed using the Gaussian kernel-density estimation method
with the learned dataset $\widehat D_{\nu_\ar}$ defined by
Eq.~\eqref{EQ19}. The construction proposed involves the empirical
covariance matrix $[C_{\widehat\bfX}]$ of $\widehat\bfX$ given by
\begin{equation}
[C_{\widehat\bfX}] = \frac{1}{\nu_\ar -1} \sum_{\ell=1}^{\nu_\ar} (\widehat\bfx^\ell-\underline{\widehat\bfx})\,
                      (\widehat\bfx^\ell-\underline{\widehat\bfx})^T \quad , \quad   \underline{\widehat\bfx} =
                      \frac{1}{\nu_\ar}  \sum_{\ell=1}^{\nu_\ar} \widehat\bfx^\ell \,  .                                     \label{EQ25}
\end{equation}
Taking int account Eqs.~\eqref{EQ11} and \eqref{EQ17},  it can be
deduced that $\underline{\widehat\bfx} =
(\underline{\widehat\bfq},\underline{\widehat\bfw}) = \bfzero$. Matrix
$[C_{\widehat\bfX}]$ is an element of $\MM_{\nu}^{+0}$ or in
$\MM_{\nu}^+$, and can be expressed in block decomposition as,
\begin{equation}
[C_{\widehat\bfX}] =  \left [  \begin{array}{cc}
                               [\, I_q\, ]       & [C_{qw}] \\
                               {[C_{qw}]^T}  &  [I_w]   \\
                               \end{array}
                     \right ] \,  ,                                                                                                       \label{EQ26}
\end{equation}
in which $[C_{qw}]\in\MM_{\nu_q,\nu_w}$ is the covariance matrix of
random vectors $\widehat\bfQ$ and $\widehat\bfW$. By the
Cauchy-Schwarz inequality, we have
\begin{equation}
\vert\, [C_{qw}]_{jk}\, \vert \leq 1 \,\, , \,\, j\in\{1,\ldots,\nu_q\} \,\, , \,\, k\in\{1,\ldots,\nu_w\}\, .                     \label{EQ27}
\end{equation}
Random vectors $\widehat\bfQ$ and $\widehat\bfW$ are statistically
dependent and are also correlated because we have introduced
independent PCA decompositions for $\bfQ$ and $\bfW$. The
following two comments are appropriate at this point.\\

\noindent (i)- If $[C_{\widehat\bfX}]$ was invertible, the estimate $p_{\widehat\bfX}^{(\nu_\arpp)}$ of $p_{\widehat\bfX}$ would be written, for all $\widehat\bfx$ in $\RR^{\nu}$, as \cite{Bowman1997,Scott2015},
\begin{equation}
p_{\widehat\bfX}^{(\nu_\arpp)}(\widehat\bfx) = c_1 \frac{1}{\nu_\ar} \sum_{\ell=1}^{\nu_\arpp} \exp\{-\frac{1}{2s_\ar^2} < \![C_{\widehat\bfX}]^{-1}(\widehat\bfx-\widehat\bfx^\ell) ,(\widehat\bfx-\widehat\bfx^\ell) \! >     \}\,  ,                                     \label{EQ28}
\end{equation}
in which  $c_1 = ( (2\pi)^{\nu/2} s_\ar^{\nu} \sqrt{\det [C_{\widehat\bfX}]} )^{-1}$ and where $s_\ar$ is the Silverman bandwidth that is written as
\begin{equation}
s_\ar = \left(\frac{4}{\nu_\ar(\nu+2)}\right )^{{1}/{(\nu+4)}}\,  .                                                                            \label{EQ29}
\end{equation}
With such a hypothesis, from Eq.~\eqref{EQ28}, it is easy to deduce $p_{\widehat\bfQ\vert\widehat\bfW}^{(\nu_\arpp)}$ and $p_{\widehat\bfW}^{(\nu_\arpp)}$.\\

\noindent (ii)- Unfortunately, in high dimensions, matrix
$[C_{\widehat\bfX}]$ can sometimes be not invertible. More critically,
and also more commonly, $[C_{\widehat\bfX}]$ is invertible in the
computational sense but it is slightly ill-conditioned. All the
numerical experiments that have been conducted have shown that, if
$[C_{\widehat\bfX}]$ is slightly ill-conditioned (for instance, with a
condition number of the order $10^3$ or $10^4$,
which is much smaller that the usual tolerance on the condition
number for computing the inverse of a matrix), and if its inverse
$[C_{\widehat\bfX}]^{-1}$ is still used, then the estimate of
$p_{\widehat\bfX}^{(\nu_\arpp)}$ defined by Eq.~\eqref{EQ28} induces
some difficulties for the MCMC generator of the posterior pdf defined
by Eq.~\eqref{EQ23}. Consequently, we propose to introduce a
regularization of $[C_{\widehat\bfX}]$ that should be viewed as an
essential part of the construction of the estimation
$p_{\widehat\bfX}^{(\nu_\arpp)}$ of $p_{\widehat\bfX}$.
\subsection{Construction of a regularization model of $[C_{\widehat\bfX}]$}
\label{Section7.2}
Let $[\widehat C_{\varepsilon}]$ be a regularization model in
$\MM_{\nu}^+$ of $[C_{\widehat\bfX}]$ such that its condition number
is of order $1$. Therefore, $[\widehat C_{\varepsilon}]^{-1}$ is in
$\MM_{\nu}^+$ and its condition number is also of order $1$. The proposed regularization is constructed as follows. Let us consider the following classical spectral representation of matrix $[C_{\widehat\bfX}]$,
\begin{equation}
[C_{\widehat\bfX}] = [\Phi]\,[\lambda]\, [\Phi]^T\,  ,                                                                                 \label{EQ30}
\end{equation}
in which the real eigenvalues are in decreasing order, $\lambda_1\geq
\lambda_2\geq \ldots \geq \lambda_{\nu}\geq 0$ and where $[\Phi]$ is
the matrix in $\MM_{\nu}$ of the corresponding eigenvectors. Due to of Eqs.~\eqref{EQ26} and \eqref{EQ27}, it is proven in Appendix~C that these eigenvalues are such that
\begin{equation}
0\leq \lambda_j \leq 2 \quad , \quad j\in\{1,\ldots , {\nu}\} \,  .                                                                \label{EQ30bis}
\end{equation}
If $[C_{qw}]$ was the zero matrix in $\MM_{\nu_q,\nu_w}$, then matrix
$[C_{\widehat\bfX}]$ would be the identity matrix and therefore, all
the eigenvalues would be equal to $1$. Since $[C_{qw}]$ is not the
zero matrix and taking into account Eq.~\eqref{EQ30bis}, there exists
and we define (by construction of the regularization model) the
integer $\nu_1$, such that,
\begin{equation}
\lambda_{\nu_1}  \geq 1 \quad , \quad  \lambda_{\nu_1+1} < 1 \quad , \quad \nu_1+1 \leq \nu \,  .                                      \label{EQ30ter}
\end{equation}
The regularization, $[\widehat C_{\varepsilon}]$  of $[C_{\widehat\bfX}]$ is defined by
\begin{equation}
[\widehat C_{\varepsilon}] = [\Phi]\,[\Lambda_\varepsilon]\, [\Phi]^T\,  ,                                                       \label{EQ30quarter}
\end{equation}
in which the diagonal matrix $[\Lambda_\varepsilon]$ is such that
\begin{equation}
[\Lambda_\varepsilon]_{jj} = \lambda_j \,\, , \,\, 1\leq j\leq \nu_1
            \quad  ; \quad   [\Lambda_\varepsilon]_{jj} = \varepsilon^2 \, \lambda_{\nu_1} \,\, , \,\,\nu_1 +1  \leq j \leq \nu\, ,    \label{EQ31}
\end{equation}
in which $\varepsilon\in[\varepsilon_\pmin , 1 [$ were $\varepsilon_\pmin > 0$ is a hyperparameter that controls the regularization and whose value will be of close to $0.5$. The methodology for choosing the value of $\varepsilon$ will be presented in Section~\ref{Section9}. The following properties can then easily be deduced:
\begin{equation}
[\widehat C_{\varepsilon}] \in \MM_{\nu}^+  \quad ,\quad [\widehat C_{\varepsilon}]^{-1}
                             = [\Phi]\,[\Lambda_\varepsilon]^{-1}\, [\Phi]^T \in \MM_{\nu}^+ \, .                               \label{EQ31bis}
\end{equation}
The condition numbers of $[\widehat C_{\varepsilon}]$ and $[\widehat
C_{\varepsilon}]^{-1}$  are thus equal to $\cond ([\widehat
C_{\varepsilon}]) = \lambda_1/(\varepsilon^2\lambda_{\nu_1})$ and
$\cond ([\widehat C_{\varepsilon}]^{-1}) =
\{1/(\varepsilon^2\lambda_{\nu_1})\} /\{1/\lambda_1\}$,
respectively. They clearly satisfy the following equation,
\begin{equation}
\cond ([\widehat C_{\varepsilon}]) = \cond ([\widehat C_{\varepsilon}]^{-1}) \leq \frac{2}{\varepsilon^2}\, .                         \label{EQ31ter}
\end{equation}
For $\varepsilon$ close to $0.5$, the condition number is less that
$8$. We next make four observations relevant to the proposed regularization.
\noindent\paragraph{(i) Remark concerning the Tikhonov regularization}
The Tikhonov regularization $[\widehat C_{\gamma}]$ of $[C_{\widehat\bfX}]$ with respect to its inverse (see for instance \cite{Tikhonov1995}),
would be such that $\widehat\bfy_\gamma = [\widehat C_{\gamma}]^{-1}\, \widehat\bfx$, in which $\widehat\bfy_\gamma$ is the unique solution in $\RR^\nu$ of the optimization problem,
\begin{equation}
\widehat\bfy_\gamma = \min_{\widehat\bfy\in\RR^\nu} \{ \Vert [C_{\widehat\bfX}]\, \widehat\bfy - \widehat\bfx\Vert^2
                                                                    + \gamma^2\Vert \widehat\bfx\Vert^2\}\, ,                        \label{EQ33}
\end{equation}
for any given $\widehat\bfx$ in $\RR^\nu$, where $\gamma > 0$ is the regularization parameter. The unique solution is such that
$([C_{\widehat\bfX}]^2 +\gamma^2\,[I_\nu])\, \widehat\bfy_\gamma = [C_{\widehat\bfX}]\, \widehat\bfx$, which yields
$[\widehat C_{\gamma}]^{-1} = ([C_{\widehat\bfX}]^2 +\gamma^2\,[I_\nu])^{-1}\,[C_{\widehat\bfX}]$. Therefore, for $j=1,\ldots,\nu$, the eigenvalues of $[\widehat C_{\gamma}]^{-1}$ are
$\lambda_j/(\lambda_j^2+\gamma^2)$ while those of $[\widehat C_{\gamma}]^{-1}$ are $\lambda_j +\gamma^2/\lambda_j$. This regularization shows that
$[\widehat C_{\gamma}]^{-1}$ is not positive definite if the rank of $[C_{\widehat\bfX}]$ is less that $\nu$, and that, if the rank of $[C_{\widehat\bfX}]$
were $\nu$, then the condition number $\cond([\widehat C_{\gamma}]^{-1})$ of
$[\widehat C_{\gamma}]^{-1}$, which is equal to
$\{\lambda_1/\lambda_\nu\}\times\{(\lambda_\nu+\gamma^2)/\lambda_1+\gamma^2\}$,
goes to infinity as $\lambda_\nu$ goes to zero, which is antinomic with the property sought. Consequently, the regularization constructed with Eq.~\eqref{EQ33} cannot be used.
\noindent\paragraph{(ii) Interpretation of the proposed regularization model as a  Tikhonov regularization}
Let us assume that the eigenvalues $\lambda_1,\ldots ,\lambda_\nu$ of $[C_{\widehat\bfX}]$ are such that, for $\varepsilon\in [\varepsilon_\pmin , 1 [$ with $\varepsilon_\pmin > 0$, and for $\nu_1$ defined by Eq.~\eqref{EQ30ter}, we have $\varepsilon^2 \,\lambda_j \lambda_{\nu_1} - \lambda_j^2 \geq 0$ for all $j\geq \nu_1 + 1$. Let $\gamma_1,\ldots ,\gamma_\nu$ be the real numbers defined by $\gamma_j=0$ for $j=1,\ldots , \nu_1$ and by
$\gamma_j = (\varepsilon^2 \,\lambda_j \lambda_{\nu_1} - \lambda_j^2)^{1/2}$ for $j=\nu_1+1,\ldots ,\nu$. Let $[\Gamma]$ be the matrix in $\MM_\nu^{+0}$ defined by $[\Gamma] =[\Phi]\,[\gamma]\, [\Phi]^T$ in which $[\gamma]$ is the diagonal matrix such that $[\gamma]_{jk} = \gamma_j \delta_{jk}$.
It can then be seen that the regularization $[\widehat
C_{\varepsilon}]$ defined by Eq.~\eqref{EQ30quarter} is such that, for
all $\widehat\bfx$ in $\RR^\nu$, $\widehat\bfy_\varepsilon = [\widehat
C_{\varepsilon}] \, \widehat\bfx$ in which $\widehat\bfy_\varepsilon$
is given by
\begin{equation}
\widehat\bfy_\varepsilon = \min_{\widehat\bfy\in\RR^\nu} \{ \Vert \,[C_{\widehat\bfX}]\, \widehat\bfy - \widehat\bfx\,\Vert^2
                                                                    + \Vert \,[\Gamma]\,\widehat\bfx\,\Vert^2\}\, .                  \nonumber
\end{equation}
\noindent\paragraph{(iii) Choice of the value of hyperparameter $\varepsilon$ that controls the regularization}
The choice of the value of hyperparameter $\varepsilon$ is presented in Section~\ref{Section9}.
\noindent\paragraph{(iv) Other remarks concerning the possible regularization models}
Other types of regularization models could {\textit{a priori}} be used.

(1) If the rank of $[C_{\widehat\bfX}]$ is less than $\nu$, the
generalized inverse (or pseudo-inverse) of $[C_{\widehat\bfX}]$ (see
for instance Chapter~6, pp. 163-226 in \cite{Puntanen2005}) could be
used. Such an approach would lead us to introduce a new
parameterization of a submanifold for $\widehat\bfX$ whose dimension
would be the rank of $[C_{\widehat\bfX}]$. The estimation
$p_{\widehat\bfX}^{(\nu_\arpp)}$ of pdf $p_{\widehat\bfX}$ could then
be constructed by using, for instance, the approach proposed in
\cite{Ozakin2009}. Nevertheless, not only the construction of the pdf
$p_{\widehat\bfW}^{(\nu_\arpp)}$ of $\bfW$ would require an
integration on the submanifold, which would induce difficulties, but
above all, the "separation" of the representations of $\widehat\bfQ$
and $\widehat\bfW$ would be lost, and such a "separation" is necessary
for our purpose. Moreover, this approach would be equivalent to doing a PCA of random vector $\bfX$ instead of two PCAs, one on $\bfQ$ and the other one on $\bfW$, a method that cannot be done as we have explained in Section~\ref{Section5}.

(2) A more classical regularization of $[C_{\widehat\bfX}]$ would consist in taking
$[\widehat C_{\eta}] = [C_{\widehat\bfX}] + [C_{\eta}]$ with $[C_{\eta}]$ a covariance matrix in $\MM_\nu^+$. A choice could be
$[C_{\eta}] = \eta^2\, [I_\nu]$. Such a model corresponds to the
introduction of an additional Gaussian noise represented by the random
vector $\widehat\bfB_\eta$ independent of $\widehat\bfX$, such that
$\widehat\bfX_\eta = \widehat\bfX +\widehat\bfB_\eta$ (taking into
account the Gaussian kernel-density estimation used for the estimate
$p_{\widehat\bfW}^{(\nu_\arpp)}$ of $p_{\widehat\bfX}$ defined by
Eq.~\eqref{EQ28}). The numerical evaluation of such a regularization
has been used for the applications presented in
Sections~\ref{Section10} and \ref{Section11}, and has demonstrated a
lack of robustness when used with the MCMC generator of $p_{\widehat\bfW}^\post$.

(3)  A regularization of the probability measure $p_{\widehat\bfX}^{(\nu_\arpp)}(\widehat\bfx)\, d\widehat\bfx$ could also be constructed using the Rao metric between two probability distributions \cite{Rao1949}, which involves the Fisher information matrix. Nevertheless, the algebraic expression of
$p_{\widehat\bfX}^{(\nu_\arpp)}$ given by Eq.~\eqref{EQ28} is not easy due to the presence of the summation over the $\nu_\ar$ realizations.
In a similar framework, another way would have been to use the Riemann metric related to the geodesic distance on the manifold related to the positive-definite matrices \cite{Bhatia2009}, which is particularly well adapted to the Gaussian case as proposed, for instance in \cite{Spantini2017}, but which induces difficulties for the non-Gaussian probability measure $p_{\widehat\bfX}^{(\nu_\arpp)}(\widehat\bfx)\, d\widehat\bfx$.
\subsection{Construction of the regularized estimate $p_{\widehat\bfX}^{(\nu_\arpp)}$ of the pdf $p_{\widehat\bfX}$ of $\widehat\bfX$}
\label{Section7.3}
The regularized estimate of $p_{\widehat\bfX}^{(\nu_\arpp)}$ defined
by Eq.~\eqref{EQ28} is obtained by using the procedures detailed in Section~\ref{Section7.2}.
For $\varepsilon$ fixed in $[\varepsilon_\pmin , 1 [$, let $[G]$ be the $(\nu\times\nu)$ real matrix such that
\begin{equation}
[G] = [\widehat C_{\varepsilon}]^{-1} \in \MM_\nu^+\quad , \quad
                                           [G]^{-1} = [\widehat C_{\varepsilon}]\in \MM_\nu^+ \, ,                        \label{EQ34}
\end{equation}
in which $[\widehat C_{\varepsilon}]$ is defined by Eqs.~\eqref{EQ30quarter} and \eqref{EQ31}. In these conditions, the regularized expression of
$p_{\widehat\bfX}^{(\nu_\arpp)}$ defined by  Eq.~\eqref{EQ28} is written (keeping the same notation) as
\begin{equation}
p_{\widehat\bfX}^{(\nu_\arpp)}(\widehat\bfx) = c_2 \frac{1}{\nu_\ar} \sum_{\ell=1}^{\nu_\arpp} \exp\left\{-\frac{1}{2s_\ar^2} < \![G](\widehat\bfx-\widehat\bfx^\ell) ,(\widehat\bfx-\widehat\bfx^\ell) \! >      \right\}\,  ,                                     \label{EQ35}
\end{equation}
in which  $s_\ar$ is the Silverman bandwidth defined by Eq.~\eqref{EQ29} and where
\begin{equation}
c_2 = \frac{\sqrt{\det [G]}}{s_\ar^\nu\,(2\pi)^{\nu/2} }\,  .                                                            \label{EQ35.1}
\end{equation}
In Eqs.~\eqref{EQ34} and \eqref{EQ35.1}, matrix $[G]$ and pdf
$p_{\widehat\bfX}^{(\nu_\arpp)}$ depend on $\varepsilon$, which will
be omitted for notationa clarity. Let $\widehat\bfX^1, \ldots , \widehat\bfX^{\nu_\arp}$ be $\nu_\ar$ independent copies of random variable $\widehat\bfX$ whose pdf is $p_{\widehat\bfX}$. For all $\widehat\bfx$ fixed in $\RR^\nu$, let $P_{\nu_\arp}(\widehat\bfx)$ be the estimator (positive-valued random variable) corresponding to the estimation $p_{\widehat\bfX}^{(\nu_\arpp)}(\widehat\bfx)$ defined by Eq.~\eqref{EQ35}, such that
\begin{equation}
P_{\nu_\arp}(\widehat\bfx) = c_2 \frac{1}{\nu_\ar} \sum_{\ell=1}^{\nu_\arpp}
      \exp\left\{-\frac{1}{2s_\ar^2} < \![G](\widehat\bfX^\ell-\widehat\bfx) ,(\widehat\bfX^\ell-\widehat\bfx) \! >      \right\}\,  .                 \label{EQ35.2}
\end{equation}
It is proved in Appendix~D that,
\begin{equation}
E\{ (P_{\nu_\arp}(\widehat\bfx) - \underline{P}_{\nu_\arp}(\widehat\bfx))^2\}  \leq
\left \{ \! \frac{1}{\nu_\ar} \!\right \}^{\! 4/(\nu+4)}\!
\left \{\! \frac{\nu \! + \! 2}{4}\! \right \}^{\! \nu/(\nu+4)} \,
\!\!\frac{\sqrt{\det [G]}}{(2\pi)^{\nu/2}} \,\underline{P}_{\nu_\arp}(\widehat\bfx)  \, ,                                           \label{EQ35.3}
\end{equation}
in which $\underline{P}_{\nu_\arp}(\widehat\bfx) =
E\{P_{\nu_\arp}(\widehat\bfx)\}$, as defined by Eq.~\eqref{EQD5} of
Appendix~D, is the mean value that tends to
$p_{\widehat\bfX}(\widehat\bfx)$ when $\nu_\ar$ goes to infinity and
consequently, the estimator is asymptotically unbiased  and
consistent. Due to the mean-square convergence of the sequence of
random variables $\{P_{\nu_\arp}(\widehat\bfx)\}_{\nu_\arp}$, as
implied by Eq.~\eqref{EQ35.3}, this sequence of estimators converges in probability to  $p_{\widehat\bfX}(\widehat\bfx)$.\\

\noindent \textit{Remark}. Below, for notational clarity, $p_{\widehat\bfX}^{(\nu_\arpp)}(\widehat\bfx)$ will simply be denoted by
$p_{\widehat\bfX}(\widehat\bfx)$, which also means that $\nu_\ar$ is chosen sufficiently large for writing that
$p_{\widehat\bfX}^{(\nu_\arpp)} \simeq p_{\widehat\bfX}$. The $\nu_\ar$-dependence of $p_{\widehat\bfQ,\widehat\bfW}$, $p_{\widehat\bfQ\vert\widehat\bfW}$,
and $p_{\widehat\bfW}$ will also be omitted.
\subsection{Deducing the pdf $p_{\widehat\bfQ,\widehat\bfW}$ of $(\widehat\bfQ,\widehat\bfW)$ and the pdf $p_{\widehat\bfW}$ of $\widehat\bfW$}
\label{Section7.4}
Vector $\widehat\bfx$ and realization $\widehat\bfx^\ell$ in $\RR^\nu$ can be decomposed as
$\widehat\bfx = (\widehat\bfq,\widehat\bfw)$ and $\widehat\bfx^\ell = (\widehat\bfq^\ell,\widehat\bfw^\ell)$ in which
$(\widehat\bfq,\widehat\bfw)$ and $(\widehat\bfq^\ell,\widehat\bfw^\ell)$ belong to $\RR^{\nu_q}\times \RR^{\nu_w}$ with $\nu=\nu_q+\nu_w$.
The $(\nu_q\times\nu_w)$ block notation of matrix $[G]$ is introduced as
\begin{equation}
[G] =  \left [  \begin{array}{cc}
                               [G_q]       & [G_{qw}] \\
                               {[G_{qw}]^T}  &  [G_w]   \\
                               \end{array}
                     \right ] \,  .                                                                                                       \label{EQ36}
\end{equation}
Since $[G]\in\MM_{\nu}^+$, we have
\begin{equation}
[G_q]\in\MM_{\nu_q}^+ \quad , \quad [G_w]\in\MM_{\nu_w}^+ \, .                                           \label{EQ37}
\end{equation}
From Eq.~\eqref{EQ35} and taking into account Eqs.~\eqref{EQ36}-\eqref{EQ37}, the joint pdf $p_{\widehat\bfQ ,\widehat\bfW}$ of
$\widehat\bfQ$ and $\widehat\bfW$ (with respect to the Lebesgue measure $d\widehat\bfq \, d\widehat\bfw$ on  $\RR^{\nu_q}\times \RR^{\nu_w}$)
can be written, for all $\widehat\bfq \in \RR^{\nu_q}$ and $\widehat\bfw \in \RR^{\nu_w}$, as
\begin{equation}
p_{\widehat\bfQ ,\widehat\bfW}(\widehat\bfq,\widehat\bfw) = c_2 \frac{1}{\nu_\ar} \sum_{\ell=1}^{\nu_\arpp}
      \exp\left\{-\frac{1}{2s_\ar^2} \, \psi(\widehat\bfq -\widehat\bfq^\ell,\widehat\bfw-\widehat\bfw^\ell)\right\}\,  ,                           \label{EQ38}
\end{equation}
in which the real-valued function $(\widehat\bfq,\widehat\bfw)\mapsto \psi(\widehat\bfq,\widehat\bfw)$ defined
on $\RR^{\nu_q}\times \RR^{\nu_w}$ is defined as
\begin{equation}
\psi(\widehat\bfq,\widehat\bfw) = < \!\! [G_q]\,\widehat\bfq \,,\widehat\bfq \! >
                               + 2 < \!\! [G_{qw}]^T\,\widehat\bfq \,,\widehat\bfw \! >
                               + < \!\! [G_w]\,\widehat\bfw \,,\widehat\bfw \! > \,  .                                                         \label{EQ39}
\end{equation}
Moreover, the prior pdf $p_{\widehat\bfW}$ of $\widehat\bfW$ (with
respect to $d\widehat\bfw$) can be expressed as,
\begin{equation}
p_{\widehat\bfW}(\widehat\bfw) = \int_{\RR^{\nu_q}} p_{\widehat\bfQ ,\widehat\bfW}(\widehat\bfq ,\widehat\bfw) \, d\widehat\bfq\,  .       \label{EQ40}
\end{equation}
From Eqs.~\eqref{EQ38} to \eqref{EQ40}, since matrix $[G]$ is positive
definite, the right-hand side of Eq.~\eqref{EQ40} can be explicitly calculated
\cite{Puntanen2005},
\begin{equation}
p_{\widehat\bfW}(\widehat\bfw) = c_3 \frac{1}{\nu_\ar} \sum_{\ell=1}^{\nu_\arpp} \exp\left\{-\frac{1}{2s_\ar^2} < \![G_0](\widehat\bfw-\widehat\bfw^\ell) ,(\widehat\bfw-\widehat\bfw^\ell) \! >      \right\}\,  ,                                                                                           \label{EQ41}
\end{equation}
in which $c_3$ is the constant of normalization and where $[G_0]$ is a
positive-definite matrix that is constructed as the following Schur complement,
\begin{equation}
[G_0] = [G_w]  - [G_{qw}]^T\, [G_q]^{-1} \, [G_{qw}]  \in \MM_{\nu_w}^+ \,  .                                                             \label{EQ42}
\end{equation}
\section{Dissipative Hamiltonian MCMC generator for the posterior pdf of $\widehat\bfW$}
\label{Section8}
In Section~\ref{Section8.1},  an MCMC generator of the posterior model
$\widehat\bfW^\post$ of $\widehat\bfW$ is presented, which is based on
a nonlinear It\^o stochastic differential equation (ISDE)
corresponding to a stochastic dissipative Hamiltonian dynamical system
for a stochastic process $\{ [\bfU(t)] , t\in\RR^+\}$ with values in
$\MM_{\nu_w,N_s}$. The number, $N_s$, of columns of $ [\bfU(t)]$ is
chosen sufficiently large (but such that $N_s\leq \nu_\ar$) in order
to increase the exploration of space $\RR^{\nu_w}$ by the MCMC algorithm
and to facilitate the construction of a reduced-order nonlinear ISDE using the diffusion-maps basis.

The posterior pdf $p_{\widehat\bfW}^\post$ defined by Eq.~\eqref{EQ41}
could require a large number of increments in the MCMC generator if
the "distance" of experimental dataset $\DD^\exper_{n_r}$ to initial
dataset $\DD_{N_d}$ is too large. For decreasing the computational burden,
the nonlinear ISDE has to be adapted with respect to the covariance matrix of $\widehat\bfW^\post$. Nevertheless, this covariance matrix is unknown and consequently, an appropriate method has to be developed for estimating an approximation of it.
Such a relatively classical problem has been addressed  for the case of Gaussian likelihoods (see for instance \cite{Flath2011}) and more recently, for non-Gaussian likelihoods  in \cite{Arnst2017} within the parametric framework.
In the present work devoted to the non-Gaussian likelihood in high dimension and in a nonparametric framework, the proposed approach consists in constructing a nonlinear ISDE adapted to the mean value and to the covariance matrix of  $\widehat\bfW^\post$, which we will call, \textit{adapted nonlinear ISDE}.  The use of an affine transformation, $\widehat\bfW^\post  = \bfu_T + [A]^{-T} \, \bfS^\post$ (constructed in Section~\ref{Section8.2}), which introduces the matrix-valued stochastic process $\{[\bfS(t)], t\in \RR^+\}$ such that $[\bfU(t)] = [u_T] + [A]^{-T} \, [\bfS(t) ]$, will transform the adapted nonlinear ISDE related to the MCMC generator of $\widehat\bfW^\post$ into a nonlinear  ISDE for the MCMC generator of $\bfS^\post$ that is a non-Gaussian $\RR^{\nu_w}$-valued random variable $\bfS^\post$, "close to" a centered random vector with an identity covariance matrix.

Finally, in order to avoid the data scattering during the generation of independent realizations of $[\bfS]$, in Section~\ref{Section8.3}, the nonlinear ISDE related to stochastic process $\{[\bfS(t)], t\in \RR^+\}$ will be projected on an diffusion-maps basis similarly to the methodology of probabilistic learning on manifolds summarized in Appendix~A. The final generation of realizations $\widehat\bfW^\post$ is summarized in Section~\ref{Section8.4}.
\subsection{Criteria for choosing a value of $N_s$}
\label{Section8.1}
A natural choice would be $N_s=\nu_\ar$. Nevertheless, in general, the number $\nu_\ar$ of additional realizations generated by the PLoM is chosen very large in order to obtain a good convergence of the statistical estimate of the probability distribution of the posterior model. Although such a choice is
always possible, it will always induce a significant increase in
computational requirements, often without attaining commensurate gains
for the MCMC generator. The choice, $N_s=N_d$, is logical and efficient because the generation of the additional realizations is done with this value by the PLoM (see Appendix~A). The choice can also be highlighted by the following criterion. The empirical estimate $[C_{\widehat\bfW}]$ of the covariance matrix of $\widehat\bfW$, performed with $\{\widehat\bfw^\ell ,\ell=1,\ldots , \nu_\ar\}$, is the identity matrix (see Eq.~\eqref{EQ17}). Let $[C^{N_s}_{\widehat\bfW}]$ be the empirical covariance matrix estimated with $\{\widehat\bfw^{\nu_\arp - j + 1} , j=1,\ldots , N_s\}$. Integer $N_s$ can then be chosen such that
$\Vert [C^{N_s}_{\widehat\bfW}] - [I_{\nu_w}]\Vert_F / \Vert
[I_{\nu_w}]\Vert_F < \varepsilon_{N_s}$. It can easily be seen that there
exists $0 < \varepsilon_{N_s} < 1$ such that $N_s=N_d$ (for instance
when $N_d=200$ and $\nu_\ar=30\,000$, $\varepsilon_{N_s}
=0.05$). Alternatively, a value of $N_s$ can be assessed, using this
same criterion, for a predetermined value of  $\varepsilon_{N_s}$.
\subsection{Adapted nonlinear ISDE as the MCMC generator of $\widehat\bfW^\post$}
\label{Section8.2}
The nonlinear ISDE of the MCMC generator of $\widehat\bfW^\post$ is constructed as proposed in \cite{Soize2008b,Soize2015}, which is based on
the works \cite{Soize1994} (in which more general stochastic Hamiltonian dynamical systems are analyzed, in particular with a general mass operator that we use hereinafter). The adapted nonlinear ISDE is deduced from it using a similar normalization as the one proposed by Arnst \cite{Arnst2017}. Nevertheless, in the present non-Gaussian case, the drift vector of the nonlinear ISDE is completely different and the affine transformation for centering and normalizing the posterior model is not the same. We then introduce the matrix $[A]$ that appears in the affine transformation $\widehat\bfW^\post  = \bfu_T + [A]^{-T} \, \bfS$ mentioned above. The method presented in Section~\ref{Section8.2} for constructing $[K]$ (and thus, $[A]$) is also different.

Let $[K]$ be a given matrix in $\MM_{\nu_w}^+$  and let us consider its Cholesky factorization
\begin{equation}
[K] = [A]\, [A]^T\,  .                                                                                                             \label{EQ43}
\end{equation}
Consequently, the inverse matrices $[K]^{-1}$ and $[A]^{-1}$ exist. As explained above, matrix $[K]$, which is constructed in Section~\ref{Section8.2}, will be an approximation of the inverse of the covariance matrix of $\widehat\bfW^\post$.
We consider, for $t > 0$, the nonlinear stochastic dissipative Hamiltonian dynamical system represented by the following nonlinear ISDE,
\begin{align}
d[\bfU(t)] & = [K]^{-1}\, [\bfV(t)] \, dt  \, ,                                                                                  \label{EQ44}\\
d[\bfV(t)] & = [L([\bfU(t)])]\, dt -\frac{1}{2} \,f_0^\post [\bfV(t)]\, dt + \sqrt{f_0^\post} \, [A]\, d[\bfW^\wien(t)] \, ,     \label{EQ45}
\end{align}
with the initial condition at $t=0$,
\begin{equation}
[\bfU(0)]  = [\widehat w_0] \quad , \quad  [\bfV(0)]  = [\widehat v_0]  \quad , a.s.  \, ,                                           \label{EQ46}
\end{equation}
in which:
\begin{enumerate}

\item[(i)] $f_0^\post > 0$ is a free parameter allowing the dissipation  to be controlled in the stochastic dynamical system. This parameter is chosen such that $f_0^\post < 4$. The value, $4$, of the upper bound corresponds to the critical damping rate for the linearized ISDE in terms of stochastic process $[\bfS]$ (see Section~\ref{Section8.3.3}).
\item[(ii)] $\{[\bfW^\wien(t)] , t\in\RR^+\}$ is the stochastic process, defined on $(\Theta,\curT,\curP)$, indexed by $\RR^+$, with values in
$\MM_{\nu_w,N_s}$, for which the columns of $[\bfW^\wien(t)]$ are $N_s$ independent copies of the $\RR^{\nu_w}$-valued normalized Wiener process $\{\bfW^\wien(t) , t\in\RR^+\}$ whose matrix-valued autocorrelation function
is such that $[R_{\bfW^\pwien}(t,t')] = E\{\bfW^\wien(t) \, \bfW^\wien(t')^T\}  =\min (t,t')\, [I_{\nu_w}]$.
\item[(iii)] $[u] \mapsto [L([u])]$ is a mapping from $\MM_{\nu_w,N_s}$ into $\MM_{\nu_w,N_s}$, which depends on $p_{\widehat\bfW}^\post$ and which is defined as follows. The posterior pdf $p_{\widehat\bfW}^\post$ defined by Eq.~\eqref{EQ24} is written as
\begin{equation}
p_{\widehat\bfW}^\post(\widehat\bfw)   = c_0\, p(\widehat\bfw) \quad , \quad
p(\widehat\bfw)  = \{\prod_{r=1}^{n_r} p_{\widehat\bfQ,\widehat\bfW}(\widehat\bfq^{\exper,r},\widehat\bfw) \}\,
                     p_{\widehat\bfW}(\widehat\bfw)^{1-n_r} \, .                                                                          \label{EQ50}
\end{equation}
Let $\widehat\bfw\mapsto\curV(\widehat\bfw)$ be the potential function on $\RR^{\nu_w}$ such that
\begin{equation}
p(\widehat\bfw)  = e^{-\curV(\widehat\bfw)} \quad , \quad \curV(\widehat\bfw) = -\log p(\widehat\bfw)  \, .                             \label{EQ51}
\end{equation}
The matrix $[u]$ is written as $[\bfu^1 \ldots \bfu^{N_s}]$ with $\bfu^j = (u_1^j,\ldots u_{\nu_w}^j)\in \RR^{\nu_w}$. Thus, mapping $[L]$ is defined, for all $[u]$ in $\MM_{\nu_w,N_s}$, as
\begin{equation}
[L([u])]_{kj} = -\frac{\partial}{\partial u^j_k}  \curV(\bfu^j) \quad , \quad  k=1,\ldots,\nu_w \quad , \quad j=1,\ldots , N_s \, ,            \label{EQ52}
\end{equation}
which can be rewritten as
\begin{equation}
[L([u])]_{kj} = \frac{1}{p(\bfu^j) } \{\bfnabla_{\bfu^j}p(\bfu^j) \}_k \, .                                                                \label{EQ53}
\end{equation}
For $j$ fixed in $\{1,\ldots , N_s\}$, the Hamiltonian of the associated conservative homogeneous dynamical system related to
stochastic process $\{(\bfU^j(t),\bfV^j(t)),$ $t\in\RR^+\}$ is thus written as
$\HH(\bfu^j,\bfv^j) = \frac{1}{2} < \! [K]^{-1}\bfv^j,\bfv^j\! > + \curV(\bfu^j)$.\\
\item[(iv)] $[\widehat w_0]\in \MM_{\nu_w,N_s}$ is defined by
$[\widehat w_0] = [\widehat\bfw^{\nu_\arp} \ldots  \widehat\bfw^{\nu_\arp - N_s + 1} ]$,
in which the $N_s$ columns correspond to the $N_s$ last additional realizations
$\{\widehat\bfw^{\nu_\arp - j + 1} ,$ $j=1,\ldots , N_s\}$
generated by the probabilistic learning on manifolds (See Section~\ref{Section4}).\\
\item[(v)] $[\widehat v_0]\in \MM_{\nu_w,N_s}$ is any realization  of a random matrix $[\widehat\bfV_0]$ independent of process
$[\bfW^\wien]$, for which the columns $\{\widehat\bfV_0^j , j=1,\ldots,N_s\}$ are $N_s$ independent Gaussian centered $\RR^{\nu_w}$-valued random variables such that the covariance matrix of $\widehat\bfV_0^j$ is $[K]^{-1}$ for all $j$.
\end{enumerate}

\noindent It can be proven (see Theorems 4 to 7 in Pages 211 to 216 and the invariant measure Page 240 of \cite{Soize1994}) that the nonlinear ISDE defined by Eqs.~\eqref{EQ44} to \eqref{EQ46} admits the unique invariant measure,
\begin{equation}
\otimes_{j=1}^{N_s} \, \{ p_{\widehat\bfW}^\post (\bfu^j) \, p_{\widehat\bfV}(\bfv^j)\, d\bfu^j\, d\bfv^j \}\, ,            \label{EQ54}
\end{equation}
in which $p_{\widehat\bfV}(\bfv^j) = (2\pi)^{-\nu_w/2}\, \exp\{
-{1}/{2} < \! [K]^{-1}\bfv^j,\bfv^j\! >\}$. In addition, these Theorems can be used to show that
Eqs.~\eqref{EQ44} to \eqref{EQ46} have a unique solution $\{ ([\bfU(t)], $ $[\bfV(t)]) , t\in\RR^+\}$, which is a second-order diffusion stochastic process that is asymptotic for $t\rightarrow +\infty$ to the stationary stochastic process $\{ ([\bfU_\st(t_\st)],[\bfV_\st(t_\st)]),$ $t_\st\in\RR^+\}$ for the right-shift semi-group on $\RR^+$. For all fixed $t_\st$, the joint probability distribution of the random matrices
$[\bfU_\st(t_\st)]$ and $[\bfV_\st(t_\st)]$ is the invariant measure defined by Eq.~\eqref{EQ54} and the probability distribution of random matrix $[\bfU_\st(t_\st)]$ is
\begin{equation}
\otimes_{j=1}^{N_s} \,  p_{\widehat\bfW}^\post (\bfu^j)\, d\bfu^j\, ,                                                       \label{EQ55}
\end{equation}
that is to say, is the probability distribution of the  random matrix
$[\widehat\bfW^\post]$ with values in $\MM_{\nu_w,N_s}$, for which the columns $\widehat\bfW^{\post,1}, \ldots ,\widehat\bfW^{\post,N_s}$
are $N_s$ independent copies of random vector $\widehat\bfW^\post$ whose pdf is $p_{\widehat\bfW}^\post$ defined by Eq.~\eqref{EQ50}. It can then be deduced that, for any fixed $t_\st$,
\begin{equation}
 [\widehat\bfW^\post] =  [\bfU_\st(t_\st)] = \lim_{t\rightarrow+\infty }  [\bfU(t)]  \, .                                     \label{EQ56}
\end{equation}
Equation~\eqref{EQ56} implies that Eqs.~\eqref{EQ44} to \eqref{EQ46}
represent an MCMC generator for $p_{\widehat\bfW}^\post$. The free
parameter $f_0$ allows for controlling the transient response
generated by the initial condition for quickly reaching the stationary
solution (note that the invariant measure is independent of $f_0$). It
can also be proven that the asymptotic stationary solution is ergodic \cite{Khasminskii2012}.\\

\noindent\textit{Expression of the mapping $[L]$ adapted to computation}.
An explicit algebraic expression is constructed for the mapping $[u]\mapsto [L([u])]$ defined by Eq.~\eqref{EQ53}, using Eqs.~\eqref{EQ50} for $p$, Eqs.~\eqref{EQ38} and \eqref{EQ39} for $p_{\widehat\bfQ,\widehat\bfW}$, and Eqs.~\eqref{EQ41} and \eqref{EQ42} for $p_{\widehat\bfW}$. These equations show the presence of a summation of exponential terms (summation over the number $\nu_\ar$ of realizations $\widehat\bfq^\ell$ and $\widehat\bfw^\ell$ of $\widehat\bfQ$ and $\widehat\bfW$). Consequently, an adapted algebraic representation must be developed in order to minimize the numerical cost for each evaluation of $[L([u])]$ and to avoid numerical noise, overflow, and underflow during the computation. Several expressions have been developed and evaluated. We present the most efficient one with respect to the above criteria.
For $k=1,\ldots,\nu_w$, for $j=1,\ldots, N_s$, and for $[u]=[\bfu^1,\ldots ,\bfu^{N_s}]$ in $\MM_{\nu_w,N_s}$,
\begin{equation}
 [L([u])]_{kj}=  \frac{1}{s_\ar^2} \{-[G_{0w}]\,\bfu^j-\bfb^\exper+(1-n_r)\,\bfa_0(\bfu^j)+\sum_{r=1}^{n_r} \bfa_1^r(\bfu^j) \}_k \, ,        \label{EQ60}
\end{equation}
where $\bfa_0(\bfu^j) = (a_{0,1}(\bfu^j) ,\ldots , a_{0,\nu_w}(\bfu^j))$ and
$\bfa_1^r(\bfu^j) = (a_{1,1}^r(\bfu^j) ,\ldots , a_{1,\nu_w}^r(\bfu^j))$  are vectors in $\RR^{\nu_w}$ such that
\begin{equation}
 \bfa_0(\bfu^j) = \left(\sum_{\ell=1}^{\nu_\ar} \widetilde\bfw_0^\ell \, \zeta_0^\ell(\bfu^j)\right)\,
                   \left(\sum_{\ell=1}^{\nu_\ar}  \zeta_0^\ell(\bfu^j)\right)^{-1} \, ,                                                                \label{EQ61}
\end{equation}
and for $r\in\{1,\ldots, n_r\}$,
\begin{equation}
 \bfa_1^r(\bfu^j) = \left(\sum_{\ell=1}^{\nu_\ar} \widetilde\bfw_1^\ell \, \zeta_1^{r\ell}(\bfu^j)\right)\,
                   \left(\sum_{\ell=1}^{\nu_\ar}  \zeta_1^{r\ell}(\bfu^j)\right)^{-1} \, .                                                           \label{EQ62}
\end{equation}
- In Eq.~\eqref{EQ60}, the symmetric $(\nu_w\times\nu_w)$ real matrix $[G_{0w}]$ is given by
\begin{equation}
[G_{0w}] = (1-n_r)\, [G_0] + n_r\, [G_w]\, .                                                                                               \label{EQ63}
\end{equation}
From Eqs.~\eqref{EQ37} and \eqref{EQ42}, it can be deduced that $[G_{0w}]\in\MM_{\nu_w}^+$ for $n_r \geq 2$.
The vector $\bfb^\exper\in\RR^{\nu_w}$ is given by
\begin{equation}
\bfb^\exper = [G_{qw}]^T \, \sum_{r=1}^{n_r}\, \widehat\bfq^{\exper,r}\, .                                                                 \label{EQ64}
\end{equation}
- In Eq.~\eqref{EQ61}, for all $\ell\in\{1,\ldots,\nu_\ar\}$, we have
\begin{equation}
\widetilde\bfw_0^\ell = [G_0]\, \widehat\bfw^\ell \in\RR^{\nu_w}\, ,                                              \label{EQ65-1}
\end{equation}
\begin{equation}
\zeta_0^\ell(\bfu^j)  = \exp \left\{ - \frac{1} {2s_\ar^2} \Vert\, [\curL_0]\,(\bfu^j - \widehat\bfw^\ell)\Vert^2 \right\}\in\RR^+ \, ,\label{EQ65-2}
\end{equation}
in which the upper triangular $(\nu_w\times\nu_w)$ real matrix $[\curL_0]$ follows from the Cholesky factorization,
$[G_0] = [\curL_0]^T\,[\curL_0]$.\\

\noindent - In Eq.~\eqref{EQ62}, for all $\ell\in\{1,\ldots,\nu_\ar\}$, we have
\begin{equation}
\widetilde\bfw_1^\ell = [G_w]\,\widehat\bfw^\ell +  [G_{qw}]^T\,\widehat\bfq^\ell \in \RR^{n_w}\, ,                                   \label{EQ65bis}
\end{equation}
and for $r\in\{1,\ldots,n_r\}$,
\begin{equation}
\zeta_1^{r\ell}(\bfu^j)  = \exp \left\{ - \frac{1} {2s_\ar^2} ( p_0^{r\ell} + p_1^{r\ell}(\bfu^j) )\right\}\in\RR^{+}\, ,              \label{EQ66}
\end{equation}
in which the positive real number $p_0^{r\ell}$ is expressed as
\begin{equation}
p_0^{r\ell}  =  \Vert\, [\curL_q]\,(\widehat\bfq^{\exper,r} - \widehat\bfq^\ell)\Vert^2 \, ,                                           \label{EQ65}
\end{equation}
with $[\curL_q]$ the real upper triangular $(\nu_q\times\nu_q)$ Cholesky factor,
$[G_q] = [\curL_q]^T\,[\curL_q]$, and where
\begin{equation}
p_1^{r\ell}(\bfu^j)  =  \Vert\, [\curL_w]\,(\bfu^j - \widehat\bfw^\ell)\Vert^2
      + 2< \![G_{qw}]^T(\widehat\bfq^{\exper,r} - \widehat\bfq^\ell\} \, , \bfu^j -\widehat\bfw^\ell \! >\, ,                                \label{EQ67}
\end{equation}
where the upper triangular $(\nu_w\times\nu_w)$ real matrix $[\curL_w]$ is obtained from the Cholesky factorization,
$[G_w] = [\curL_w]^T\,[\curL_w]$.\\

\noindent - The numerical experiments that have been carried out have shown that, for the computation of
$\zeta_1^{r\ell}(\bfu^j)$ defined by Eq.~\eqref{EQ66}, the term in the exponential must be computed before exponentiation in order to avoid underflow and numerical noise.
\subsection{Transformation of the adapted nonlinear ISDE for the generation of $\widehat\bfW^\post$}
\label{Section8.3}
In this section, we construct the transformation introduced at the beginning of Section~\ref{Section8}, we deduce the nonlinear ISDE from the adapted nonlinear ISDE, we verify that the construction proposed satisfies the criteria, and finally, we present the numerical aspects for the computation.
\subsubsection{Construction of the transformation}
\label{Section8.3.1}
The covariance matrix of $\widehat\bfW^\post$ can neither explicitly be calculated using pdf $p_{\widehat\bfW}^\post$ nor estimated by computational statistics. Indeed such an estimation would require an integration on $\RR^{n_w}$, integration that has to be estimated using the Monte Carlo method \cite{Givens2013,Soize2017b} with respect to a pdf for which a large number of realizations would be drawn (for instance using the $\nu_\ar$ additional realizations of the prior model of $\bfW$, or using a uniform pdf). The use of $p_{\widehat\bfW}^\post$ is not possible since the generator is under construction and as of yet unknown. Even in relatively high dimension, this approach can be prohibitive since the normalization constant $c_0$ of $p_{\widehat\bfW}^\post$ is unknown and has to be numerically estimated. Consequently, an approximation of the covariance matrix of $\widehat\bfW^\post$ is performed using a linearization of mapping $[u]\mapsto [L([u])]$ around an approximation, denoted by $\underline{\widehat\bfw}^\exper$, of the mean value $E\{\widehat\bfW^\post\}$ of $\widehat\bfW^\post$ that is also unknown (because only experimental realizations $\{\widehat\bfq^{\exper,r}, r =1,\ldots , n_r\}$ of $\widehat\bfQ$ are assumed to be available. Let us assume that $\underline{\widehat\bfw}^\exper$ is a given vector in $\RR^{n_w}$, which will be identified in Section~\ref{Section8.3.5}. For given vector $\bfu$ in $\RR^{\nu_w}$, let $\bfL(\bfu)=(L_1(\bfu),\ldots,L_{\nu_w}(\bfu))$ be the vector in $\RR^{\nu_w}$ such that, for $k=1,\ldots, \nu_w$ and for $j=1,\ldots, N_s$, the component $L_k(\bfu^j)$ of $\bfL(\bfu^j)$ is
\begin{equation}
L_k(\bfu^j) = [L([u])]_{kj} \quad, \quad [u]=[\bfu^1\ldots\bfu^{N_s}] \, ,                                                            \label{EQ68}
\end{equation}
in which  $[L([u])]$ is defined by Eq.~\eqref{EQ52}. Matrix $[K]\in\MM_{\nu_w}^+$
introduced in Section~\ref{Section8.2}, for which the factorization $[K]=[A]\,[A]^T$ is given by Eq.~\eqref{EQ43}, is then defined for all $k$ and $k'$ in $\{1,\ldots,\nu_w\}$, as
\begin{equation}
[K]_{kk'} = \left \{ \frac{\partial^2\curV(\bfu)}{\partial u_k\partial u_{k'}}\right \}_{\bfu = \underline{\widehat\bfw}^\exper}
       = - \left\{ \frac{\partial}{\partial u_{k'}} L_k(\bfu)\right \}_{\bfu = \underline{\widehat\bfw}^\exper}\, .                       \label{EQ69}
\end{equation}
From this definition, matrix $[K]$ is symmetric, but there is not necessarily positive definite for any value of
$\underline{\widehat\bfw}^\exper$, because function $\bfu\mapsto\curV(\bfu)$ defined by Eq.~\eqref{EQ51}, is not, \textit{ a priori}, convex on $\RR^{\nu_w}$ for the non-Gaussian pdf $p_{\widehat\bfW}^\post$. However, it can be assumed that $\bfu\mapsto\curV(\bfu)$ is locally convex in the neighborhood of $\bfu = \underline{\widehat\bfw}^\exper$ if this vector is correctly estimated (see Section~\ref{Section8.3.5}). Therefore, $[K]$ will be in $\MM_{\nu_w}^+$ (this property will effectively be checked numerically in the algorithm (see Section~\ref{Section8.3.4})).
The first-order Taylor development of  $\bfu\mapsto\bfL(\bfu)$ around $\bfu = \underline{\widehat\bfw}^\exper$ is written as
\begin{equation}
\bfL(\bfu) = \bfL( \underline{\widehat\bfw}^\exper)  +
    [\bfnabla_\bfu \bfL(\bfu)]_{\bfu = \underline{\widehat\bfw}^\exper } (\bfu - \underline{\widehat\bfw}^\exper)
    + o(\Vert \bfu - \underline{\widehat\bfw}^\exper\Vert) \, ,                                                 \nonumber
\end{equation}
which yields the following linearized expression,
\begin{equation}
\bfL^\linear(\bfu) = \bfL( \underline{\widehat\bfw}^\exper) -
    [K]\, (\bfu - \underline{\widehat\bfw}^\exper) \quad , \quad
    [K] = -[\bfnabla_\bfu \bfL(\bfu)]_{\bfu = \underline{\widehat\bfw}^\exper }\, .                                                 \label{EQ70}
\end{equation}
Let $\bfu_T\in\RR^{\nu_w}$ be the solution of the equation $\bfL^\linear(\bfu_T) =\bfzero$, which is such that
\begin{equation}
\bfu_T = \underline{\widehat\bfw}^\exper + [K]^{-1}\, \bfL(\underline{\widehat\bfw}^\exper) \, .                                   \label{EQ71}
\end{equation}
The transformation of stochastic process $\{[\bfU(t)] ,[\bfV(t)], t\in\RR^+\}$, involved in the adapted nonlinear ISDE defined by Eqs.~\eqref{EQ44} to \eqref{EQ46}, is written as
\begin{align}
[\bfU(t)] = &  {[u_T]} + {[A]^{-T}} \, [\bfS(t)] \, ,                                                                               \label{EQ69bis} \\
[\bfV(t)] = &  [A]\, [\bfR(t)]\, ,                                                                                                  \label{EQ70bis}
\end{align}
in which $[u_T] = [\bfu_T \ldots \bfu_T]\in \MM_{\nu_w,N_s}$, where $[A]$ is defined by Eq.~\eqref{EQ43}, and where
$\{([\bfS(t)] ,[\bfR(t)]), t\in\RR^+\}$ is the new stochastic process with values in $\MM_{\nu_w,N_s}\times \MM_{\nu_w,N_s}$,
such that
\begin{align}
[\bfS(t)] = &  {[A]^{T}} \, ([\bfU(t)] -{[u_T]}) \, ,                                                                               \label{EQ71bis} \\
[\bfR(t)] = &  [A]^{-1}\, [\bfV(t)]\, .                                                                                             \label{EQ72}
\end{align}
\subsubsection{Nonlinear ISDE for stochastic process $\{([\bfS(t)],[\bfR(t)]),t\in\RR^+\}$}
\label{Section8.3.2}
Substituting Eqs.~\eqref{EQ69bis} and \eqref{EQ70bis} into Eqs.~\eqref{EQ44} and \eqref{EQ45}, using   Eqs.~\eqref{EQ71bis} and \eqref{EQ72}
for transforming the initial conditions defined by Eq.~\eqref{EQ46}, simple algebraic manipulations yield,  for $t >0$, the following nonlinear ISDE,
\begin{align}
d[\bfS(t)] & = [\bfR(t)] \, dt  \, ,                                                                                              \label{EQ73}\\
d[\bfR(t)] & = [\widetilde L([\bfS(t)])]\, dt -\frac{1}{2} \,f_0^\post [\bfR(t)]\, dt + \sqrt{f_0^\post} \, d[\bfW^\wien(t)] \, ,  \label{EQ74}
\end{align}
with the almost-sure initial condition at $t=0$,
\begin{equation}
[\bfS(0)]  = [s_0] \quad , \quad  [\bfR(0)]  = [r_0]  \, .                                                                    \label{EQ75}
\end{equation}
The matrices $[s_0]$ and  $[r_0]$ in $\MM_{\nu_w,N_s}$ are given by
\begin{align}
[s_0] = &  {[A]^{T}} \, ([\widehat w_0] -{[u_T]}) \, ,                                                                           \label{EQ76} \\
[r_0] = &  [A]^{-1}\, [\widehat v_0]\, .                                                                                          \label{EQ77}
\end{align}
The mapping $[s]\mapsto [\widetilde L([s])]$ from $\MM_{\nu_w,N_s}$ into $\MM_{\nu_w,N_s}$ is written as
\begin{equation}
[\widetilde L([s])] = [A]^{-1}\, [L({[u_T]} + {[A]^{-T}} \, [s])] \, .                                                                   \label{EQ78}
\end{equation}
\subsubsection{Verifying that the linearized ISDE is well adapted for stochastic process $\{([\bfS(t)],[\bfR(t)]),t\in\RR^+\}$}
\label{Section8.3.3}
Using Eqs.~\eqref{EQ70} and \eqref{EQ71}, the linearization $[\widetilde L^\linear([s])]$ of $[\widetilde L([s])]$
defined by Eq.~\eqref{EQ78} is such that $[\widetilde L^\linear([s])] = -[s]$. From Eqs.~\eqref{EQ73} and \eqref{EQ74}, it
can be deduced that the linearized ISDE is written as
\begin{align}
d[\bfS^\linear(t)] & = [\bfR^\linear(t)] \, dt  \, ,                                                                                       \nonumber \\
d[\bfR^\linear(t)] & = -[\bfS^\linear(t)])]\, dt -\frac{1}{2} \,f_0^\post [\bfR^\linear(t)]\, dt + \sqrt{f_0^\post} \, d[\bfW^\wien(t)] \, .  \nonumber
\end{align}
Let us write $[\bfS^\linear(t)] = [\bfS^{\linear,1}(t) \ldots \bfS^{\linear,N_s}(t)]$ whose columns are statistically dependent  for $t >0$ due to the coupling by the initial conditions defined by Eqs.~\eqref{EQ76} and \eqref{EQ77}. Nevertheless, for the asymptotic solution ($t\rightarrow +\infty$), denoted as $\{([\bfS^\linear (t_\st)],$ $[\bfR^\linear (t_\st)]), t_\st \in\RR^+\}$, these are statistically independent and it is  known (see for instance Page 241 of \cite{Soize1994}) that each column $\{\bfS^{\linear,j}(t_\st), t_\st\in\RR^+\}$ of $\{[\bfS^{\linear}(t_\st)], t_\st\in\RR^+\}$ is a Gaussian, stationary, centered stochastic process whose covariance matrix $[C^\linear] = E\{\bfS^{\linear,j}(t_\st) \, \bfS^{\linear,j}(t_\st)^T\}$ is independent of $j$ and such that $[C^\linear]= [I_{\nu_w}]$. This result shows that the nonlinear ISDE defined by
Eqs.~\eqref{EQ73} to \eqref{EQ78} is well adapted to the covariance matrix of the asymptotic stochastic process $\{[\bfS(t_\st)],t_\st\in\RR^+\}$ and therefore, to the covariance of $[\bfW^\post]$ {\textit via}
the transformation defined by Eqs.~\eqref{EQ69bis} and \eqref{EQ70bis}. It can be seen that $f_0^\post=4$ corresponds to the critical damping rate of the linearized dynamical system.
\subsubsection{Numerical aspects for computing matrix $[K]$}
\label{Section8.3.4}
Assuming that $\underline{\widehat\bfw}^\exper$ is given in $\RR^{\nu_w}$, we must calculate $[K]$ defined by Eq.~\eqref{EQ69}.
Although the algebraic calculation can actually be carried out, the
corresponding numerical implementation carries a numerical cost that
is greater than the direct numerical calculation of the gradient. This
last approach will thus be pursued. Let $\{\Delta t _\alpha, \alpha =1,2,\ldots\}$ be a decreasing sequence of positive real numbers that goes to zero. Let $[K_\alpha]$ be the sequence of matrices in $\MM_{\nu_w}$ such that
\begin{equation}
[K_\alpha]_{kk'}  =  -\frac{1}{\Delta t_\alpha}\, ( L_k(\underline{\widehat\bfw}^\exper+\frac{\Delta t_\alpha}{2} \bfe^{k'})
                                  - L_k(\underline{\widehat\bfw}^\exper -\frac{\Delta t_\alpha}{2} \bfe^{k'})) \, ,                              \label{EQ79}
\end{equation}
in which $\{\bfe^1,\ldots , \bfe^{\nu_w}\}$ is the canonical basis of $\RR^{\nu_w}$. Matrix $[K]$ is then defined as $[K_{\alpha^\popt}]$
in which, for all $\alpha > \alpha^\opt$, the symmetrization error is sufficiently small for the Frobenius norm and all the eigenvalues are strictly positive.
\subsubsection{Estimating $\underline{\widehat\bfw}^\exper$}
\label{Section8.3.5}
The algorithm proposed for estimating
$\underline{\widehat\bfw}^\exper$ is based on a predictor-corrector
method. The predictor is based on the fact that the size $\nu_\ar$ of
the learned dataset, $\DD_{\nu_\ar}$, constructed in
Section~\ref{Section4} using the PLoM, can be chosen as large as required.\\

\noindent \textit{(i)- Predictor}. The predictor of $\underline{\widehat\bfw}^\exper$ is the vector $\underline{\widehat\bfw}^{\exper,\predictor} \in \RR^{\nu_w}$ such that
\begin{equation}
\underline{\widehat\bfw}^{\exper,\predictor}  =  E\{ \widehat\bfW \,\vert \, \widehat\bfQ = \underline{\widehat\bfq}^\exper\}\, ,           \label{EQ80}
\end{equation}
in which $\underline{\widehat\bfq}^\exper = (1/n_r )\sum_{r=1}^{n_r} {\widehat\bfq}^{\exper ,r}$ is the vector in $\RR^{\nu_q}$  where
${\widehat\bfq}^{\exper ,r}$ is defined by Eq.~\eqref{EQ13}. Therefore, we have
\begin{equation}
\underline{\widehat\bfw}^{\exper,\predictor}  =
   \frac { \int_{\RR^{\nu_w}} \widehat\bfw \, p_{\widehat\bfQ,\widehat\bfW}(\underline{\widehat\bfq}^\exper ,\widehat\bfw) \, d\widehat\bfw}
         { \int_{\RR^{\nu_w}} p_{\widehat\bfQ,\widehat\bfW}(\underline{\widehat\bfq}^\exper ,\widehat\bfw) \, d\widehat\bfw}\, ,            \label{EQ81}
\end{equation}
where $p_{\widehat\bfQ,\widehat\bfW}$ is defined by Eqs.~\eqref{EQ38}
and \eqref{EQ39}. The calculation of the integrals in Eq.~\eqref{EQ81}
can be explicitly evaluated yielding,
\begin{equation}
\underline{\widehat\bfw}^{\exper ,\predictor} =
   \frac { \sum_{\ell=1}^{\nu_\ar}  \widetilde\bfw_2^\ell\, \zeta_2^\ell}
         { \sum_{\ell=1}^{\nu_\ar}  \zeta_2^\ell}\, ,                                                                              \label{EQ82}
\end{equation}
in which $\widetilde\bfw_2^\ell$ belongs to $\RR^{\nu_w}$ and is written as
\begin{equation}
\widetilde\bfw_2^\ell = \widehat\bfw^\ell - [G_w]^{-1}\, [G_{qw}]^T\,(\underline{\widehat\bfq}^\exper -\widehat\bfq^\ell) \, ,      \label{EQ83}
\end{equation}
and where $\zeta_2^\ell$ is positive and such that
\begin{equation}
\zeta_2^\ell = \exp\left\{ -\frac{1}{2s_\ar^2} < \![G_1]\,(\underline{\widehat\bfq}^\exper -\widehat\bfq^\ell)\, , \underline{\widehat\bfq}^\exper -\widehat\bfq^\ell\! >\right\}  \, .                                                                                                             \label{EQ84}
\end{equation}
The matrix $[G_1]$  is the Schur complement defined by
\begin{equation}
[G_1] = [G_q] - [G_{qw}]\, [G_w]^{-1}\, [G_{qw}]^T \in \MM_{\nu_q}^+\, .                                                                \label{EQ85}
\end{equation}

\noindent \textit{(ii)- Corrector}. We introduce the maximum log-likelihood of the posterior model,
\begin{equation}
\underline{\widehat\bfw}^\exper  =  \max_{\widehat\bfw\in\RR^{\nu_w}} \log p_{\widehat\bfW}^\post(\widehat\bfw)\, .                    \label{EQ86}
\end{equation}
Using Eq.~\eqref{EQ50} for $p_{\widehat\bfW}^\post$ with Eqs.~\eqref{EQ38} and \eqref{EQ39} for $p_{\widehat\bfQ,\widehat\bfW}$, and Eq.~\eqref{EQ41}
for $p_{\widehat\bfW}$, the non convex optimization problem can be rewritten as
\begin{equation}
\underline{\widehat\bfw}^\exper  =  \max_{\widehat\bfw\in\RR^{\nu_w}} J(\widehat\bfw)\, ,                                                \label{EQ87}
\end{equation}
in which $J(\widehat\bfw)$ is written as
\begin{equation}
J(\widehat\bfw) = (1-n_r)\,\log\left\{\sum_{\ell=1}^{\nu_\arp} \zeta_0^\ell(\widehat\bfw)\right\} + \sum_{r=1}^{n_r} \log\left\{\sum_{\ell=1}^{\nu_\arp}\zeta_1^{r\ell}(\widehat\bfw) \right\}\, ,                                                                         \label{EQ88}
\end{equation}
where $\zeta_0^\ell(\widehat\bfw)$ is defined by Eq.~\eqref{EQ65-2} and where $\zeta_1^{r\ell}(\widehat\bfw)$ is defined by Eq.~\eqref{EQ66}.
The corrector of $\underline{\widehat\bfw}^{\exper,\predictor}$ is the
vector $\underline{\widehat\bfw}^\exper$ that is constructed  by
solving the nonconvex optimization problem defined by Eq.~\eqref{EQ87}
using the interior-point algorithm for which the initial point is
chosen as $\widehat\bfw_0 =
\underline{\widehat\bfw}^{\exper,\predictor}$ that is computed using
Eq.~\eqref{EQ82}.
%
\subsection{Projection of the nonlinear ISDE for stochastic process $\{([\bfS(t)],[\bfR(t)]),t\in\RR^+\}$ using a diffusion-maps basis}
\label{Section8.4}
In order to avoid a possible scattering of the generated realizations constructed by solving the nonlinear ISDE defined by Eqs.~\eqref{EQ73}
to \eqref{EQ78} and in order to preserve a possible concentration of the measure
$P_{\widehat\bfW^\post}(d\widehat\bfw)=p_{\widehat\bfW^\post}(\widehat\bfw)\, d\widehat\bfw$ on $\RR^{\nu_w}$, a projection of the ISDE is carried out using the diffusion-maps basis following the methodology of the PLoM that is summarized in Appendix~A. We then obtain a reduced-order nonlinear ISDE.
\subsubsection{Construction of the diffusion-maps basis for the posterior model}
\label{Section8.4.1}
The diffusion-maps basis is represented by the matrix
\begin{equation}
[g_s] = [\bfg_s^1 \ldots \bfg_s^{m_\post}]\in\MM_{N_s,m_\post} \quad \hbox{with}\quad
1 < m_\post \leq N_s \leq \nu_\ar \, ,                                                                                                 \label{EQ89}
\end{equation}
which is constructed using the set of independent realizations $\{\bfs^j,j=1,\ldots, N_s\}$ that result from the transformation defined by Eq.~\eqref{EQ71bis} of the set $\{\widehat\bfw^{\nu_\ar-j+1}, j =1,\ldots, N_s\}$ extracted form the learned dataset ${\widehat D}_{\nu_\ar}$ (see Eq.~\eqref{EQ19}). We then have,
\begin{equation}
\bfs^j = [A]^T\, ( \widehat\bfw^{\nu_\ar-j+1}  -\bfu_T)\in\RR^{\nu_w} \quad , \quad j=1,\ldots , N_s\, ,                             \label{EQ90}
\end{equation}
in which $\bfu_T \in \RR^{\nu_w}$ is defined by Eq.~\eqref{EQ71}. The construction of this diffusion-maps basis is summarized in Appendix~E.
\subsubsection{Reduced-order nonlinear ISDE}
\label{Section8.4.2}
The reduced-order nonlinear ISDE is  obtained by projection on diffusion-maps basis $[g_s]\in\MM_{N_s,m_\post}$ of the nonlinear ISDE relative to
the $(\MM_{\nu_w,N_s}\! \times \!\MM_{\nu_w,N_s})$-valued
stochastic process $\{([\bfS(t)],$ $[\bfR(t)]),t\in\RR^+\}$ defined by Eqs.~\eqref{EQ73} to \eqref{EQ78}. We then introduced the $(\MM_{\nu_w,m_\ppost} \!\times\! \MM_{\nu_w,m_\ppost})$-valued stochastic process $\{([\bfcurZ(t)],$ $[\bfcurY(t)]),t\in\RR^+\}$ such that,
\begin{equation}
[\bfS(t)] = [\bfcurZ(t)] \, [g_s]^T \quad ,\quad  [\bfR(t)] = [\bfcurY(t)] \, [g_s]^T \quad , \quad t\geq 0\, .                     \label{EQ91}
\end{equation}
Stochastic process $\{([\bfcurZ(t)],[\bfcurY(t)]),t\in\RR^+\}$ is then the solution of the reduced-order nonlinear ISDE (obtained by projection) such that,
for all $t >0$,
\begin{align}
d[\bfcurZ(t)] & = [\bfcurY(t)] \, dt  \, ,                                                                                             \label{EQ92}\\
d[\bfcurY(t)] & = [\widetilde \curL([\bfcurZ(t)])]\, dt -\frac{1}{2} \,f_0^\post [\bfcurY(t)]\, dt
                                         + \sqrt{f_0^\post} \, d[\bfcurW^\wien(t)] \, ,                                                \label{EQ93}
\end{align}
with the almost-sure initial condition at $t=0$,
\begin{equation}
[\bfcurZ(0)]  = [z_0] \quad , \quad  [\bfcurY(0)]  = [y_0]  \, .                                                                        \label{EQ94}
\end{equation}
The matrices $[z_0]$ and  $[y_0]$ in $\MM_{\nu_w,m_\ppost}$ are written as
\begin{equation}
[z_0] = [s_0]\, [a_s] \quad , \quad  [y_0] = [r_0]\, [a_s]   \, ,                                                                       \label{EQ95}
\end{equation}
in which matrices $[s_0]$ and  $[r_0]$ in $\MM_{\nu_w,N_s}$ are defined by Eqs.~\eqref{EQ76} and \eqref{EQ77}, and where $[a_s]$ is the matrix such that
\begin{equation}
 [a_s] =[g_s]\, ([g_s]^T\,[g_s])^{-1} \in \MM_{N_s,m_\ppost} \, .                                                                       \label{EQ96}
\end{equation}
In Eq.~\eqref{EQ93}, $[\widetilde \curL([\bfcurZ(t)])]$ is such that
\begin{equation}
[\widetilde \curL([\bfcurZ(t)])] = [\widetilde L([\bfZ(t)]\, [g_s]^T)] \, [a_s]\, ,                                                     \label{EQ97}
\end{equation}
in which $[\widetilde L([s])]$ is defined by Eq.~\eqref{EQ78}, and where
\begin{equation}
[\bfcurW^\wien(t)] = [\bfW^\wien(t)] \, [a_s]\, .                                                                                       \label{EQ98}
\end{equation}
\subsection{Construction of realizations of $\widehat\bfW^\post$}
\label{Section8.5}
The independent realizations $\{\widehat\bfw^{\post,\ell},\ell=1,\ldots,\nu_\post\}$ (used in Eq.~\eqref{EQ19bis}) of $\widehat\bfW^\post$ whose pdf is $p_{\widehat\bfW}^\post$ defined by Eq.~\eqref{EQ50}, are constructed using the discretization of the reduced-order ISDE defined by Eqs.~\eqref{EQ92} to \eqref{EQ94}. The number, $\nu_\post$, of realizations is reparameterized as
\begin{equation}
\nu_\post = n_\MC^\post \times N_s \, ,                                                                                                  \label{EQ99}
\end{equation}
in which $n_\MC^\post$ is a given integer.
Let $[\bfW^\wien(\cdot;\theta)]$ with $\theta\in\Theta$ be a realization of the Wiener stochastic process $[\bfW^\wien]$ defined in Section~\ref{Section8.2}-(ii). Let $\{([\bfcurZ(t;\theta)],[\bfcurY(t;\theta)]),t\in\RR^+\}$ be one realization of the $(\MM_{\nu_w,m_\ppost} \!\times\! \MM_{\nu_w,m_\ppost})$-valued stochastic process $\{([\bfcurZ(t)],[\bfcurY(t)]),t\in\RR^+\}$, which is computed by solving Eqs.~\eqref{EQ92} to \eqref{EQ94} with the St\"{o}rmer-Verlet scheme detailed in Appendix~F for which the sampling step is $\Delta t$. Let $\ell_0^\post$ be the integer such that, for $t \geq \ell_0^\post\, \Delta t$, the solution of Eqs.~\eqref{EQ92} to \eqref{EQ94} is asymptotic to the stationary solution. Therefore, the independent realizations of $\widehat\bfW^\post$ can be generated as follows. Let $M_0^\post$ be a given positive integer. Using Eqs.~\eqref{EQ69bis} and \eqref{EQ91}, for $n=1,\ldots , n_\MC^\post$ and
for $t_{\ell'} =\ell'\,\Delta t$ with $\ell' = \ell_0^\post + n\, M_0^\post$, we have, for $j=1,\ldots, N_s$ and for $k=1,\ldots, \nu_w$,
\begin{equation}
\widehat\bfw_k^{\post,\ell}  = [u^{\ell'}]_{kj} \quad , \quad \ell = j +(\ell'-1)\, N_s \, ,                                         \label{EQ100}
\end{equation}
\begin{equation}
[u^{\ell'}]  = [u_T] + [A]^{-T}\, [s^{\ell'}]  \quad , \quad [s^{\ell'}] = [\bfcurZ(t_{\ell'};\theta)]\, [g_s]^T\, .                 \label{EQ101}
\end{equation}
In this method of generation, only one realization $\theta$ is used and $M_0^\post$ is chosen sufficiently large in order that
$[\bfcurZ(t_{\ell'})]$ and $[\bfcurZ(t_{\ell'+M_0^\ppost})]$ be two random matrices that are approximatively independent.
\section{Choice of the value of the regularization parameter $\varepsilon$}
\label{Section9}
The regularization introduced in Section~\ref{Section7.2} was aimed to
facilitate the nonparametric statistical estimation of the pdf $p_{\widehat\bfX}$ of random
variable $\widehat\bfX=(\widehat\bfQ,\widehat\bfW)$ with values in
$\RR^{\nu} = \RR^{\nu_q}\times\RR^{\nu_w}$, using the multidimensional
Gaussian kernel-density estimation (see Eq.~\eqref{EQ35}).
As already explained in that section, the proposed regularization
depends on the parameter $\varepsilon$ and on the criterion for
selecting $\nu_1$. Consequently, the posterior pdf
$p_{\widehat\bfW}^\post$ of $\widehat\bfW$, which is directly deduced
from $p_{\widehat\bfX}$, depends on $\varepsilon$. There is no
\textit{prior} information constraining $\varepsilon$
chosen, which is typical when regularizations are introduced. Further,
a mathematical exploration of Eq.~\eqref{EQ35.3}, aimed at deducing
such constraints, seems intractable.

It may seem possible to compute an optimal value of $\varepsilon$ by
minimizing the $L^1$-norm of the difference between the pdf of
$\widehat\bfQ^\post$ and the pdf of $\widehat\bfQ^{\exper}$.
This is not possible if $n_r$ is small, because the quality of the
nonparametric estimation of the pdf of $\widehat\bfQ^{\exper}$ would
not be sufficiently good. If $n_r$ is sufficiently large for obtaining
a good estimation of $\widehat\bfQ^{\exper}$, then an algorithm could
proceed as follows. For a given value of $\varepsilon$, the first
stage would consist of using the algorithm presented in this paper for
estimating the pdf of $\widehat\bfW^\post$  and then generating the
$\nu_\post$ realizations
$\{\widehat\bfw^{\post,\ell},\ell=1,\ldots,\nu_\post\}$ of
$\widehat\bfW^\post$ (which depend on $\varepsilon$). The second stage
would consist of estimating the  pdf of $\widehat\bfQ^\post$ using the
conditional pdf of $\widehat\bfW^\post$ given $\widehat\bfQ =
\widehat\bfq$, which has to be evaluated for the $\nu_\post$
realizations $\{\widehat\bfw^{\post,\ell},\ell=1,\ldots,\nu_\post\}$
(and not, using the conditional pdf of $\widehat\bfW$ given
$\widehat\bfQ=\widehat\bfq$, which should then be evaluated for the
experimental realizations of $\widehat\bfW$, which are not
available). Note that the complexity of such an approach would be
similar to the one that we have used for estimating the pdf of
$\widehat\bfW^\post$. The pdf of $\widehat\bfQ^\post$ that would be
estimated would depend on $\varepsilon$. The third stage would consist
of solving an optimization problem with respect to $\varepsilon$ for
which the objective function would be the $L^1$-norm of the
error. Such a non convex optimization problem would be relatively
tricky and numerically expensive.

Consequently, we propose to fix the value of $\varepsilon$ to an
"average value" that has been estimated by numerical experiments. In
order to estimate this "average" value, the following method has been
used. Let $p_{\WW}^\post$ be the posterior pdf of $\WW$ that is
estimated with the $\nu_\post$ realizations
$\{\ww^{\post,\ell},\ell=1,\ldots,\nu_\post\}$ that are deduced from
$\{\widehat\bfw^{\post,\ell},\ell=1,\ldots,\nu_\post\}$ computed in
Section~\ref{Section8.5}, using Eqs.~\eqref{EQ19bis} and
\eqref{EQ20}.\\

The methodology used for validating the range of the values of $\varepsilon$ consists in estimating an optimal value of $\varepsilon$, which minimizes a "distance" between the pdf $p_{\WW_k}^\post$ of component $\WW_k^\post$, for $k=1,\ldots, n_w$ (which depends on $\varepsilon)$, and an experimental reference,
$p_{\WW_k}^\exper$, that is assumed to be known for the applications used for the validation. Obviously, in the framework of the Bayesian inference, the family of $\{p_{\WW_k}^\exper , k =1,\ldots , n_w\}$ are unknown and consequently, cannot be used for estimating $\varepsilon$ \textit{a priori}. It is recalled that only $n_r$ experimental realizations $\{\qq^{\exper,r}, r=1,\ldots,n_r\}$ of $\QQ$ are available and that the corresponding experimental realizations  $\{\ww^{\exper,r}, r=1,\ldots,n_r\}$ of $\WW$ are not available. We thus introduce the error function,
$\varepsilon \mapsto \underline\OVL (\varepsilon)$, defined by
\begin{equation}
\underline\OVL (\varepsilon)  =
\frac{1}{n_w} \sum_{k=1}^{n_w} \frac{\int_\RR \vert p_{\WW_k}^\post(w)- p_{\WW_k}^\exper(w)\vert\, dw}
                                    {\int_\RR p_{\WW_k}^\exper(w)\, dw}\, .                                                                  \label{EQ102}
\end{equation}
Let $\bfp = (p_1,\ldots,p_{n_w})$ be a function in the space $L^1(\RR,\RR^{n_w})$ equipped with the $L^1$-norm,
\begin{equation}
\Vert\bfp\Vert_{L^1} = \int_\RR \vert \bfp(w) \Vert_1 \, dw = \int_\RR \sum_{j=1}^{n_w} \vert p_j(w)\vert \, dw \, . \nonumber
\end{equation}
Introducing the functions $\bfp^\post = (p^\post_{\WW_1},\ldots,p^\post_{\WW_{n_w}})$ and $\bfp^\exper = (p^\exper_{\WW_1},\ldots,p^\exper_{\WW_{n_w}})$
that belong to $L^1(\RR,\RR^{n_w})$, it can be seen that
\begin{equation}
\frac{\Vert\bfp^\post - \bfp^\exper\Vert_{L^1}}{\Vert \bfp^\exper\Vert_{L^1}}  \leq n_w\, \underline\OVL (\varepsilon)\, , \nonumber
\end{equation}
because, for $j=1,\ldots, n_w$, we have  $A_j/(a_1+\ldots + a_{n_w}) \leq A_j/a_j$ for $a_j >0$ and $A_j >0$.
All the numerical experiments that have been conducted, in particular the applications presented in Section~\ref{Section10}, show that the value $0.5$ seems an appropriate value for $\varepsilon$.
\section{Applications (AP1) and (AP2) for validating the methodology}
\label{Section10}
In this section, two applications are presented and are used for performing the validation of the methodology and algorithms presented. All the random variables are defined on probability space $(\Theta,\curT,\curP)$. These two applications will be referenced as (AP1) and (AP2) for application 1 and 2.
These two applications are relatively simple and can be easily reproduced.
\subsection{Stochastic model and simulated experiments for applications (AP1) and (AP2)}
\label{Section10.1}
\paragraph{Stochastic model for (AP1) and (AP2)}
The stochastic model used for generating the initial dataset
$\DD_{N_d} =\{ \xx_d^j = (\qq_d^j,\ww_d^j) , j=1,\ldots , N_d\}$ (see Eq.~\ref{EQO1})
relative to random variable $\XX=(\QQ,\WW)$ in which $\QQ=(\QQ_1,\ldots,\QQ_{n_q})$ and $\WW=(\WW_1,\ldots,\WW_{n_w})$,
 is written, for (AP1) and (AP2), as
\begin{equation}
\QQ = [B(\UU)]\,(\WW + V\,\bfb)\, ,                                                                                                        \nonumber
\end{equation}
in which $\UU$, $V$, and $\WW$ are independent random variables. The maximum value of $N_d$ is $200$ and $n_w=20$. We have $n_q=200$ for (AP1) and
$n_q = 20\,000$ for (AP2).
The deterministic vector $\bfb$ in $\RR^{n_w}$ is written as $\bfb = 0.2\, \bfu + 0.9$ in which all the components of $\bfu$ belongs to $]0,1[$ (generated with the Matlab script: $rng$('default'); $\bfu = rand (n_w,1)$). The real-valued random variable $V = 0.2\,\curU + 0.9$ for (AP1) and $V = 0.2\,\curU - 0.1$ for (AP2) in which $\curU$ is a uniform random variable on $[0,1]$.
The random vector $\UU=(\UU_1,\ldots ,\UU_{n_u})$ with $n_u = 6$ is written, for $\alpha=1,\ldots, n_u$, as
$\UU_\alpha =2\, u_\alpha \, \curU_\alpha +1 -u_\alpha$ in which $\curU_1,\ldots, \curU_{n_u}$ are $n_u$ independent uniform random variables on $[0,1]$ and where, $u_\alpha= 0.2(\alpha-1)/(n_u-1)$ for (AP1) and $u_\alpha= 0.7(\alpha-1)/(n_u-1)$ for (AP2).
The entries $[B(\UU)]_{kj}$ of the $(n_q\times n_w)$ random matrix are defined by
$[B(\UU)]_{kj} = \sum_{\alpha=1}^{n_u} \lambda_\alpha(\UU_\alpha)\, \varphi_k^\alpha(\UU_\alpha) \, \varphi_k^\alpha(\UU_\alpha)
\, \varphi_{j+n_q/2}^\alpha(\UU_\alpha)$. For (AP1), $\varphi_k^\alpha(\UU_\alpha) = \sin\{\alpha\, k\pi/(n_q+1)\}$  is independent of $\UU_\alpha$ (deterministic) and $\lambda_\alpha(\UU_\alpha) = 1/(\alpha\UU_\alpha)^2$. For (AP2), $\varphi_k^\alpha(\UU_\alpha) = \sin\{\alpha\,\UU_\alpha\, k\pi/(n_q+1)\}$  and $\lambda_\alpha(\UU_\alpha) = 5(1-\UU_\alpha) + 1/(\alpha\UU_\alpha)^2$.
The random vector $\WW$ is written as $\WW = \sum_{\beta=1}^{3} \sqrt{\mu_\beta} \, \bfphi^\beta\, \eta_\beta$,
in which $\mu_\beta=1/\beta^2$ and $\bfphi^\beta= (\phi_1^\beta,\ldots, \phi_{n_w}^\beta)$ with
$\phi_j^\beta= \sin \{\beta \pi j/(1+n_w) \}$. The non-Gaussian centered random vector $\bfeta = (\eta_1,\eta_2,\eta_3)$ is written as
$\bfeta = \sum_{\gamma=1}^{27}\bfy^\gamma \, \psi_{\alpha_1^{(\gamma)}}(\Xi_1)\, \psi_{\alpha_2^{(\gamma)}}(\Xi_2)$ in which $\Xi_1$ and $\Xi_2$ are independent normalized Gaussian random variables. The indices $\alpha_1^{(\gamma)}$ and $\alpha_2^{(\gamma)}$ are such that
$0 < \alpha_1^{(\gamma)} + \alpha_2^{(\gamma)} \leq 6$, and $\psi_{\alpha_1^{(\gamma)}}(\Xi_1)$ and $\psi_{\alpha_2^{(\gamma)}}(\Xi_2)$  are the polynomial Gaussian chaos. The matrix $[y] = [\bfy^1 \ldots \bfy^{27}]$ is such that $[y] \, [y]^T = [I_3]$ and is generated
using the Matlab script: $rng$('default'); $a_1$ = $randn$(27,27); [$a_2$,~] = $eig$($a_1$*$(a_1)'$); $a_2$(:\,,\,4:27) = []; [y] = $(a_2)'$.\\
\paragraph{Simulated experiments for (AP1) and (AP2)}
The experimental dataset $\DD^\exper_{n_r}$ is generated with $n_r=200$ independent experimental realizations $\{\qq^{\exper,r},r=1,\ldots n_r\}$
of $\QQ^\exper = (\QQ_1^\exper,\ldots ,\QQ_{n_q}^\exper)$. As already explained, we also generate the independent experimental realizations $\{\ww^{\exper,r},r=1,\ldots n_r\}$ of $\WW^\exper = (\WW_1^\exper,\ldots ,\WW_{n_w}^\exper)$ in order to validate the choice of the regularization parameter $\varepsilon$ (see Section~\ref{Section9}). The experimental model is written, for (AP1) and (AP2), as
\begin{equation}
\QQ^\exper = [B(\UU^\exper)]\,(\WW^\exper+ V^\exper\,\bfb)\, ,                                                                                                        \nonumber
\end{equation}
in which $\UU^\exper$, $V^\exper$, and $\WW^\exper$ are independent random variables that are also independent of $\UU$, $V$, and $\WW$.
The real-valued random variable $V^\exper = 0.2\,\curU^\exper + 0.9$ for (AP1) and $V^\exper = 0.2\,\curU^\exper - 0.1$ for (AP2) in which $\curU^\exper$ is a uniform random variable on $[0,1]$ independent of $\curU$.
The random vector $\UU^\exper=(\UU^\exper_1,\ldots ,\UU^\exper_{n_u})$ is written, for $\alpha=1,\ldots, n_u$, as $\UU^\exper_\alpha =2\, u^\exper_\alpha \, \curU^\exper_\alpha +1 -u^\exper_\alpha$ in which $\curU^\exper_1,\ldots, \curU^\exper_{n_u}$ are $n_u$ independent uniform random variables on $[0,1]$ and where, $u^\exper_\alpha= 0.3(\alpha-1)/(n_u-1)$ for (AP1) and $u_\alpha= 0.7(\alpha-1)/(n_u-1)$ for (AP2). Note that for (AP1), the coefficient is $0.3$ and not $0.2$ as in the stochastic model.
The mapping $\uu\mapsto [B(\uu)]$ is the same as the one of the stochastic model.
The random vector $\WW^\exper$ is written as $\WW^\exper = 0.2 \times \bfun + \widetilde\WW^\exper$ in which $\bfun\in\RR^{n_w}$ is the vector whose components are equal to $1$ and where $\widetilde\WW^\exper$ is an independent copy of the stochastic model of $\WW$.
\subsection{Values of the numerical parameters for the computation of (AP1) and (AP2)}
\label{Section10.3}
Table~\ref{Table1} summarizes the values of all the numerical parameters introduced in the algorithms.  Except for regularization parameter $\varepsilon$ and for the convergence learning with respect to dimension $N_d$ of initial dataset $\DD_{N_d}$ (theses two parameters  will be the subject of a particular analysis presented later), the other values of the numerical parameters have been obtained by using the existing criteria or by performing a local convergence analysis.
%
%
\begin{figure}[h!]
  \centering
  \includegraphics[width=6.5cm]{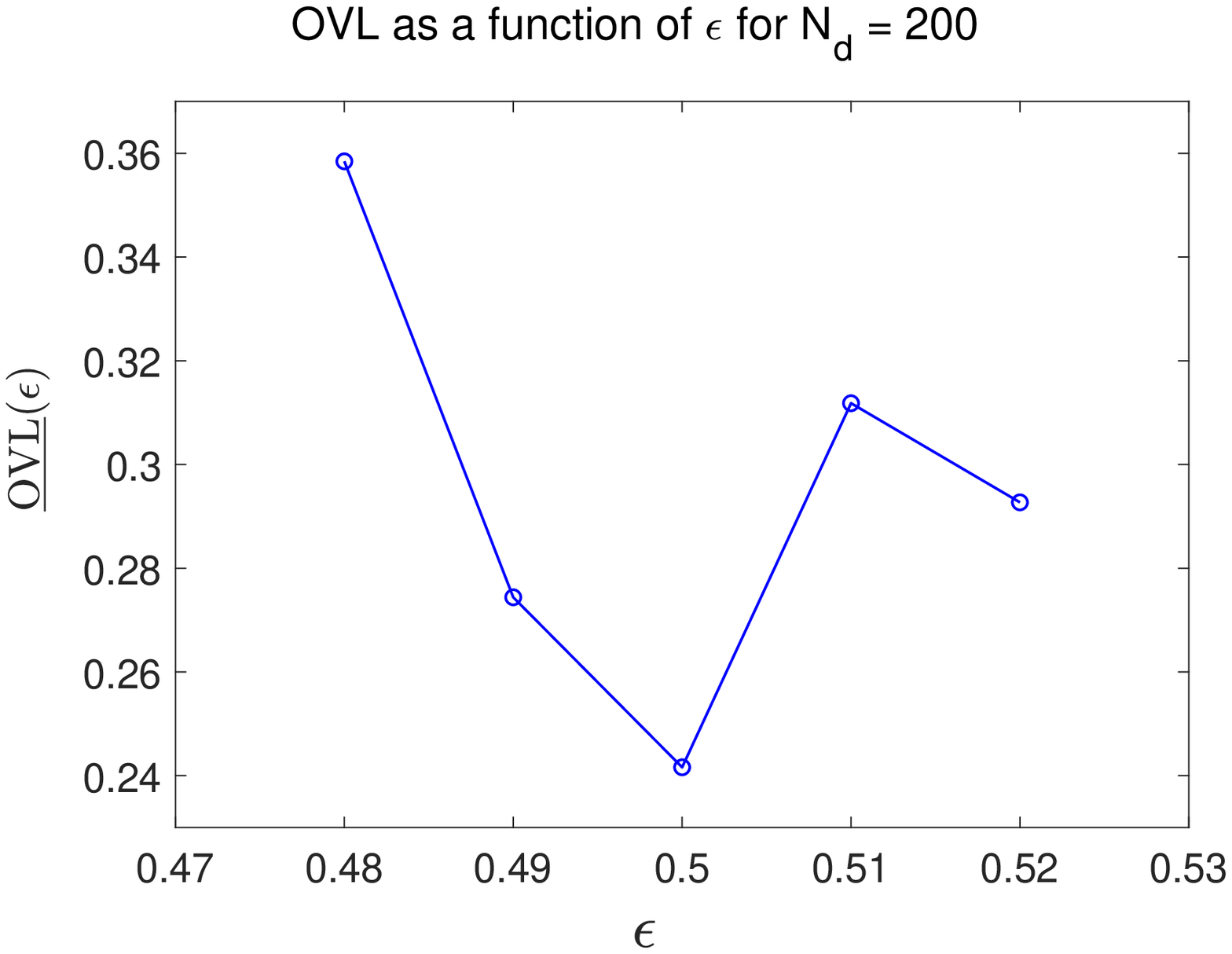} \hfill \includegraphics[width=6.5cm]{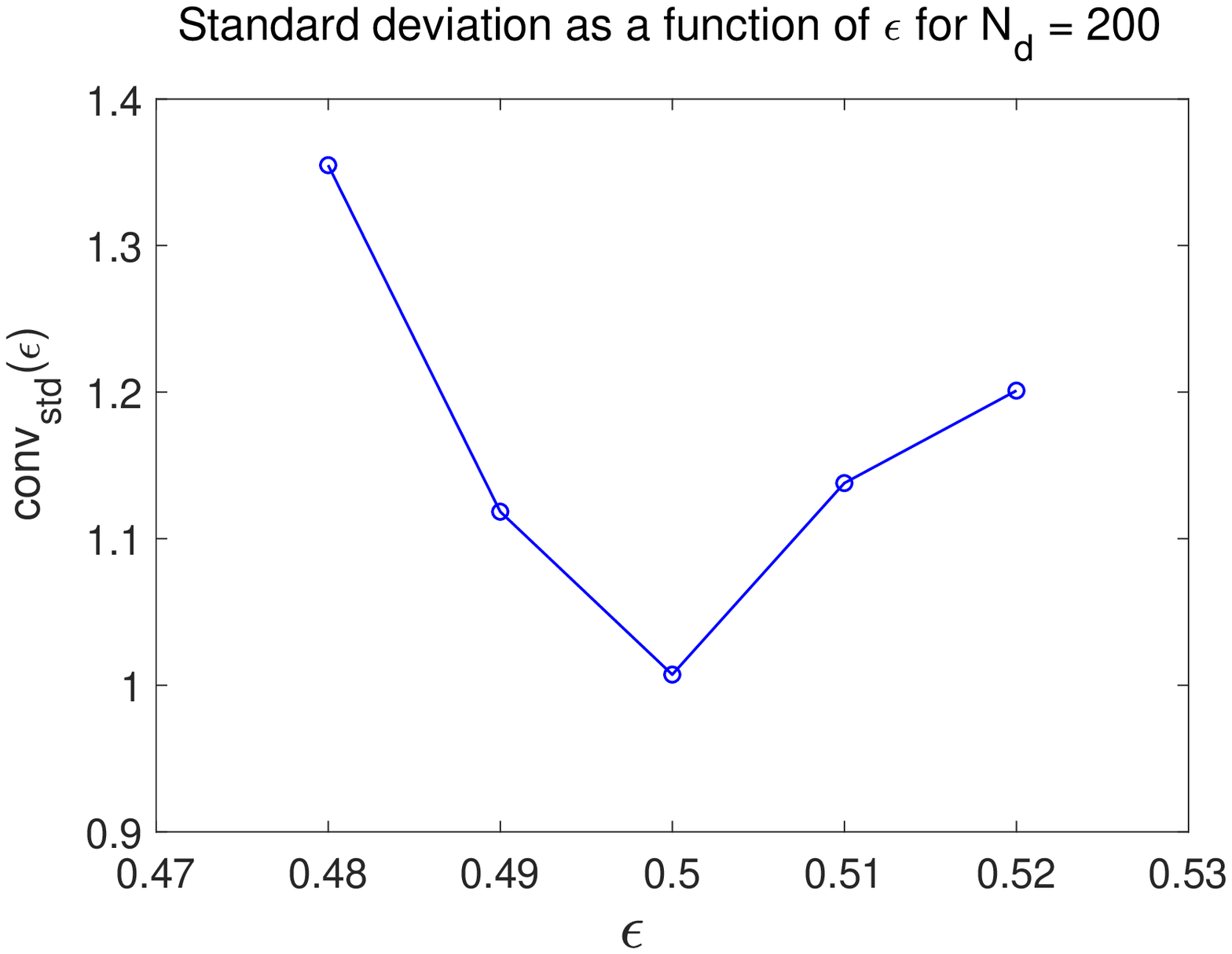}
\caption{Application AP1: validation of the choice $\varepsilon=0.5$. For $N_d = 200$, graph of
$\varepsilon\mapsto\underline\OVL (\varepsilon)$ (left) and  graph of $\varepsilon\mapsto \conv_\std (\varepsilon)$ (right).}  \label{figure1}
\end{figure}
\begin{figure}[h!]
  \centering
  \includegraphics[width=6.5cm]{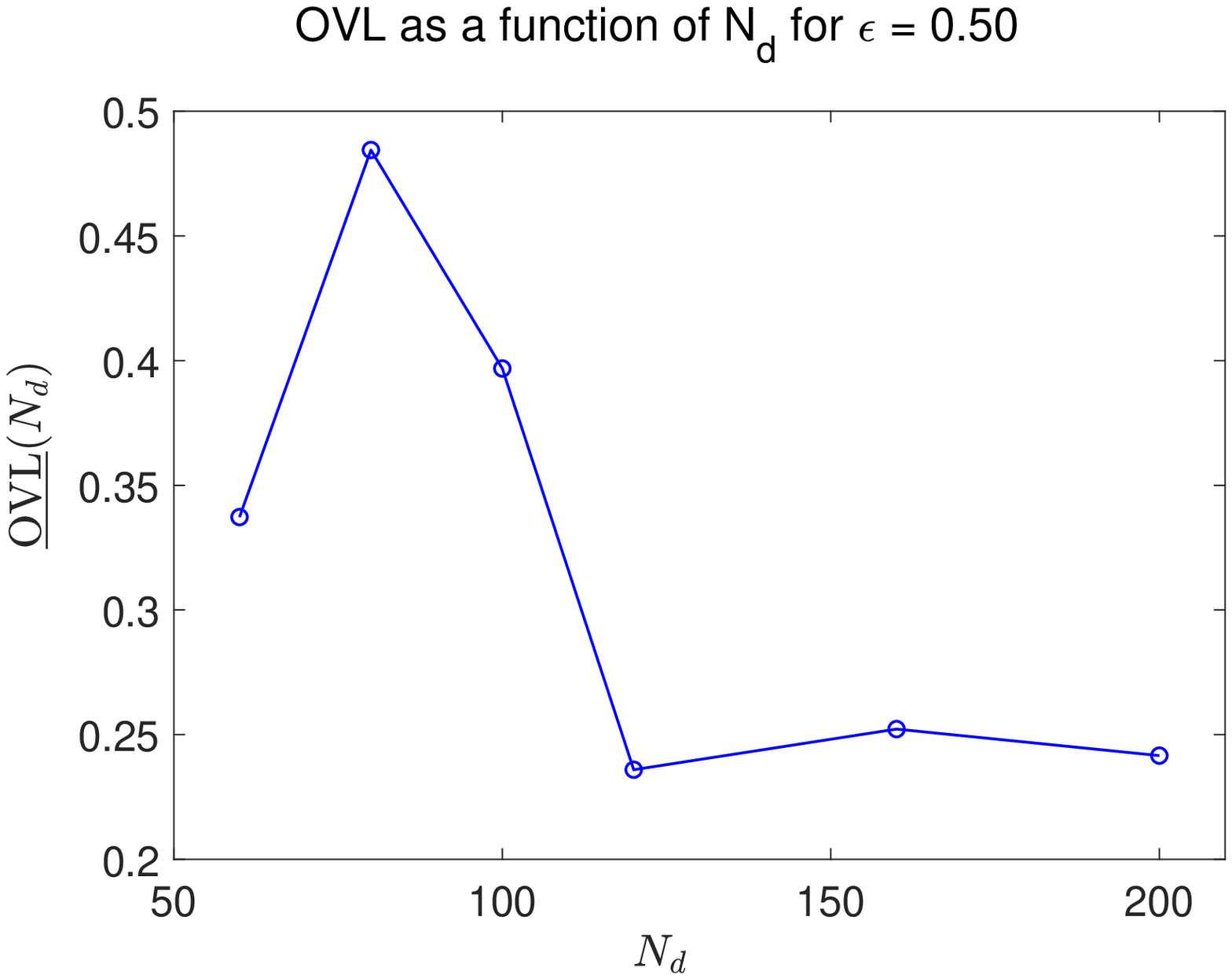} \hfill \includegraphics[width=6.5cm]{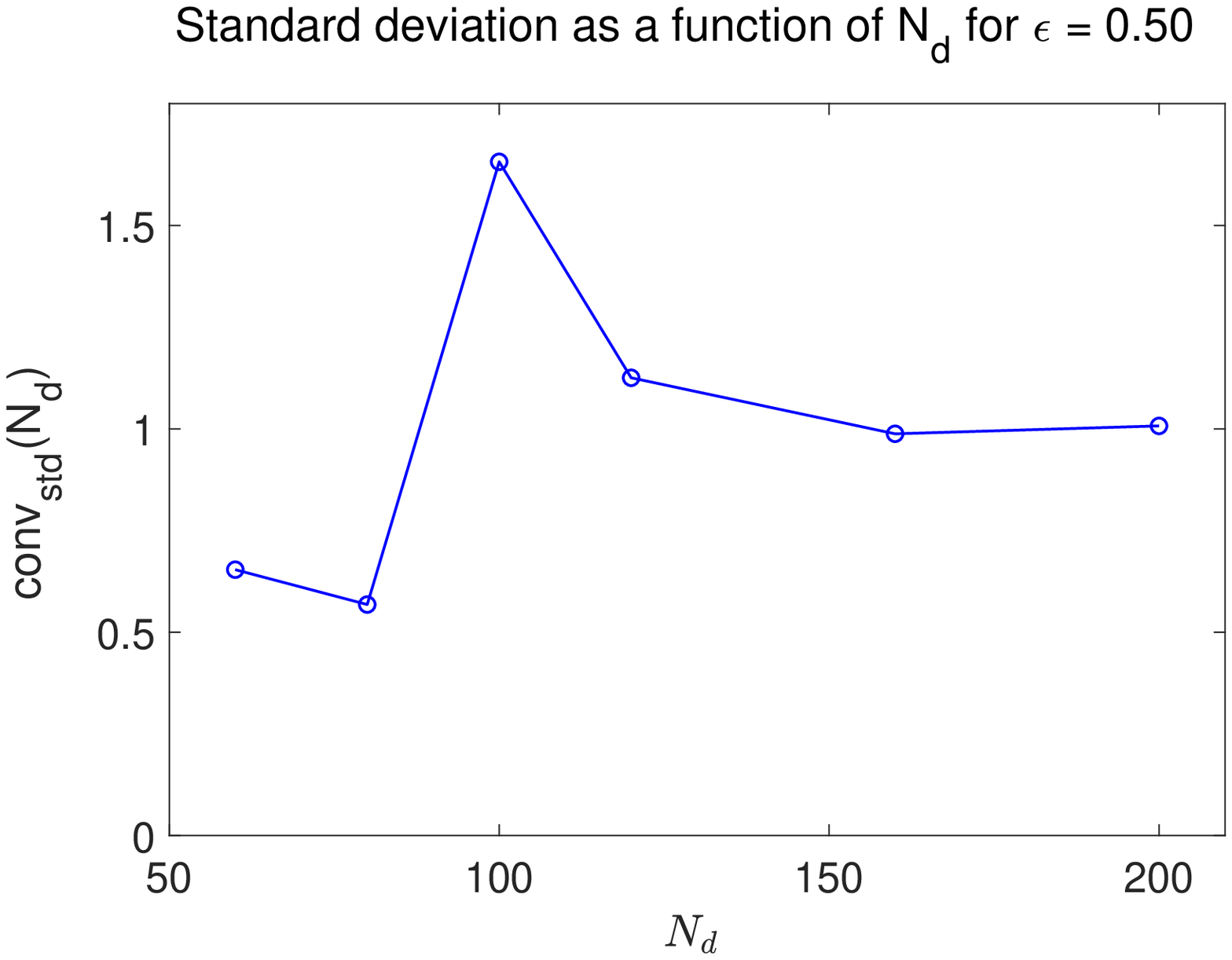}
\caption{Application AP1: convergence of the probabilistic learning with respect to $N_d$. For $\varepsilon=0.5$, graph of
$N_d\mapsto\underline\OVL (N_d)$ (left) and  graph of $N_d\mapsto \conv_\std (N_d)$ (right).}  \label{figure2}
\end{figure}
\begin{figure}[h!]
  \centering
  \includegraphics[width=6.5cm]{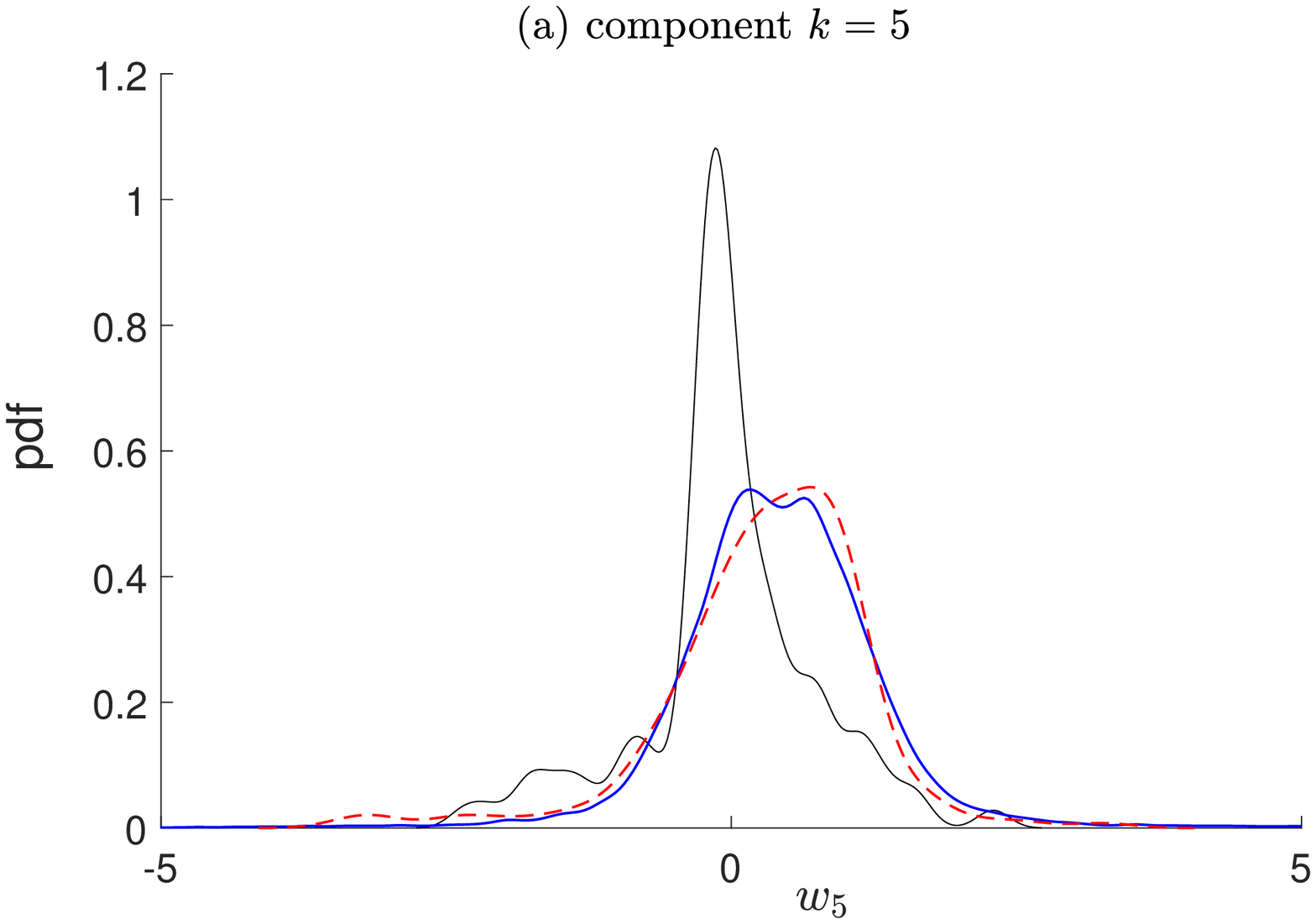} \hfill \includegraphics[width=6.5cm]{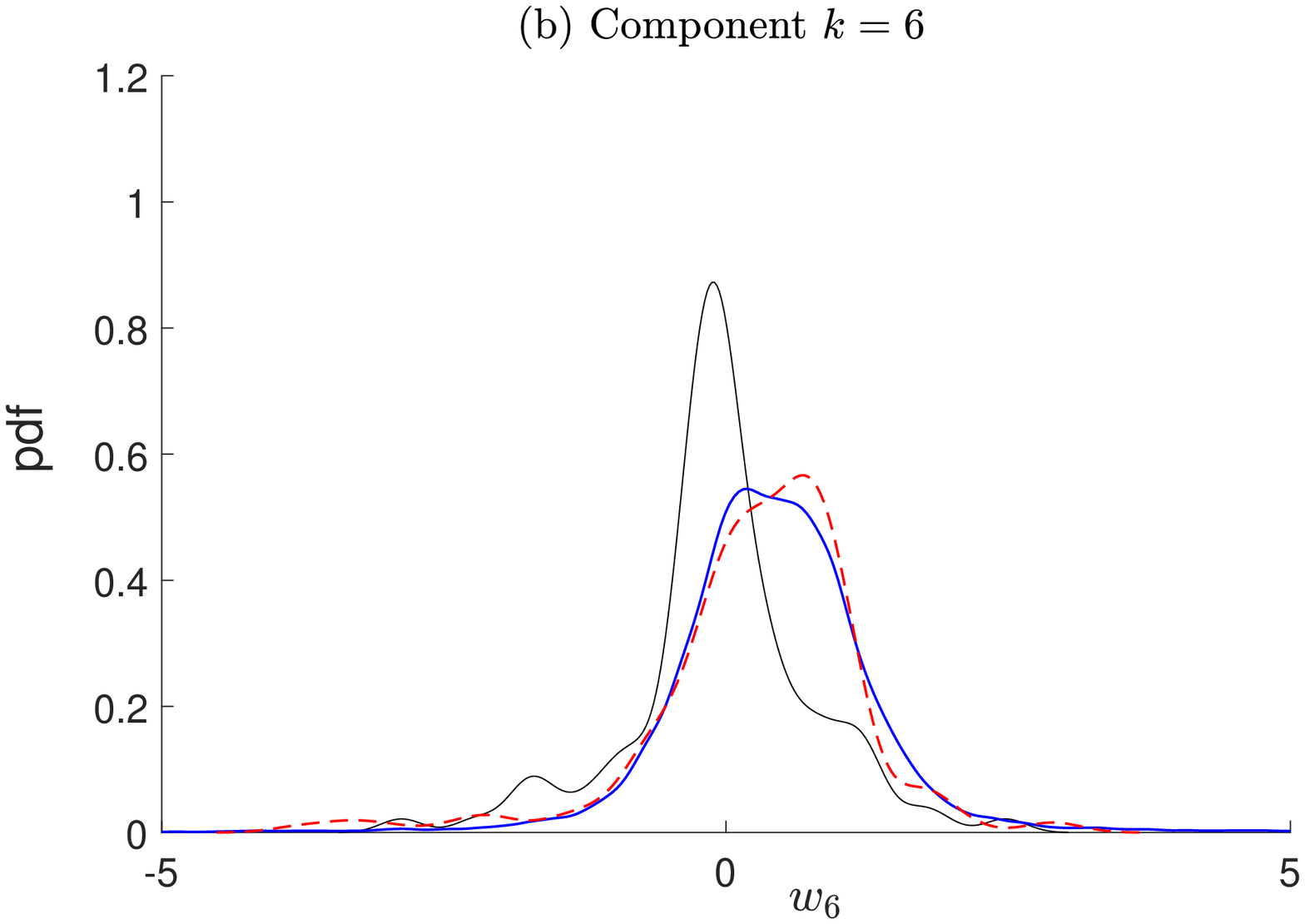}
  \includegraphics[width=6.5cm]{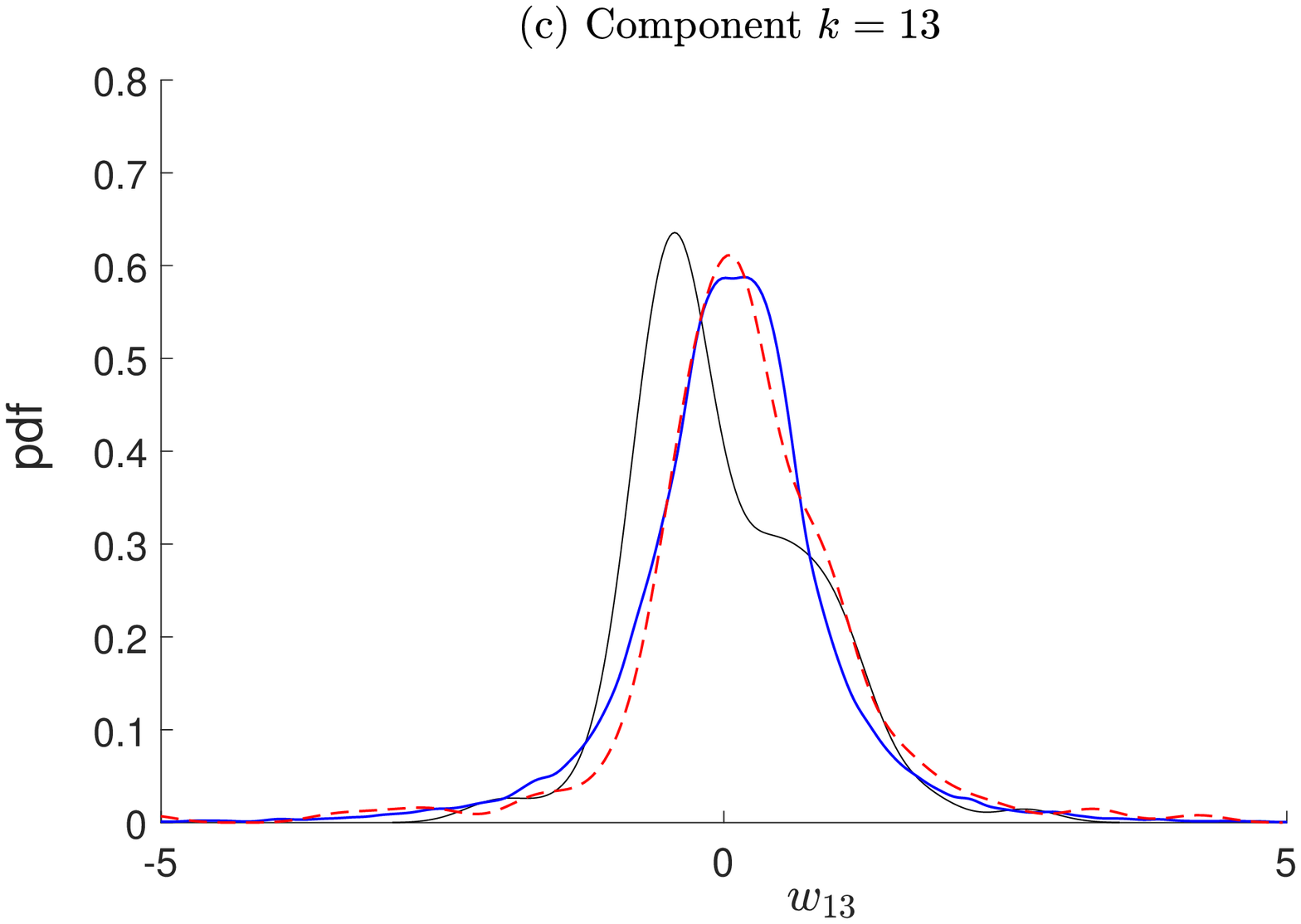} \hfill \includegraphics[width=6.5cm]{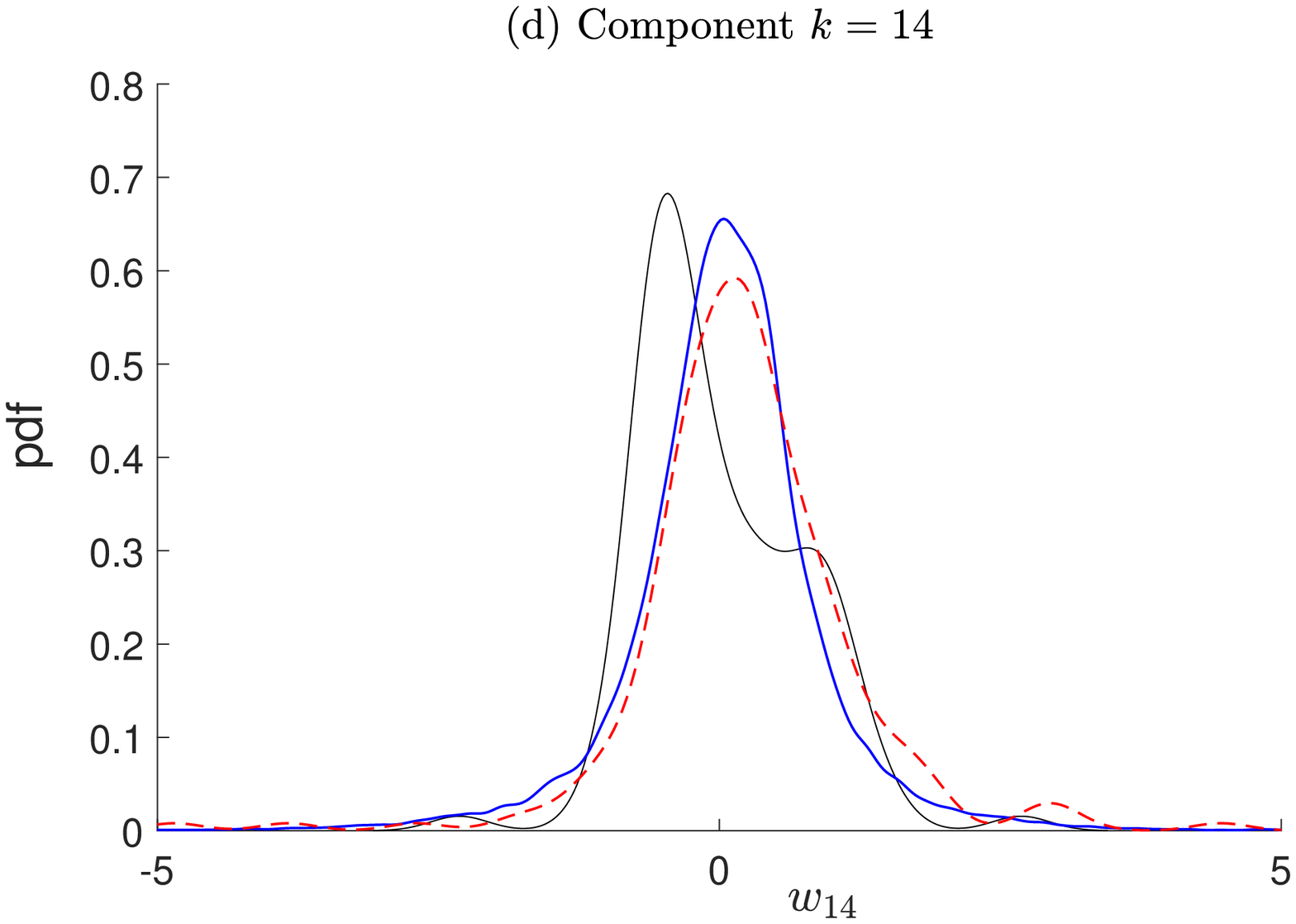}
  \caption{Application AP1: pdf $w\mapsto p_{\WW_k}^d(w)$ of $\WW_k$ estimated with the initial dataset $\DD_{N_d}$ of $N_d=200$ realizations (thin black line), pdf $w\mapsto p_{\WW_k}^\exper(w)$ of $\WW_k$ estimated with the experimental dataset $\DD_{n_r}^\exper$ of $n_r=200$ realizations (thick red dashed line), pdf $w\mapsto p_{\WW_k}^\post(w)$ of $\WW_k^\post$ estimated with $\varepsilon=0.5$, $N_d=200$, and $\nu_\post= 40\, 000$ realizations (thick blue line), for $k=5$ (a), $k=6$ (b), $k=13$ (c), and $k=14$ (d).}  \label{figure3}
\end{figure}
%
%
\begin{figure}[h!]
  \centering
  \includegraphics[width=6.5cm]{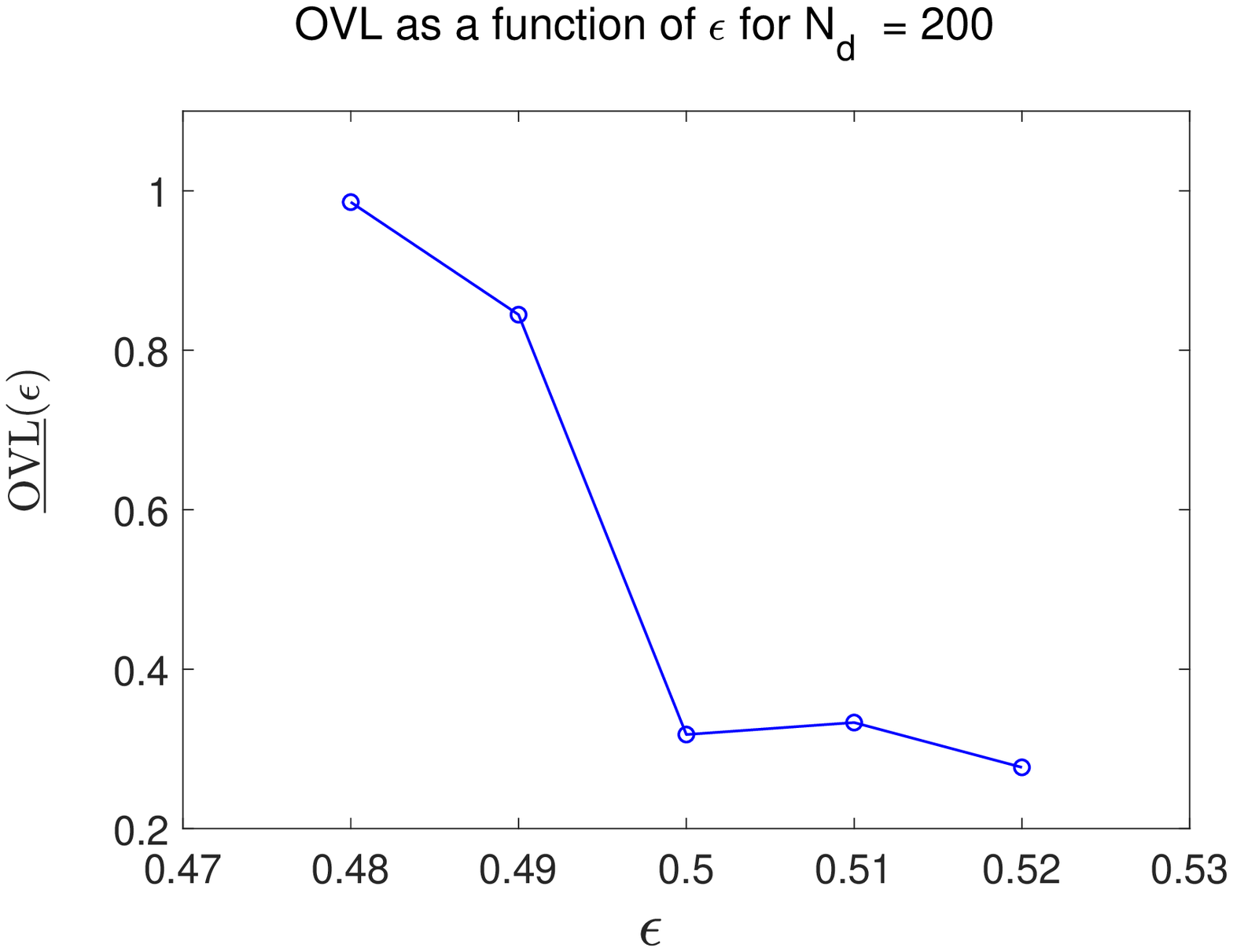} \hfill \includegraphics[width=6.5cm]{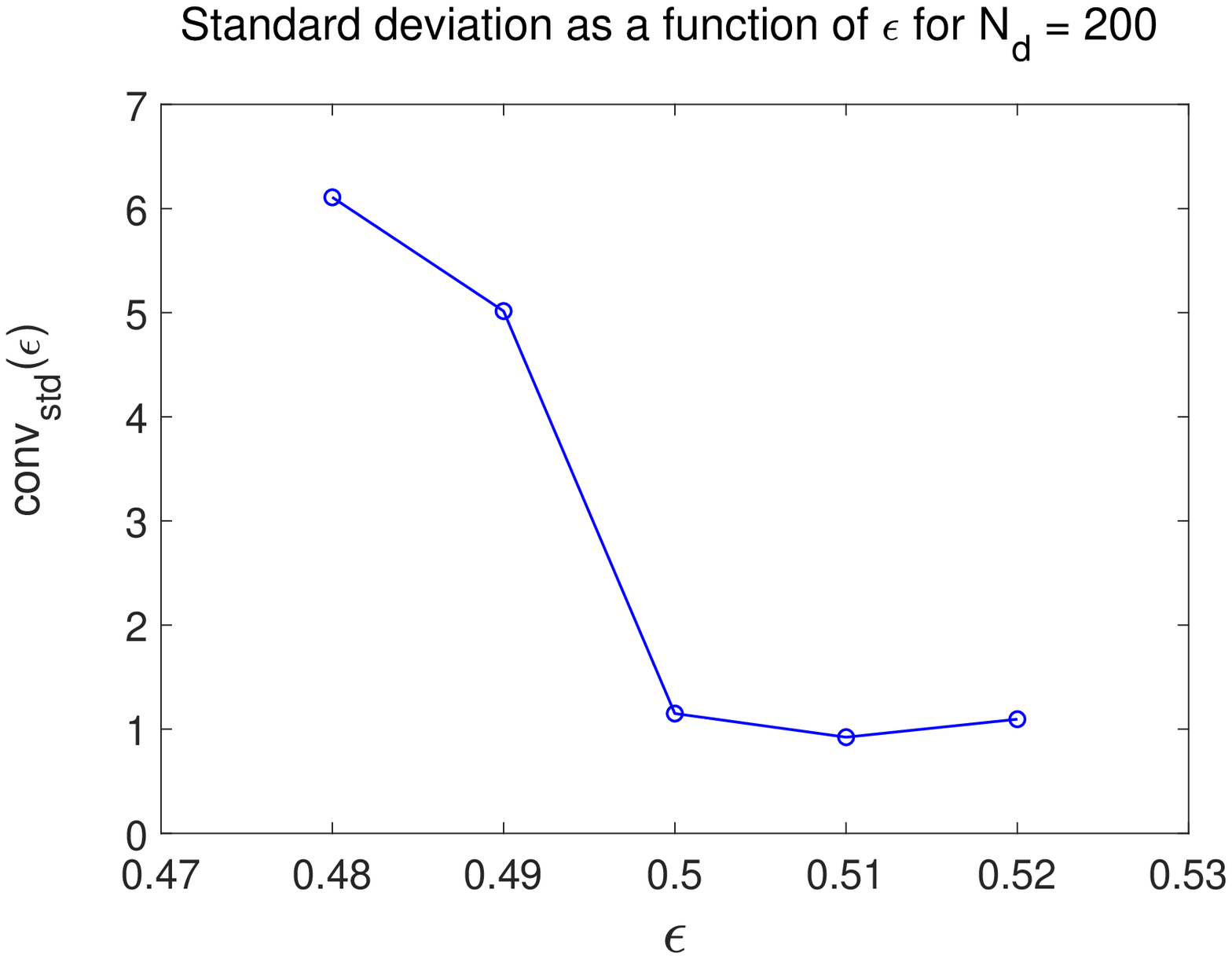}
\caption{Application AP2: validation of the choice $\varepsilon=0.5$. For $N_d = 200$, graph of
$\varepsilon\mapsto\underline\OVL (\varepsilon)$ (left) and  graph of $\varepsilon\mapsto \conv_\std (\varepsilon)$ (right).}  \label{figure4}
\end{figure}
\begin{figure}[h!]
  \centering
  \includegraphics[width=6.5cm]{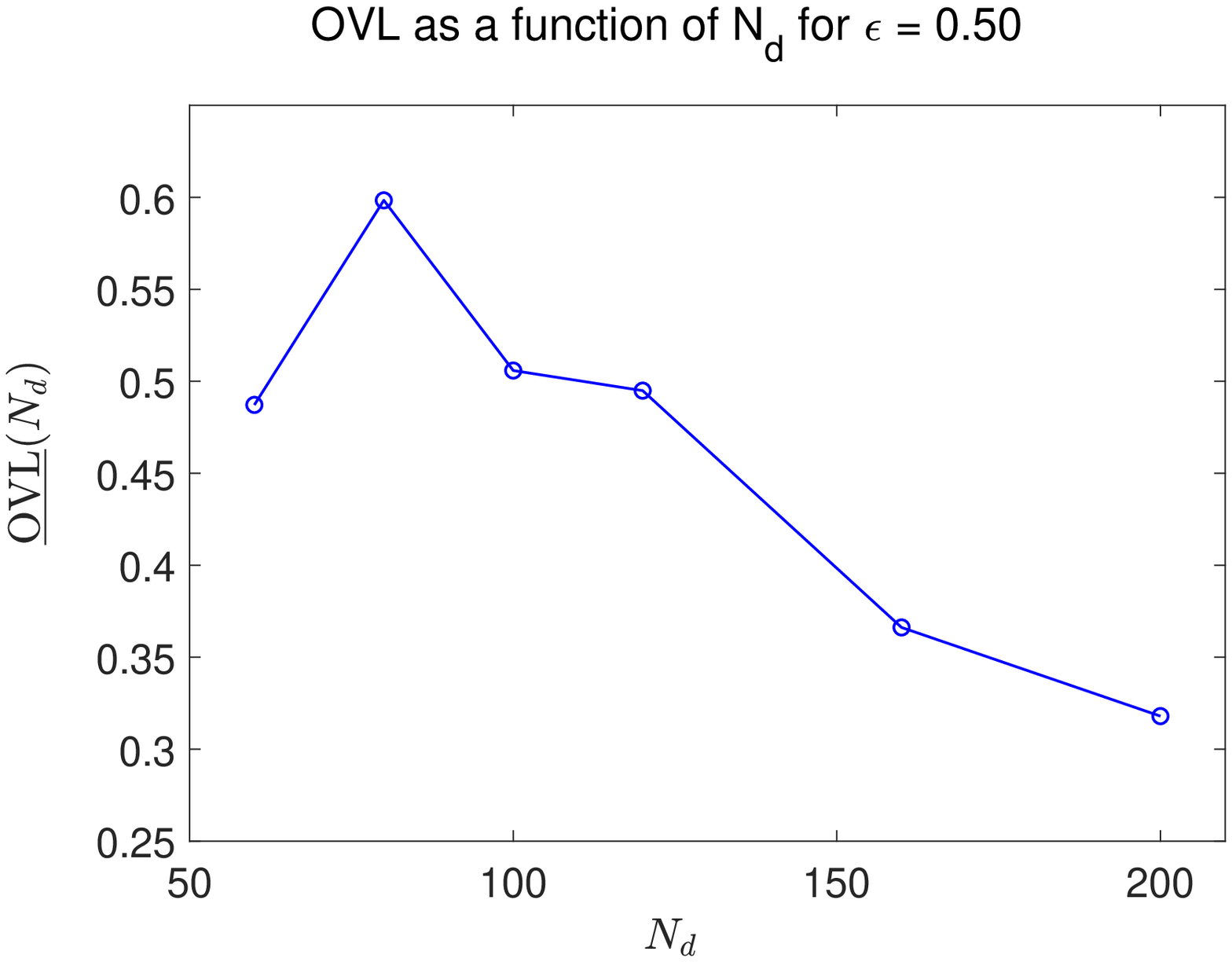} \hfill \includegraphics[width=6.5cm]{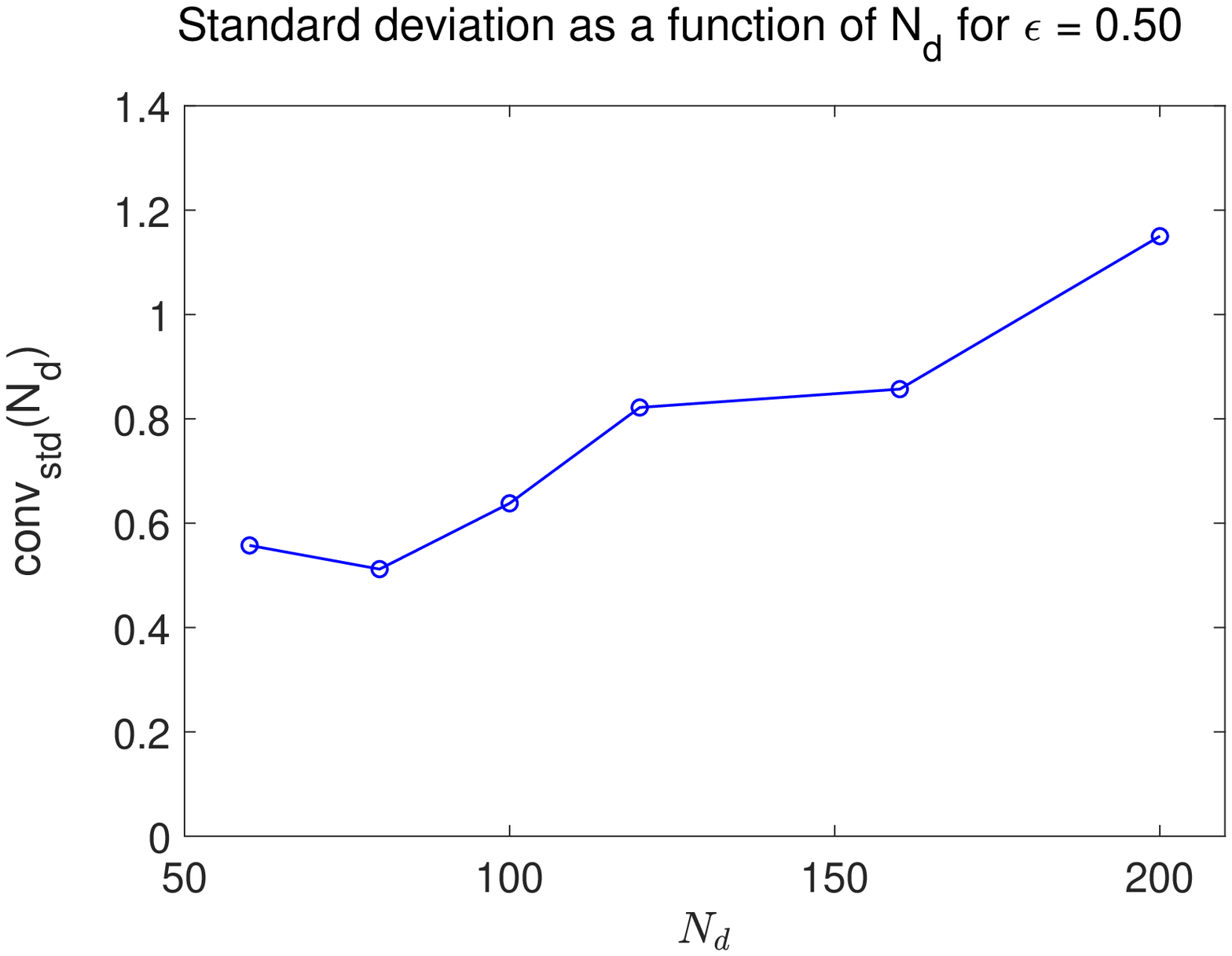}
\caption{Application AP2: convergence of the probabilistic learning with respect to $N_d$. For $\varepsilon=0.5$, graph of
$N_d\mapsto\underline\OVL (N_d)$ (left) and  graph of $N_d\mapsto \conv_\std (N_d)$ (right).}  \label{figure5}
\end{figure}
\begin{figure}[h!]
  \centering
  \includegraphics[width=6.5cm]{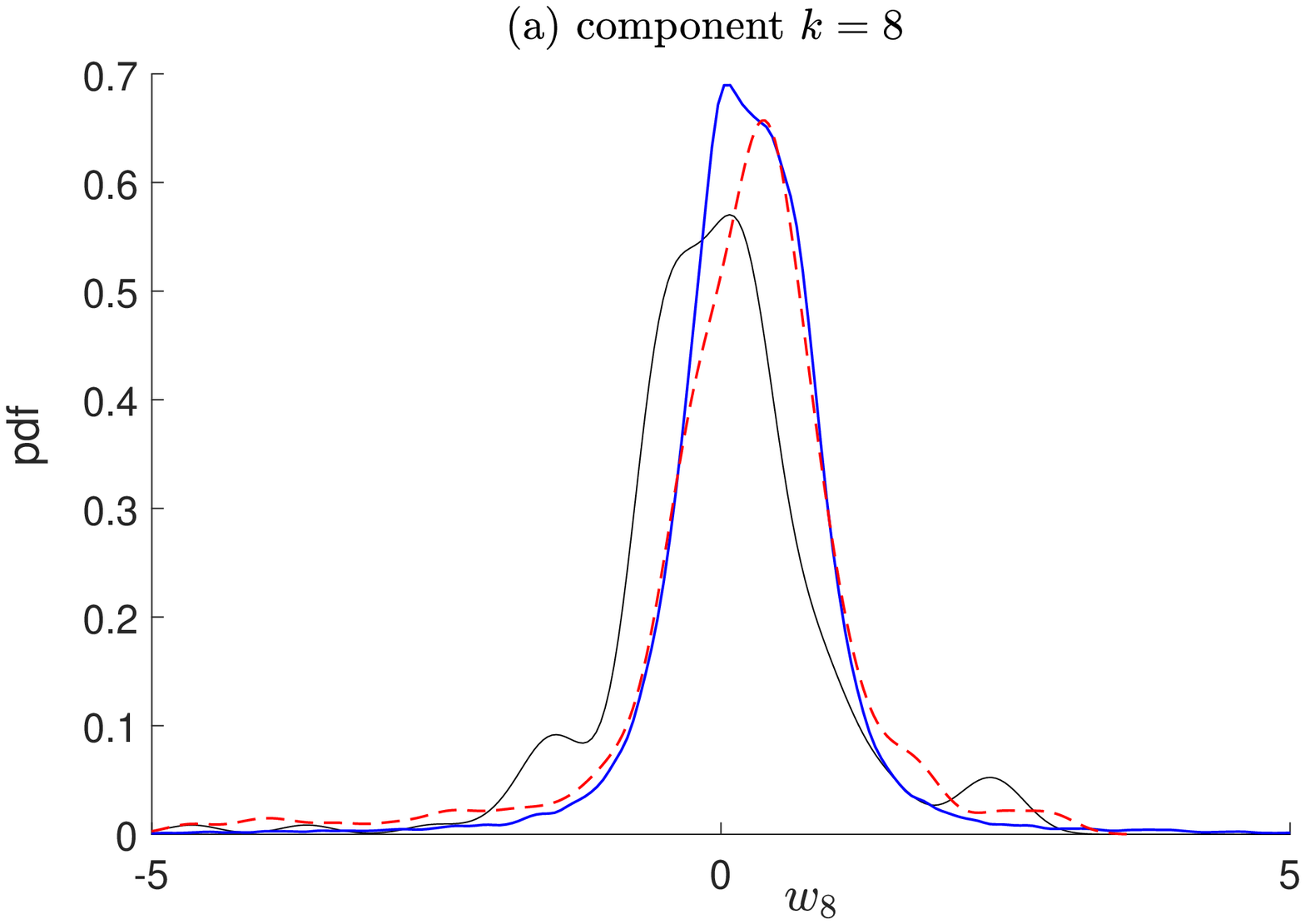} \hfill \includegraphics[width=6.5cm]{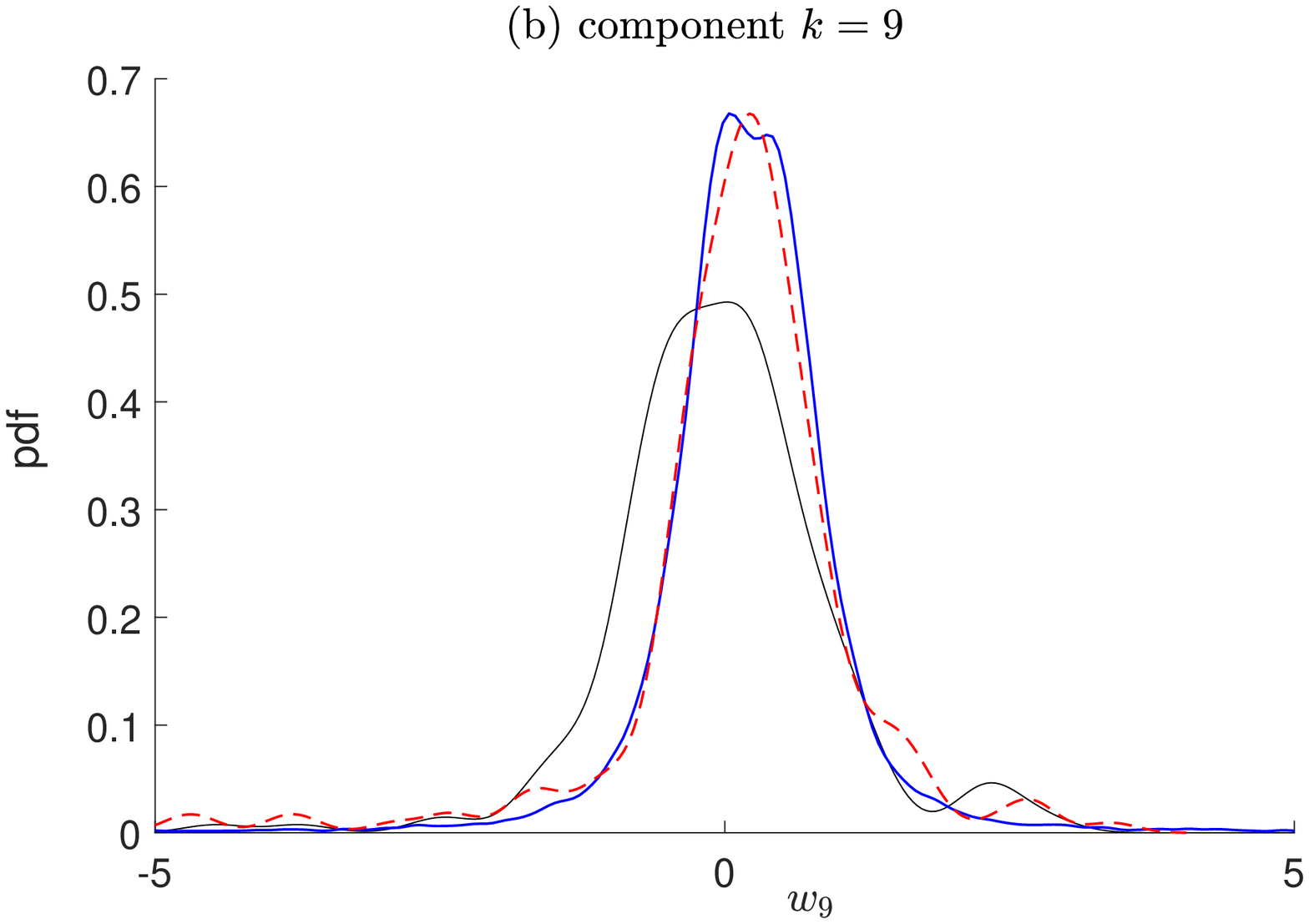}
  \includegraphics[width=6.5cm]{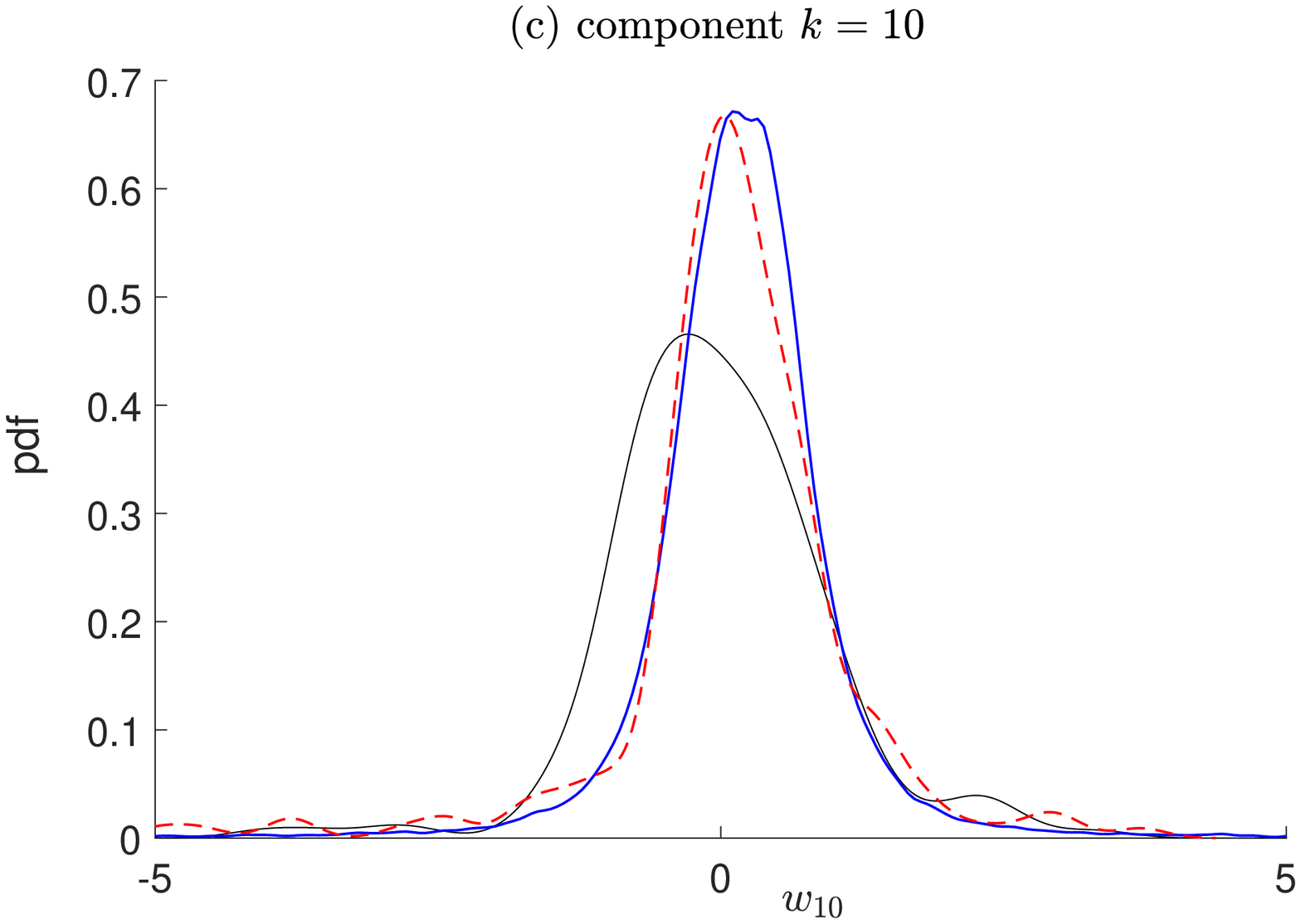} \hfill \includegraphics[width=6.5cm]{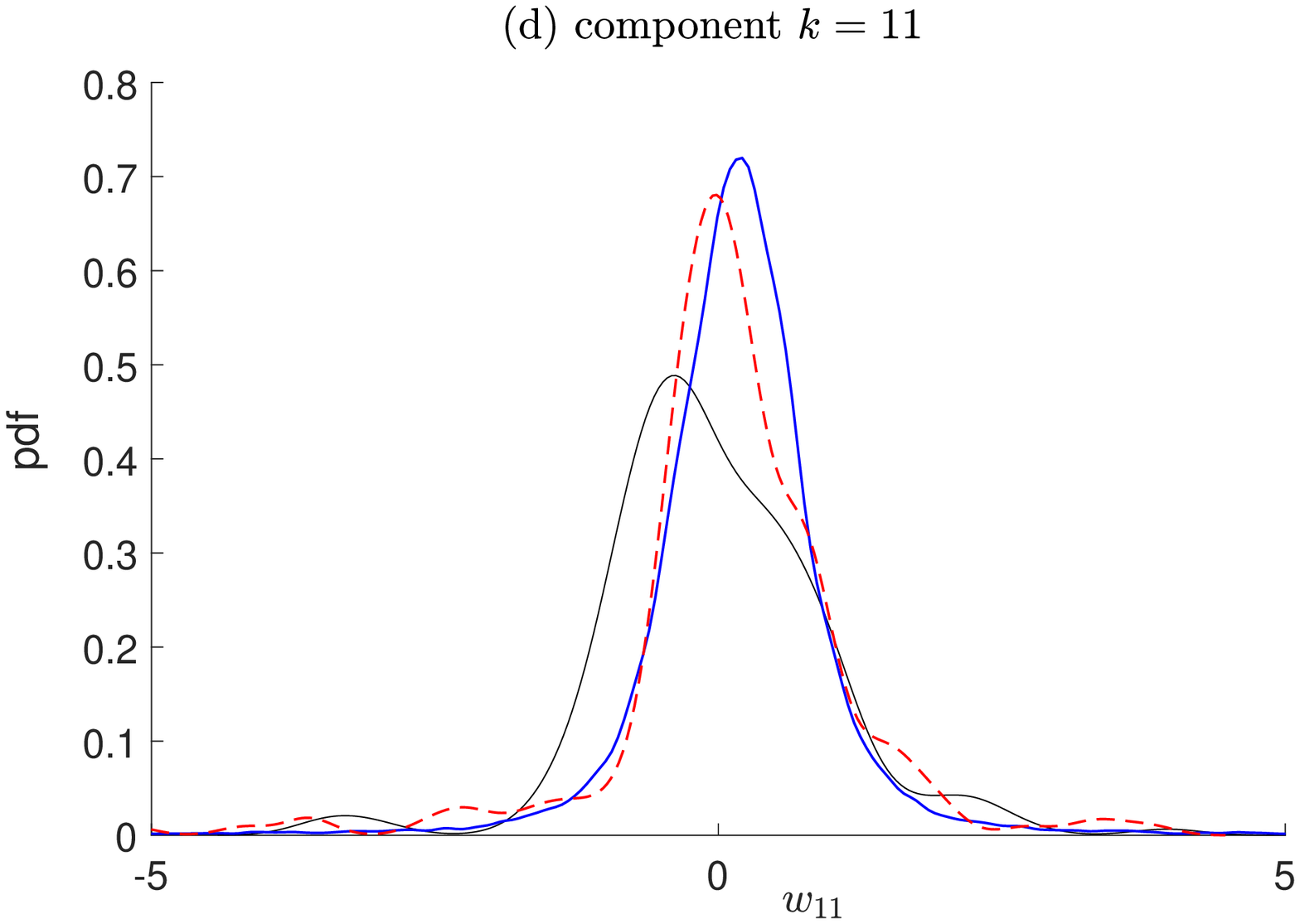}
  \caption{Application AP2: pdf $w\mapsto p_{\WW_k}^d(w)$ of $\WW_k$ estimated with the initial dataset $\DD_{N_d}$ of $N_d=200$ realizations (thin black line), pdf $w\mapsto p_{\WW_k}^\exper(w)$ of $\WW_k$ estimated with the experimental dataset $\DD_{n_r}^\exper$ of $n_r=200$ realizations (thick red dashed line), pdf $w\mapsto p_{\WW_k}^\post(w)$ of $\WW_k^\post$ estimated with $\varepsilon=0.5$, $N_d=200$, and  $\nu_\post= 40\, 000$ realizations (thick blue line), for $k=5$ (a), $k=6$ (b), $k=13$ (c), and $k=14$ (d).}  \label{figure6}
\end{figure}
\subsection{Quantities used for validating the choice of the value of
  the regularization parameter, for studying the convergence of the
  probabilistic learning, and for validating the proposed method}
\label{Section10.4}

\paragraph{Definition of the graphs that are plotted}

As already explained in Section~\ref{Section9}, we propose the value $0.5$ for the regularization parameter $\varepsilon$. In order to validate this choice, for the three applications and for $N_d$ fixed, we have plotted:

(i) the graph of function $\varepsilon\mapsto\underline\OVL (\varepsilon)$ defined by Eq.~\eqref{EQ102}, which has to be minimum in the neighborhood of $\varepsilon=0.5$;

(ii) the graph of function $\varepsilon\mapsto \conv_\std (\varepsilon)$ that is defined hereinafter and whose value should be close to $1$ in the neighborhood of $\varepsilon=0.5$. Let $\bfsigma^\post_\WW(\varepsilon)= (\sigma^\post_{\WW_1}(\varepsilon),$ $\ldots ,$ $\sigma^\post_{\WW_{m_w}}(\varepsilon))$ be the vector of the standard deviations estimated with the $\nu_\post$ realizations of the components of random vector $\WW^\post$ (estimated with the $\nu_\post$ realizations) and let $\bfsigma^\exper_\WW= (\sigma^\exper_{\WW_1},\ldots , \sigma^\exper_{\WW_{m_w}})$ be the standard deviations  of the components of random vector $\WW$ (estimated with experimental dataset $\DD^\exper_{n_r}$). The function
$\conv_\std$  is defined by $\conv_\std (\varepsilon) = \Vert \bfsigma^\post_\WW(\varepsilon)\Vert / \Vert \bfsigma^\exper_\WW\Vert$.

\paragraph{Studying the convergence of the probabilistic learning for the posterior model}
For $\varepsilon$ fixed at $0.5$, the convergence of the probabilistic learning is analyzed with respect to $N_d$ by studying the function
$N_d\mapsto\underline\OVL (N_d)$ defined by Eq.~\eqref{EQ102} (replacing $\varepsilon$ by $N_d$) and the function
$N_d\mapsto \conv_\std (N_d)$  such that $\conv_\std (N_d) = \Vert \bfsigma^\post_\WW(N_d)\Vert / \Vert \bfsigma^\exper_\WW\Vert$.

\paragraph{Validating the proposed method}
In addition to the quantities just described and for several components of index $k$, we will compare the graph of the pdf $w\mapsto p_{\WW_k}^d(w)$ of $\WW_k$ estimated with the initial dataset $\DD_{N_d}$ of $N_d=200$ realizations, with the graph of the pdf $w\mapsto p_{\WW_k}^\exper(w)$ of $\WW_k$ estimated with the experimental dataset $\DD_{n_r}^\exper$ realizations, and with the graph of the pdf $w\mapsto p_{\WW_k}^\post(w)$ of $\WW_k^\post$ estimated for $\varepsilon=0.5$ and $N_d=200$, and  $\nu_\post= 40\, 000$ realizations. The pdf $w\mapsto p_{\WW_k}^\post(w)$ should be close to $w\mapsto p_{\WW_k}^\exper(w)$ (the reference).
\subsection{Results and comments for  applications (AP1) and (AP2)}
\label{Section10.5}
The results are presented in Figs.~\ref{figure1} to \ref{figure3} for application (AP1) and in Figs.~\ref{figure4} to \ref{figure6} for application (AP2).

\noindent (i) Concerning the validation of the choice $\varepsilon =0.5$ of the regularization parameter, Figs.~\ref{figure1}-(left) and \ref{figure4}-(left) show that function $\varepsilon\mapsto\underline\OVL (\varepsilon)$  has effectively a minimum in the neighborhood of $\varepsilon=0.5$ for these two applications.\\

\noindent (ii) For the two applications with $\varepsilon=0.5$, the convergence of the probabilistic learning with respect to the size $N_d$ of the initial dataset that is  used in all the calculations detailed in Sections~\ref{Section3} to \ref{Section8}, Figs.~\ref{figure2} and \ref{figure5} show the results obtained for the functions $N_d\mapsto\underline\OVL (N_d)$ (left figure) and $N_d\mapsto \conv_\std (N_d)$ (right figure). For application (AP2), the convergence of the learning is slower and a best convergence could certainly be obtained by increasing the maximum value of $N_d$ that should be considered. Nevertheless, this slower convergence of the learning with respect to $N_d$ does not interfere with the validation of the proposed methodology, because, for a fixed value of $N_d$, the results obtained show that the posterior model that is estimated allows the prior model to be  significantly improved; a better convergence of the learning with respect to $N_d$ would lead to even greater improvement of the posterior model.\\

\noindent (iii) Concerning the validation of the proposed method, Figs.~\ref{figure1}-(right) and \ref{figure4}-(right) show that for $N_d=200$ and $\varepsilon=0.5$, the norm $\conv_\std (\varepsilon)$ of the vector of the standard deviations, normalized by its counterpart for the experiments, is close to $1$. For these two applications, Figs.~\ref{figure3} and \ref{figure6} show, for selected components $\WW_k$ of random vector $\WW$,
the comparison of three probability density functions: the pdf $w\mapsto p_{\WW_k}^d(w)$ of $\WW_k$ estimated with the initial dataset $\DD_{N_d}$ with $N_d=200$, the pdf $w\mapsto p_{\WW_k}^\exper(w)$ of $\WW_k$ estimated with the experimental dataset $\DD_{n_r}^\exper$ with  $n_r=200$, and the pdf $w\mapsto p_{\WW_k}^\post(w)$ of the posterior $\WW_k^\post$ estimated with $\varepsilon=0.5$, $N_d=200$, and $\nu_\post= 40\, 000$.
For each value of $k$ that is considered, the comparison between $p_{\WW_k}^d$ and $p_{\WW_k}^\exper$ shows that there are significant differences (mean value, standard deviation, non-Gaussianity) between these two pdf's, which justify the use of the Bayesian approach for improving $p_{\WW_k}^d$ with
$p_{\WW_k}^\post$. The values of $k$ selected for plotting, for each application, correspond to the greatest difference between these two pdfs.
An important element for the validation is the comparison between
$p_{\WW_k}^\post$ and $p_{\WW_k}^\exper$. It can be seen that the
results are quite good for these two applications.
\section{Application (AP3) to the ultrasonic wave propagation in biological tissue}
\label{Section11}
In this section, the methodology is applied to the ultrasonic wave
propagation in biological tissue for which $\WW$ is the vector of the
spatial discretization related to the non-Gaussian tensor-valued
random elasticity field of a damaged cortical bone due to
osteoporosis. This application will be referred to (AP3). All the
data concerning this application are described in order that the
application can be reproduced.
\subsection{Stochastic boundary value problem}
\label{Section11.1}
\begin{figure}[h!]
\centering
\includegraphics[width=10cm]{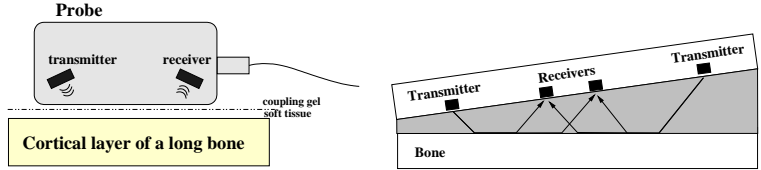}
\caption{Application AP3: scheme of the axial transmission technique.}
\label{figure7}
\end{figure}

This application deals with the numerical simulation of the axial transmission technique that is used in biomechanics for the identification of the cortical bone microstructure. The principle of the axial transmission technique is illustrated in Fig.~\ref{figure7}. An impulse is generated by a transmitter placed on the skin of a patient and then, the backscattered pressure field is recorded at distant receivers in the ultrasonic range.
\begin{figure}[h!]
\centering
\includegraphics[width=8cm]{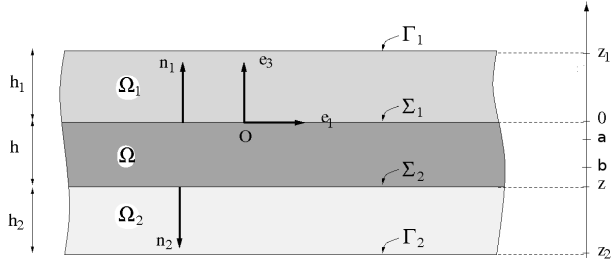}
\caption{Application AP3: Geometry of the multilayer system for the boundary value problem.}
\label{figure8}
\end{figure}

\paragraph{Boundary value problem}
The 2D physical space is equipped with a cartesian frame $(O, \bfe_1, \bfe_3)$ in which the coordinates of a point are denoted by $(x_1,x_3)$. A boundary value problem has been introduced \cite{Desceliers2008,Desceliers2012} for modeling the ultrasonic wave propagation in human cortical bone. It consists of a 2D semi-infinite multilayer medium  in the $\bfe_1$ longitudinal direction (see Fig.~\ref{figure8}). The model consists of an elastic semi-infinite layer $\Omega$ (cortical bone) with thickness $h$ in the $\bfe_3$ radial direction.
This elastic semi-infinite layer  is sandwiched between two acoustical fluid layers, $\Omega_1$ (skin and soft tissues) and $\Omega_2$ (bone marrow) with thicknesses $h_1$ and $h_2$ in the $\bfe_3$ radial direction. The media occupying domains $\Omega_1$ and $\Omega_2$ are homogeneous while
the cortical bone that occupies domain $\Omega$ is heterogeneous in the $\bfe_3$ direction. The probe (transmitter and receivers) is located in $\Omega_1$.

A mean (nominal) boundary value problem is written  in time and space domains considering the three coupled layers: linear acoustic wave equation formulated with the pressure field $P_1(\bfx,t)$ and $P_2(\bfx,t)$ in domains $\Omega_1$ and $\Omega_2$, and linear elastodynamics equation formulated with the displacement field $\bfD(\bfx,t)$ in domain $\Omega$. Such a formulation  requires to define,

\noindent (1) for the heterogeneous cortical bone $\Omega$, its mass density $\rho(x_3)$ and its $(3\times 3)$ matrix-valued effective elasticity field $\{[C(x_3)]\, , \ x_3\in ]-h,0[\}$;

\noindent (2) for the acoustic fluids $\Omega_1$ and $\Omega_2$, their mass densities $\rho_1 = \rho_2 = 1 000\, \rm kg.m^{-3}$, and their sound velocities $c_1=c_2 = 1 500\, \rm m.s ^{-1}$.  Note that the acoustic fluid $\Omega_2$ can also be viewed as an elastic solid for which the non zero components of its $(3\times 3)$ elasticity matrix, denoted as $[C^F]$, are equal to $\rho_2\, c_2^2$.

Introducing $a$ and $b$ such that $-h < b < a < 0$ (see Fig.~\ref{figure8} in which $z=-h$). In case of osteoporosis, there is a gradient of porosity in domain $\Omega$ in the $\bfe_3$ direction:

\noindent (1) for  $-h < x_3 < b $,   the cortical bone is a damaged material mostly made up of an acoustic fluid, which has the same behavior
as the acoustic fluid $\Omega_2$.

\noindent (2) for $ a < x_3 < 0$,  the cortical bone is an elastic solid that is modeled by a homogeneous  transverse isotropic elastic medium for which its $(3\times 3)$ elasticity matrix is denoted by $[C^S]$: the transverse Young modulus and the Poisson coefficient are $E_T = 9.8\,\rm GPa$ and $\nu_T=0.4$; the longitudinal Young modulus, the Poisson coefficient, and the shear modulus are respectively $E_L = 17.7\,\rm GPa$, $\nu_L = 0.38$, and $G_L=4.79\,\rm GPa$; its  mass density is $\rho_S = 1600\,\rm Kg.m^{-3}$.

\noindent (3) for $b \leq x_3 \leq a $, there is a gradient of porosity in the cortical bone.

The model proposed in \cite{Desceliers2012} is used for constructing the mean (nominal) model, based on the hypotheses defined in paragraphs (1) to (3) above, which is written, for all $x_3 \in ]-h,0[$, as
\begin{equation}
\rho(x_3)= (1-f(x_3))\, \rho_S + f(x_3)\, \rho_2 \, ,   \nonumber
\end{equation}
\begin{equation}
  [C(x_3)] = \beta_C\,\Big((1-f(x_3))\, [C^S] + f(x_3)\, [C^F]\Big)\, ,                                                 \label{EQ103}
\end{equation}
in which $\beta_C$ is a parameter that allows a bias to be introduced in the model, where $f(x_3) = 1$ if $x_3 < b$, $f(x_3)=0$ if $x_3 >a $, and $f(x_3) = \alpha_0 + \alpha_1\, x_3 + \alpha_2\, x_3^2 + \alpha_3\, x_3^3$ if $b\leq x_3 \leq a$ in which $\alpha_0 = a^2\,(a-3\,b)/(a-b)^3$, $\alpha_1 = 6\,a\,b/(a-b)^3$, $\alpha_2 = -3(a+b)/(a-b)^3$, and $\alpha_3= 2/(a-b)^3$.

\paragraph{Prior stochastic model of the matrix-valued effective elasticity field of the cortical bone}
In practice, the effective elasticity field of the cortical bone, which occupies domain $\Omega$, is a non-Gaussian random field and is modeled by a $(3\times 3)$ matrix-valued random field $\{[\bfC(x_3)]\, , \in ]-h,0[\}$  whose mean value is the field $[C(x_3)]$ defined by Eq.~\eqref{EQ103},
\begin{equation}
E\{[\bfC(x_3)]\} = [C(x_3)]  \quad , \quad \forall \, x_3 \in ]-h,0[ \, . \nonumber
\end{equation}
The prior probabilistic model of this non-Gaussian random field is taken in the set $\rm SFE^+$ introduced in \cite{Soize2017b}. The construction of this set of non-Gaussian matrix-valued random fields is based on the use of the Maximum Entropy principle for constructing a set of positive-definite random matrices. This prior probabilistic model depends only on two hyperparameters, a dispersion coefficient  $\delta_S$ and a spatial correlation length $\ell_S$.

\paragraph{Stochastic model for the acoustical source}
The transmitter is an acoustic point source located in domain $\Omega_1$, which delivers a random acoustical impulse, and is modeled by a random acoustical source density $Q$ such that
\begin{equation*}
  {\partial Q\over \partial t}(\bfx,t)= \rho_1\, F(t)\delta_0(x_1) \delta_0(x_3)\ ,
\end{equation*}
in which $\delta_0$ is the Dirac function on the real line at the origin. Time function $F$ is written as $F(t) = f_0\, \sin(2\pi F_c \,t)\, e^{-4(t\, F_c - 1 )^2}$ in which $F_c$ is the random central frequency whose probability distribution is uniform  on $[800,1200]\,\rm kHz$ and where $f_0=100\,\rm N$.  At time $t=0$, the system is assumed to be at rest. For each given realization of random field $[\bfC]$, the corresponding realization of (1) the random displacement field $\bfD$ and its associated Von Mises stresses fields $\bfS^{\rm VM}$ are computed in  $\Omega$, (2) the random pressure fields $P_1$ and $P_2$ are computed in $\Omega_1$ and $\Omega_2$. These numerical calculations are performed using the fast and efficient hybrid solver detailed in \cite{Desceliers2008}. It involves a spatial Fourier transform of the random boundary value problems  into the longitudinal direction ($\bfe_1$ direction) and a 1D finite element discretization into the radial direction ($\bfe_3$ direction).
\subsection{Illustration of results obtained with the stochastic boundary value problem}
\label{Section11.2}
\begin{figure}[h!]
\centering
\includegraphics[width=6.5cm]{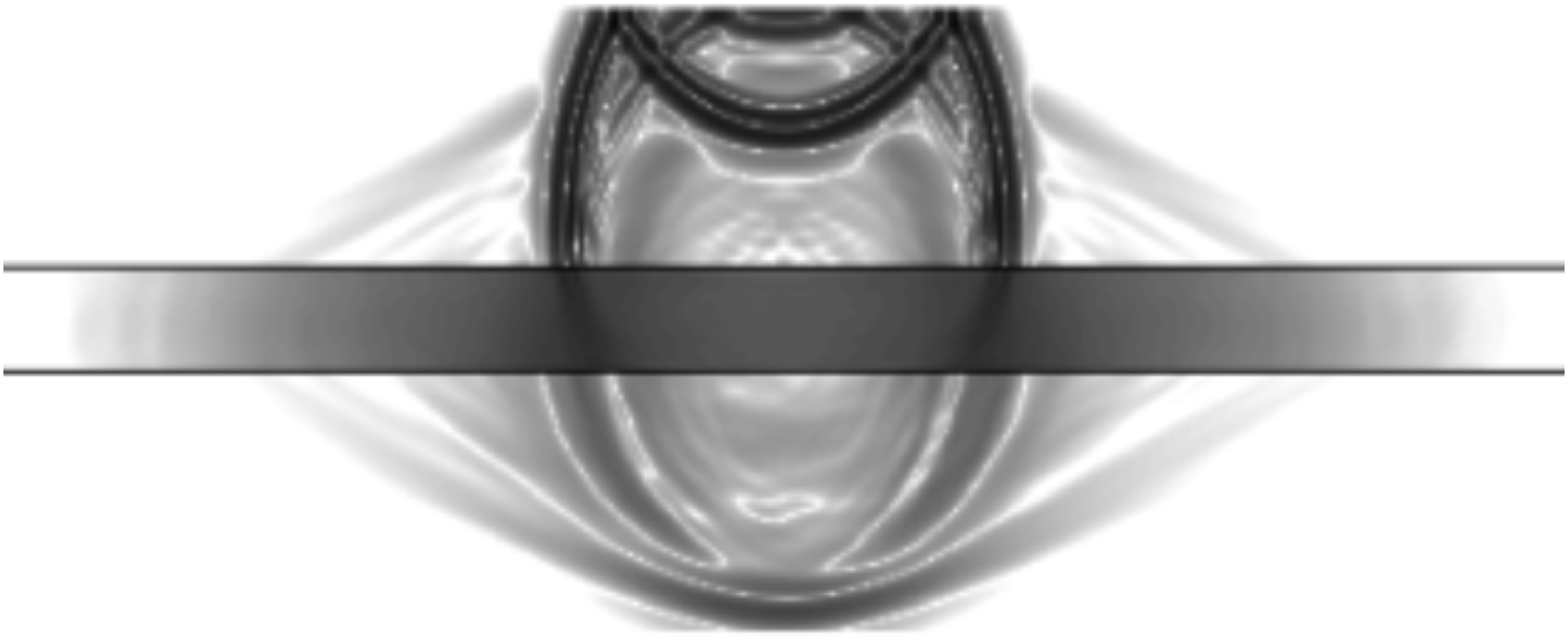} \hfill  \includegraphics[width=6.5cm]{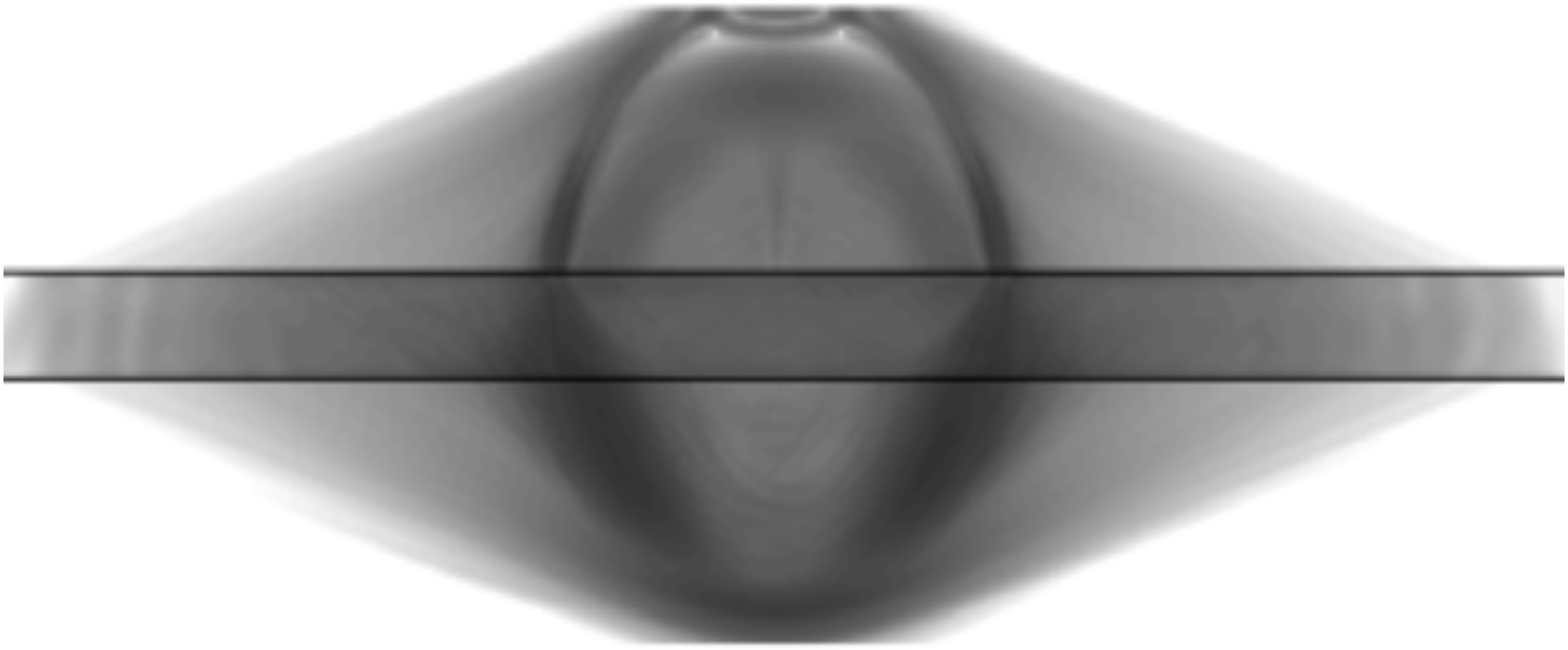}
\caption{Application AP3: Propagation of the mean (left figure) and variance (right figure) of the random wave in the three layers at $t = 9.72\times10^{-6}\rm s$ with for  $h_1 = 10^{-2}\,\rm m$, $h = 8\times10^{-3}\,\rm m$, $h_2 = 10^{-2}\,\rm m$.
Vertical direction: $x_3$. In grey color, mean and variance of wave fields $(x_1,x_3)\mapsto  P_1(x_1,x_3,t)$ (upper layer),  $(x_1,x_3)\mapsto  S^{\rm VM}(x_1,x_3,t)$ (sandwiched layer), $(x_1,x_3)\mapsto  P_2(x_1,x_3,t)$ (bottom layer).
}
\label{figure9}
\end{figure}

This section deals with an illustration of the ultrasonic wave propagation in the three-layers system using the stochastic boundary value problem defined in Section~\ref{Section11.1}, but for which the following particular configuration and parameterization are used. Note that, in presence of a gradient of porosity, the ultrasonic wave propagation is complex and the plot of such waves is difficult to interpret; consequently, for this illustration, it will be assumed that there is a damaged cortical bone without porosity gradient, which means that $-h < b = a < 0$. We consider the case $h_1 = 10^{-2}\,\rm m$, $h = 8\times10^{-3}\,\rm m$, $h_2 = 10^{-2}\,\rm m$, $b = a= -h/2$, and $z=-h$ (obviously, for generating the initial dataset used by  the probabilistic learning and for generating the experimental dataset required for the Bayesian approach, we will consider a porosity gradient ($-h < b < a < 0$)). The calculation has been performed with $\beta_C = 1$ and  $\rho_S = 1722\,\rm Kg.m^{-3}$. For this illustration, parameters $\delta_S$, $\ell_S$, and $F_c$ are considered deterministic (that will not be the case for generating the initial dataset) and are such that $\delta_S= 0.2$, $\ell_S = 3\times 10^{-3}\, \rm m$, and $F_c = 1 000\,\rm kHz$. The acoustical point source (transmitter) is located in $\Omega_1$ with coordinates  $(x_1 = 0 ,x_3 = 0.001) \,\rm m$.
In the $\bfe_3$ radial direction, the number of nodes used in the finite element interpolation of fields $P_1$, $P_2$, and $\bfD$ are $101$, $101$, and $82$, respectively. The Monte Carlo numerical simulation method is used as stochastic solver. The sampling time step is $\Delta t = 2.94\times10^{-6}\,\rm s$ and the number of time sampling points is $330$. The sampling spectral step is $\Delta k = 15.70\,\rm rad.m^{-1}$
and the number of spectral sampling points is $1 024$. At observation
time $t=9.72\!\times\! 10^{-6}\, \rm s$, Fig.~\ref{figure9} displays
the mean and variance of random fields $P_1$, $\bfS^{\rm VM}$, and
$P_2$. Figure~\ref{figure9} shows the lateral wave (or head wave)  propagating from the fluid-solid interface (plane wave front, which links the reflected P-wave front and the interface).
\subsection{Generation of the initial dataset}
\label{Section11.3}
Using the stochastic boundary value problem defined in Section~\ref{Section11.1}, the objective is to generate the initial dataset
$\DD_{N_d} =\{ \xx_d^j = (\qq_d^j,\ww_d^j) , j=1,\ldots , N_d\}$ (see Eq.~\ref{EQO1})
relative to random variable $\XX=(\QQ,\WW)$ in which $\QQ=(\QQ_1,\ldots,\QQ_{n_q})$ and $\WW=(\WW_1,\ldots,\WW_{n_w})$.
We then have to define the vector-valued random QoI, $\QQ$, the vector-valued random system parameter, $\WW$, and the $\RR^{n_u}$-valued random variable $\UU$, which are such that
\begin{equation}
\QQ= \ff(\UU,\WW) \, .                                                                                                 \label{EQ104}
\end{equation}
Note that the deterministic mapping $\ff$ cannot explicitly be described because this mapping is associated with the solution of the boundary value problem. The generation is carried out with $N_d=200$.\\

\noindent (i) Initial dataset $\DD_{N_d}$ is constructed for the case analyzed in \cite{Desceliers2012}, that is to say, for a damaged cortical bone with a gradient of porosity such that $h_1 = 2\!\times\! 10^{-3}\,\rm m$, $h = 8\!\times\! 10^{-3}\,\rm m$, $h_2 = 10^{-2}\,\rm m$, $a = -h/2$, $b=-h$, $z=-h$ and $\rho_S=1 600\,\rm Kg.m^{-3}$. The acoustical point source (transmitter) is located in $\Omega_1$ with coordinates  $(x_1 = 0 ,x_3 = 0.001) \,\rm m$. The dispersion coefficient $\delta_S$ and the spatial correlation length $\ell_S$ are modeled by random variables with uniform probability distributions on $[0,\, 0.7977]$ and  $[1, \, 8]\!\times\! 10^{-3}\, \rm m$ respectively. The central frequency $F_c$  is the uniform random variable defined in Section~\ref{Section11.1}.   The is no bias introduced in the model and consequently,  $\beta_C = 1$ in Eq.~\eqref{EQ103}.\\

\noindent (ii) The number of nodes for the finite element discretization  of $P_1$, $P_2$, and $\bfD$ in the $\bfe_3$ radial direction  are  $21$, $101$, and $162$, respectively. The sampling time step is $\Delta t = 4.2565\!\times\! 10^{-6}\,\rm s$ and the sampling spectral step is $\Delta k = 44.88\,\rm rad.m^{-1}$. The number of time sampling points is $300$ and the number of  spectral sampling points is $2 048$.\\

\noindent (iii) Let $\QQ$ be the random vector of the $300$ time sampling points of the random pressure field $P_1$  at positions $\{(x_1^\ell, x_3 =10^{-3}\,\rm m) ,\ell=1,\ldots,14\}$ ($14$ receivers) in which $x_1^\ell = 13.1\!\times\! 10^{-3} + \ell\,\Delta x_1$ with $\Delta x_1 = 0.8\!\times\! 10^{-3}\,\rm m$. Thus, $\QQ$ is a $\RR^{n_q}$-valued random vector with $n_q = 4 200$.\\

\noindent (iv) Let $\WW$ be the random vector of all the random variables
\begin{equation}
\{[\bfL(x_3^\ell)]_{ij}\, , 1 \leq i < j \leq 3\} \cup \{\log([\bfL(x_3^\ell)]_{jj})\, , 1\leq  j \leq 3\} \, , \nonumber
\end{equation}
in which $\{ [\bfL(x_3^\ell)]_{ij}\}_{ij}$ are the entries of the random upper triangular matrix $[\bfL(x_3^\ell)]$ constructed by the following Cholesky factorization, $[\bfC(x_3^\ell)] = [\bfL(x_3^\ell)]^T[\bfL(x_3^\ell)]$. The points $\{ x_3^\ell , 1 \leq \ell \leq 120\}$ are the coordinates for the $\ell$-th integration points of the finite element mesh of interval $[-h,0]$ (see Fig.~\ref{figure7}). Thus,  $\WW$ is a $\RR^{n_w}$-valued random vector with $n_w = 720$.\\

\noindent (v) The $\RR^{3}$-valued random variable $\UU$ is written as $\UU = (\delta_s,\ell_s,F_c)$.
\subsection{Generation of the experimental dataset}
\label{Section11.4}
The experimental dataset $\DD^\exper_{n_r}$ is generated with $n_r=200$ independent experimental realizations $\{\qq^{\exper,r},r=1,\ldots n_r\}$
of $\QQ^\exper = (\QQ_1^\exper,\ldots ,\QQ_{n_q})$, which are such that
\begin{equation}
\QQ^\exper = \ff(\UU^\exper,\WW^\exper) \, ,                                                                                  \label{EQ105}
\end{equation}
in which the deterministic mapping $\ff$ is the same as the one used in Eq.~\eqref{EQ103} and corresponds to the use of the boundary value problem
defined in Sections~\ref{Section11.1} and \ref{Section11.3}. The random vectors $\UU^\exper$ and $\WW^\exper$ are constructed as independent copies of random vectors $\UU$ and $\WW$ define in Sections{\ref{Section11.1} and \ref{Section11.3},  but the bias on the mean model of the random effective elasticity matrix introduced in Eq.~\eqref{EQ103} is chosen as $\beta_C = 0.9$. It should be noted that the $n_r$ independent realizations $\{\ww^{\exper,r},r=1,\ldots n_r\}$ of random vector $\WW^\exper$ are generated in order to construct the simulated experiments $\{\qq^{\exper,r},r=1,\ldots n_r\}$ using  Eq.~\eqref{EQ105}, but these realizations are not used in the Bayesian approach proposed. Nevertheless, these realizations of $\WW_\exper$ will be used for estimating the probability density functions $\{w\mapsto p_{\WW_k}^\exper(w)\}_k$ of the components $\{\WW^\exper_k\}_k$ of $\WW^\exper$ in order to validate the methodology proposed (comparing $p_{\WW_k}^\exper$ to the posterior pdf $p_{\WW_k}^\post$).
\subsection{Values of the numerical parameters, observed quantities for convergence analyses, and validation}
\label{Section11.5}
The values of the numerical parameters introduced in the algorithm are summarized in Table~\ref{Table1} (column relative to (AP3)). All the given values of the numerical parameters have been obtained by using the criteria introduced in the theory or have been estimated by performing a local convergence analysis. The quantities used for validating the choice of the value of the regularization parameter $\varepsilon$, for studying the convergence of the probabilistic learning with respect to $N_d$, and for validating the method proposed, are similar to those introduced in Section~\ref{Section10.4} for Applications (AP1) and (AP2).
\begin{figure}[h!]
  \centering
  \includegraphics[width=6.5cm]{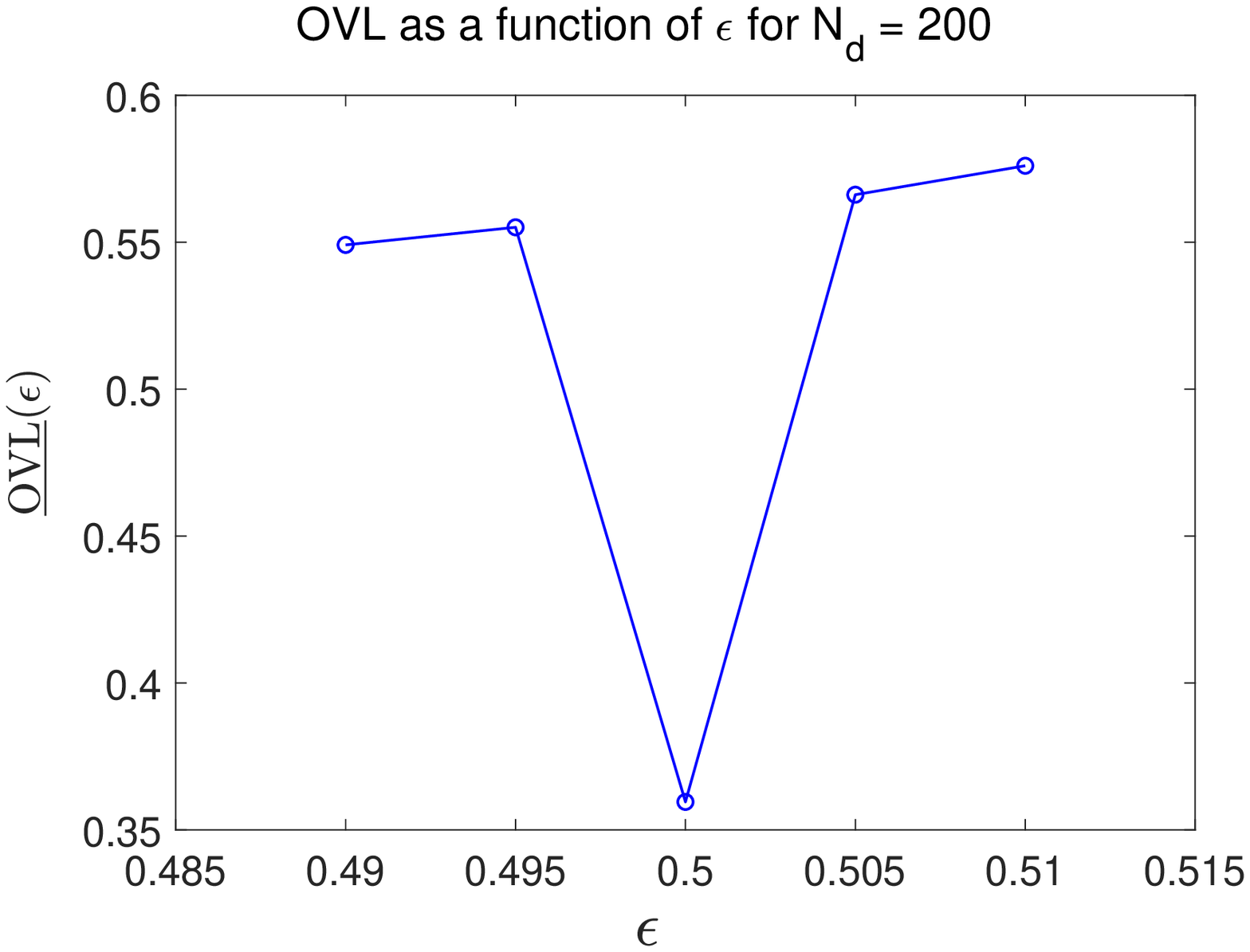} \hfill \includegraphics[width=6.5cm]{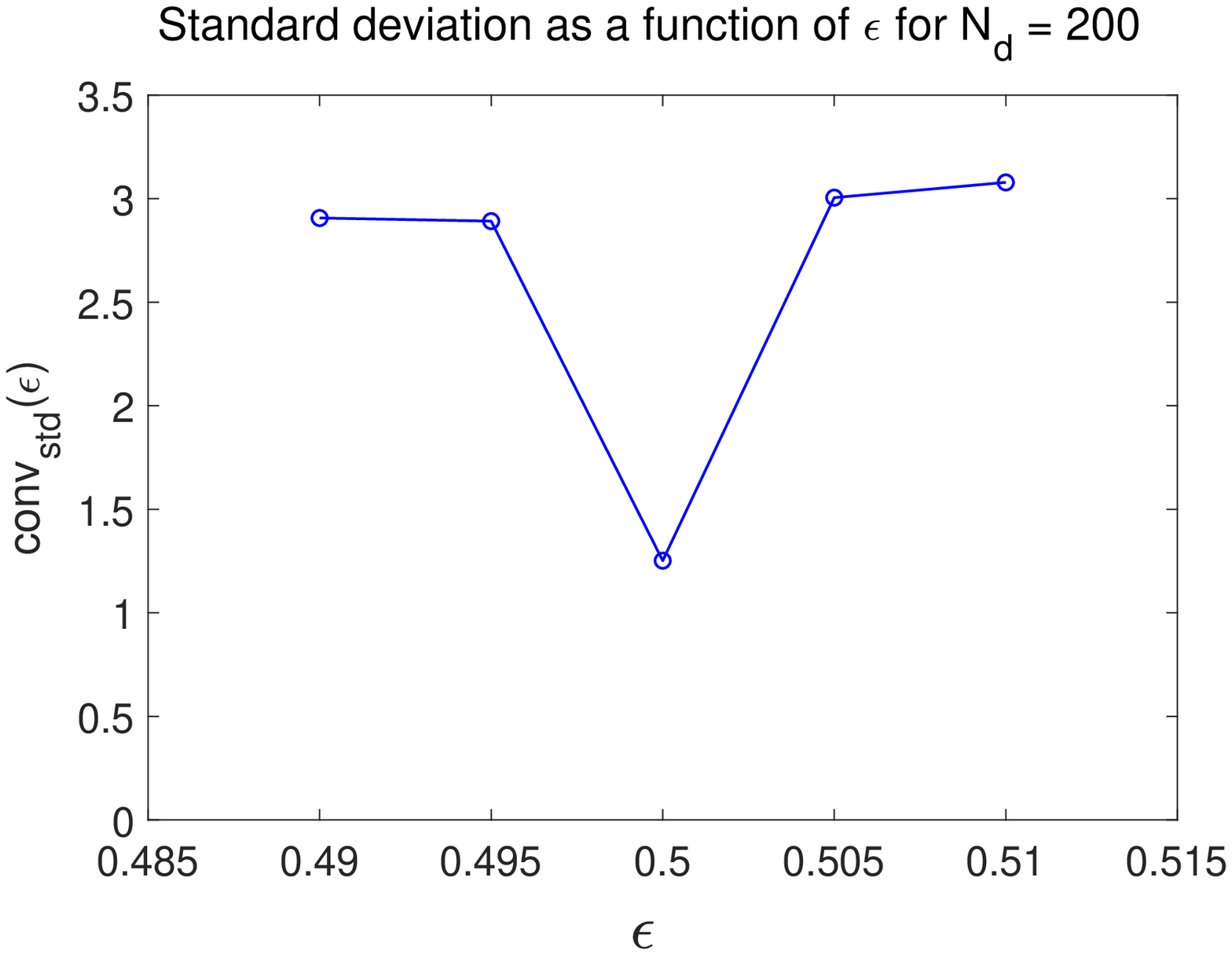}
\caption{Application AP3: validation of the choice $\varepsilon=0.5$. For $N_d = 200$, graph of
$\varepsilon\mapsto\underline\OVL (\varepsilon)$ (left) and  graph of $\varepsilon\mapsto \conv_\std (\varepsilon)$ (right).}  \label{figure10}
\end{figure}
\begin{figure}[h!]
  \centering
  \includegraphics[width=6.5cm]{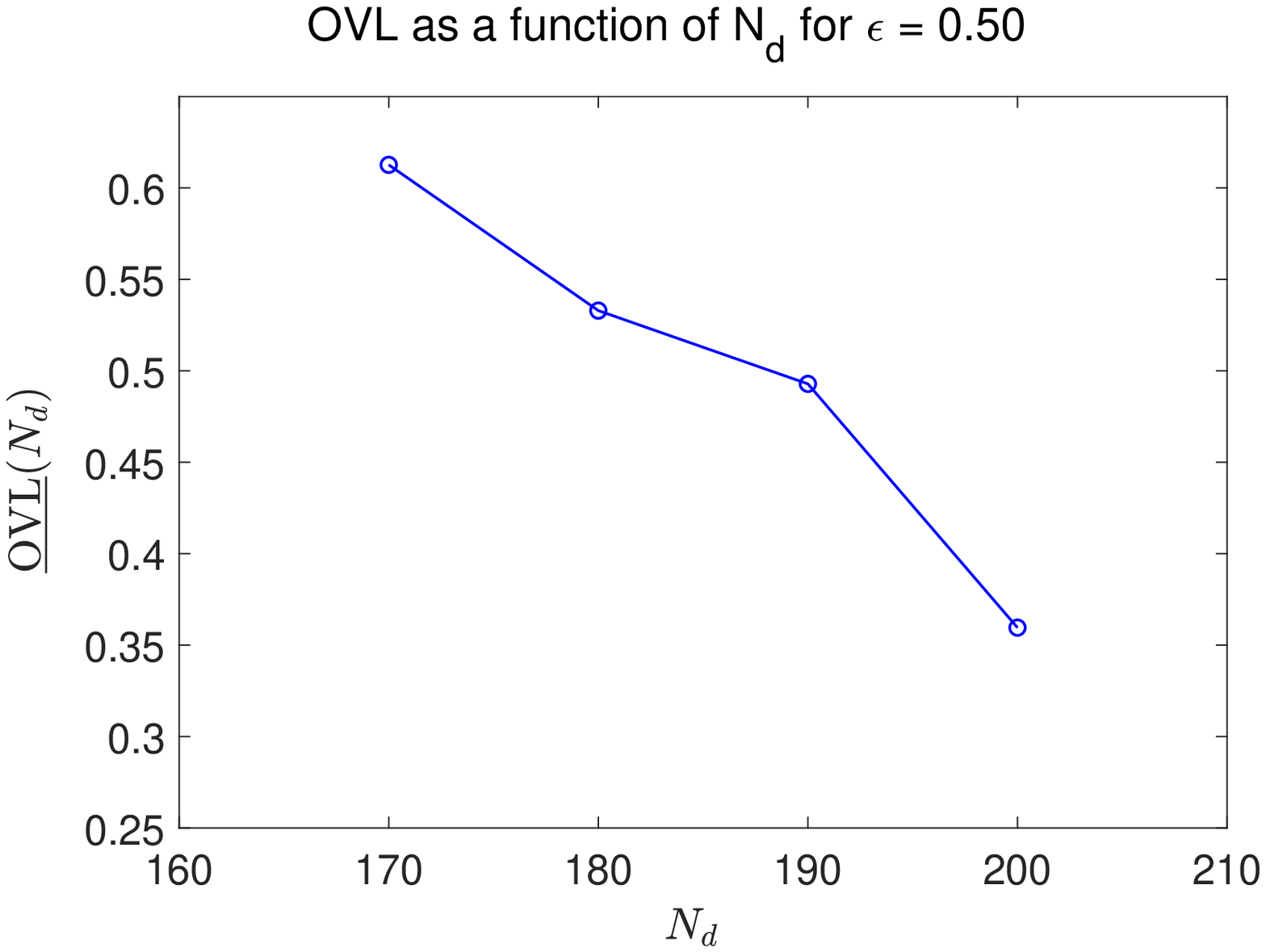} \hfill \includegraphics[width=6.5cm]{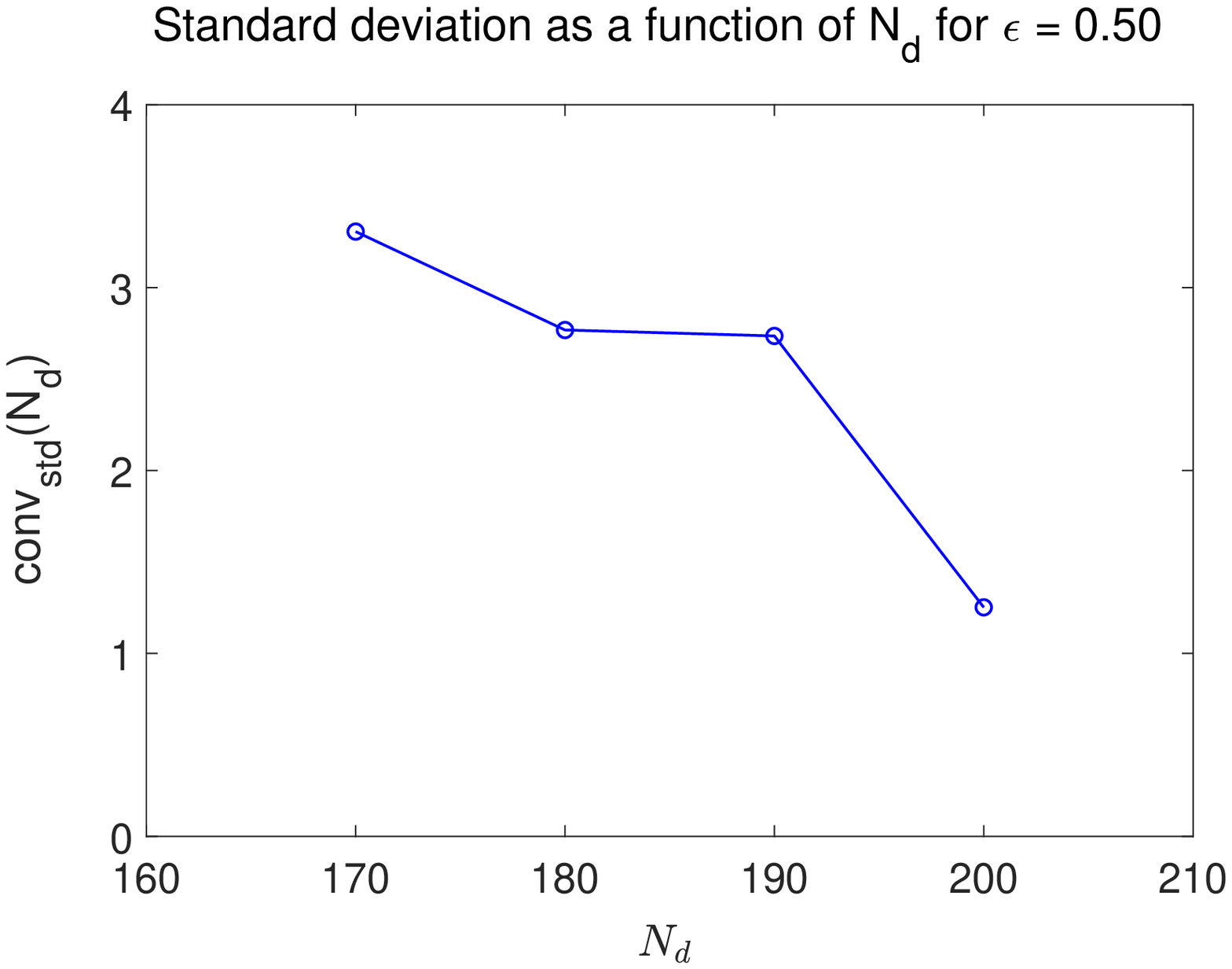}
\caption{Application AP3: convergence of the probabilistic learning with respect to $N_d$. For $\varepsilon=0.5$, graph of
$N_d\mapsto\underline\OVL (N_d)$ (left) and  graph of $N_d\mapsto \conv_\std (N_d)$ (right).}  \label{figure11}
\end{figure}
\begin{figure}[h!]
  \centering
  \includegraphics[width=6.5cm]{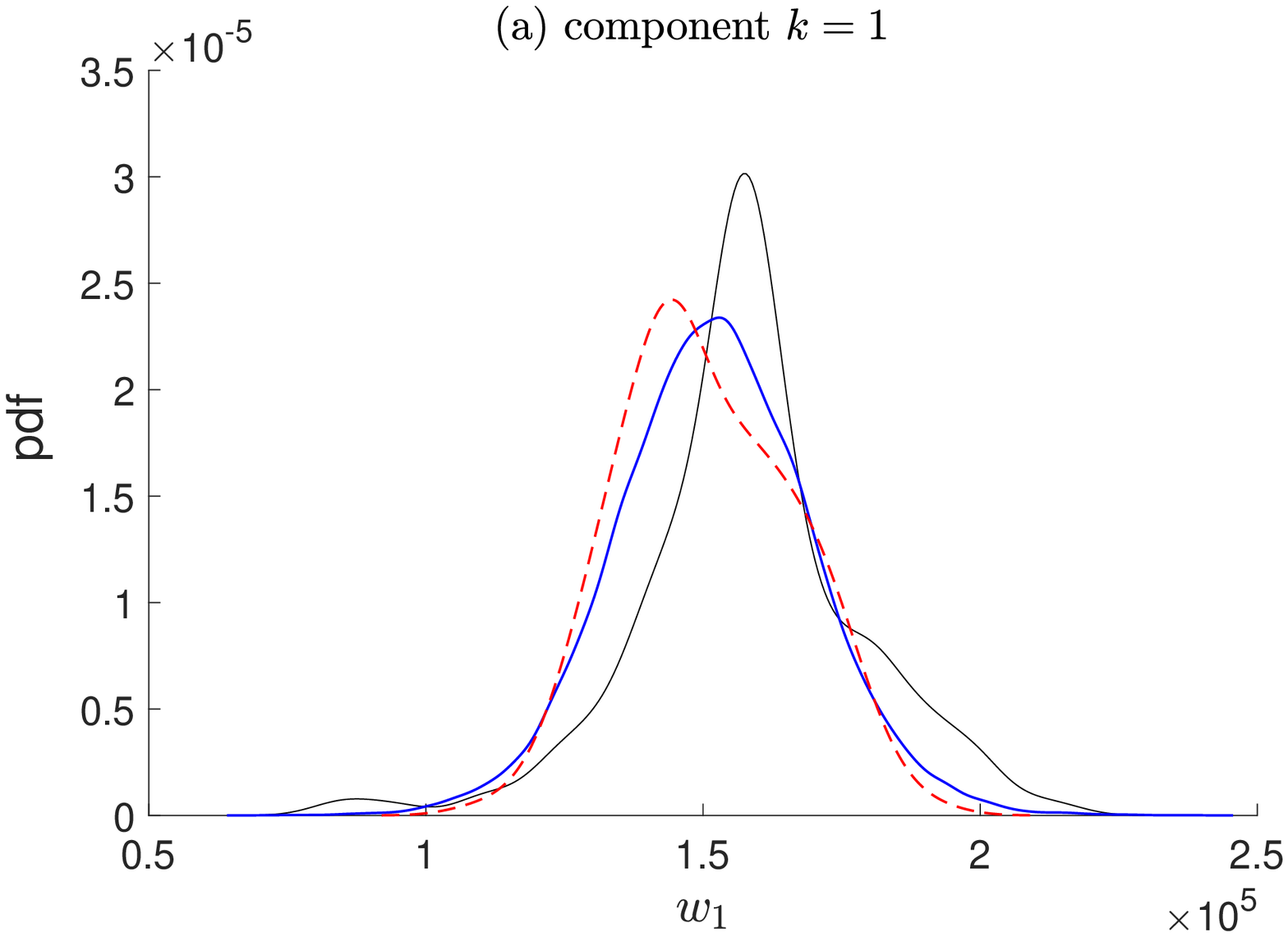} \hfill \includegraphics[width=6.5cm]{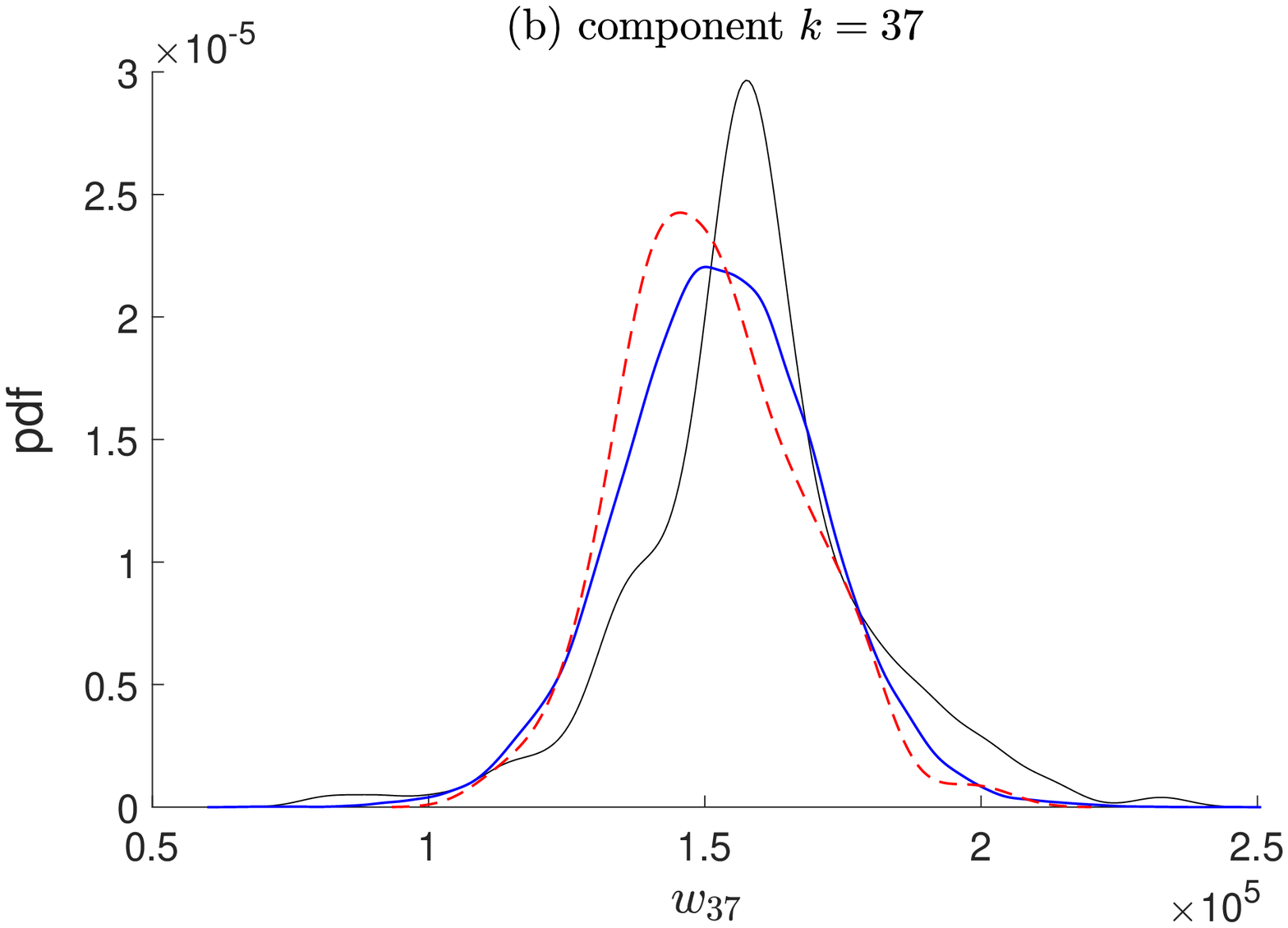}
  \includegraphics[width=6.5cm]{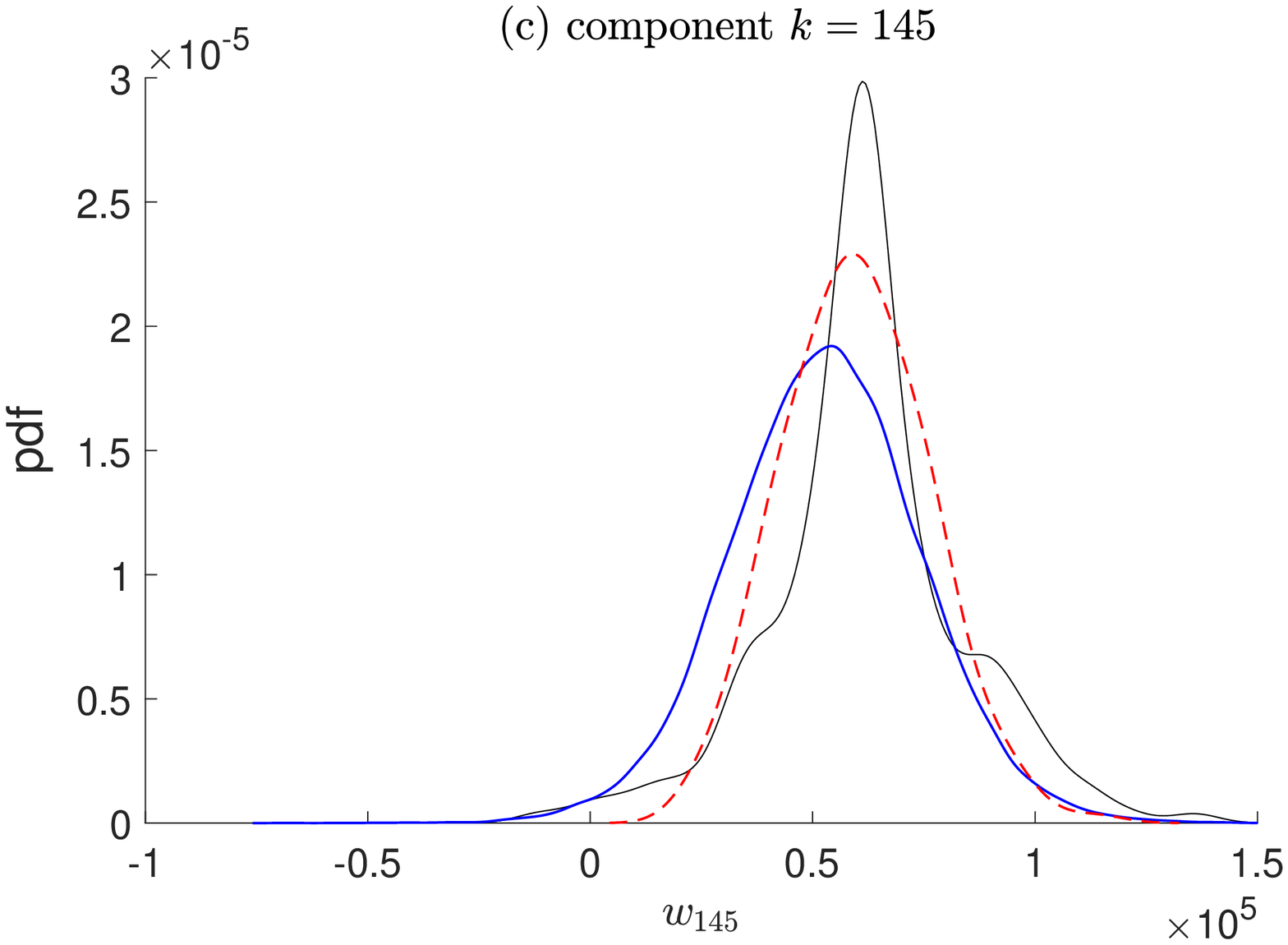} \hfill \includegraphics[width=6.5cm]{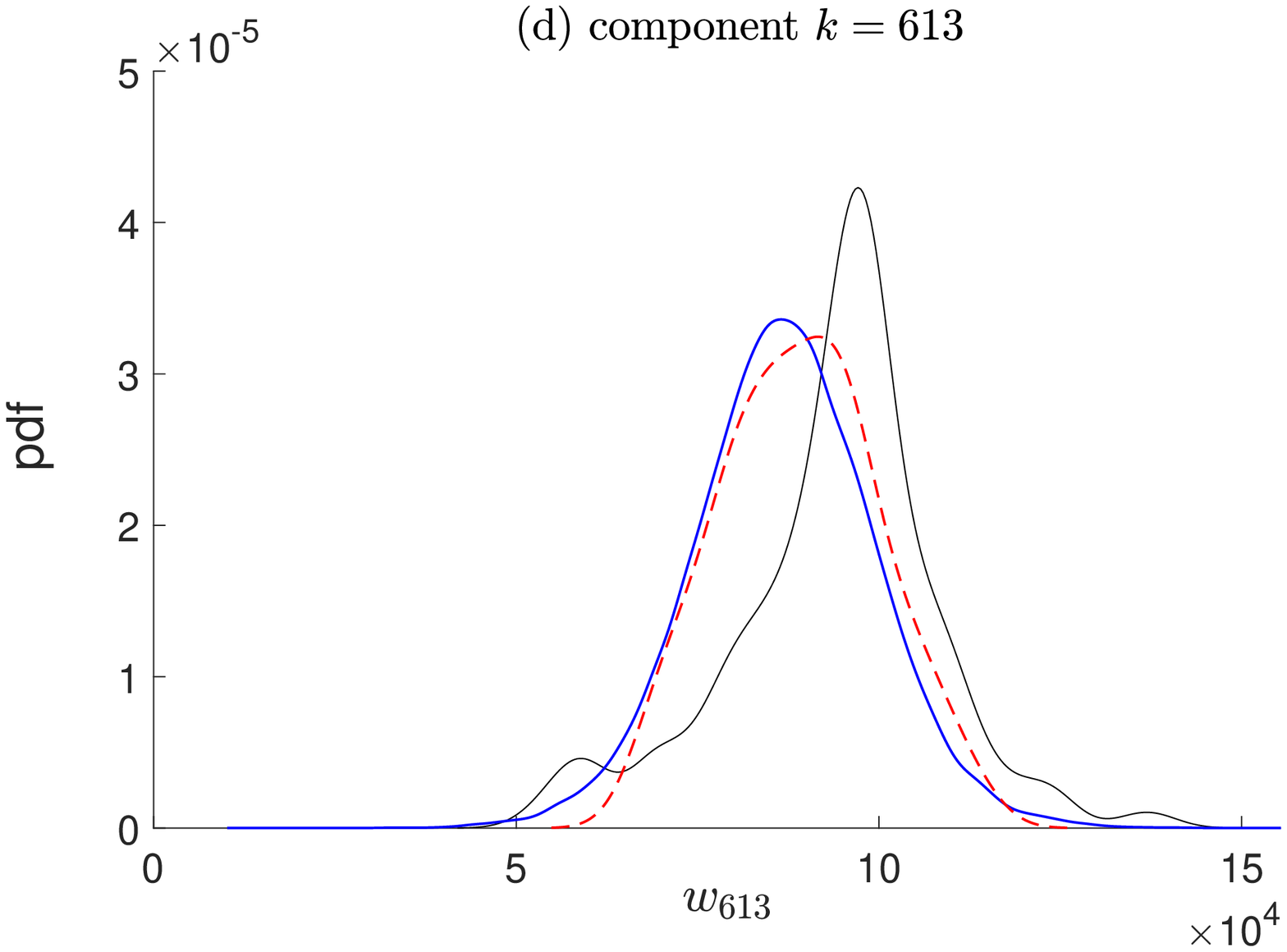}
  \caption{Application AP3: pdf $w\mapsto p_{\WW_k}^d(w)$ of $\WW_k$ estimated with the initial dataset $\DD_{N_d}$ of $N_d=200$ realizations (thin black line), pdf $w\mapsto p_{\WW_k}^\exper(w)$ of $\WW_k$ estimated with the experimental dataset $\DD_{n_r}^\exper$ of $n_r=200$ realizations (thick red dashed line), pdf $w\mapsto p_{\WW_k}^\post(w)$ of $\WW_k^\post$ estimated with $\varepsilon=0.5$, $N_d=200$, and  $\nu_\post= 40\, 000$ realizations (thick blue line), for $k=5$ (a), $k=6$ (b), $k=13$ (c), and $k=14$ (d).}  \label{figure12}
\end{figure}
\subsection{Results and comments for  application (AP3)}
\label{Section11.6}
The results are presented in Figs.~\ref{figure10} to \ref{figure12}.

\noindent (i) Concerning the validation of the choice $\varepsilon =0.5$ of the regularization parameter, Fig.~\ref{figure10}-(left) shows that function $\varepsilon\mapsto\underline\OVL (\varepsilon)$  has effectively a minimum in the neighborhood of $\varepsilon=0.5$, as obtained for applications (AP1) and (AP2) presented in Section~\ref{Section10}.

\noindent (ii) Concerning the convergence of the probabilistic learning with respect to size $N_d$ of the initial dataset that is used in all the calculations detailed in Sections~\ref{Section3} to \ref{Section8}, Fig.~\ref{figure11} shows the results obtained for the functions $N_d\mapsto\underline\OVL (N_d)$ (left figure) and $N_d\mapsto \conv_\std (N_d)$ (right figure) with $\varepsilon=0.5$. The convergence of the learning is slower and a best convergence could certainly be obtained by increasing the maximum value of $N_d$ that should be considered, but as already explained in Section~\ref{Section10.5},  this slower convergence of the learning with respect to $N_d$ does not interfere with the validation of the proposed methodology (see the explanation given in Section~\ref{Section10.4}-(ii)).

\noindent (iii) Concerning the validation of the method proposed, Fig.\ref{figure10}-(right) shows that, for $N_d=200$ and $\varepsilon=0.5$, the norm $\conv_\std (\varepsilon)$ of the vector of the standard deviations, normalized by its counterpart for the experiments, is close to $1$. Figure~\ref{figure12} shows, for selected components $\WW_k$ of random vector $\WW$, the comparison of three probability density functions: the pdf $w\mapsto p_{\WW_k}^d(w)$ of $\WW_k$ estimated with the initial dataset $\DD_{N_d}$ with $N_d=200$, the pdf $w\mapsto p_{\WW_k}^\exper(w)$ of $\WW_k$ estimated with the experimental dataset $\DD_{n_r}^\exper$ with  $n_r=200$, and the pdf $w\mapsto p_{\WW_k}^\post(w)$ of the posterior $\WW_k^\post$ estimated with $\varepsilon=0.5$, $N_d=200$, and $\nu_\post= 40\, 000$.
For each value of $k$ that is considered, the comparison between $p_{\WW_k}^d$ and $p_{\WW_k}^\exper$ shows that there are significant differences (mean value, standard deviation, non-Gaussianity) between these two pdf's, which justifies the use of the Bayesian approach for improving $p_{\WW_k}^d$ with
$p_{\WW_k}^\post$. An important element for the validation is the comparison between $p_{\WW_k}^\post$ and $p_{\WW_k}^\exper$. It can be seen that the results are not perfect. This appearance is due to the fact that a high-dimensional inverse statistical problem must be solved and that the capacity of the Bayes formulation to solve it depends on many factors, the most important of which is certainly the sensitivity of the quantities of interest with respect to certain components of the elastic tensor field at certain spatial points of the elastic medium of the dynamical  fluid-structure coupled problem. If these observed
quantities of interest are not very sensitive to the realizations of $\WW_k $, which represent random values at a certain spatial point of a certain component of the tensor of elasticity, then it is difficult to identify their probability density functions.

\begin{table}[tbhp]
\caption{For applications (AP1), (AP2), and (AP3), the table defines the values of all the parameters introduced in the algorithm, and which are used in computations.}
\label{Table1}
\centering
\begin{tabular}{|l|l|c|c|c|}
\hline
                            & Parameters                        & (AP1)                & (AP2)                & (AP3)               \\
\hline
Dimensions                  & $n_q$                             & $200$                & $20\, 000$           & $4\,200$            \\
                            & $n_w$                             & $20$                 & $20$                 & $720$               \\
                            & $n=n_q+n_w$                       & $220$                & $20\, 020$           & $4\, 920$           \\
                            & $N_d$                             & $\leq 200$           & $\leq  200$          & $\leq 200$          \\
                            & $\max N_d$                        & $200$                & $200$                & $200$               \\
                            & $n_r$                             & $200$                & $200$                & $200$               \\
\hline
Learning step               & $\nu_x$                           & $9$                  & $15$                 & $164$               \\
(Appendix A)                & $\varepsilon_\diff$               & $48$                 & $120$                & $350$               \\
                            & $m$                               & $12$                 & $17$                 & $166$               \\
                            & $f_0$                             & $1.5$                & $1.5$                & $1.5$               \\
                            & $n_\MC$                           & $150$                & $150$                & $150$               \\
                            & $\nu_\ar = N_d\times n_\MC$       & $30\, 000$           & $30\, 000$           & $30\, 000$          \\
                            & $\Delta t$                        & $0.1649$             & $0.0903$             & $0.2163$            \\
                            & $M_0$                             & $100$                & $100$                & $100$               \\
                            & $\ell_0$                          & $100$                & $100$                & $100$               \\
\hline
PCA of                      & $\nu_q$                           & $6$                  & $12$                 & $125$               \\
$\bfQ$ and $\bfW$           & $\err_\bfQ(\nu_q)$                & $4.3\times 10^{-5}$  & $5.3\times 10^{-5}$  & $3.9\times 10^{-4}$ \\
                            & $\nu_w$                           & $3$                  & $3$                  & $72$                \\
                            & $\err_\bfW(\nu_w)$                & $3.7\times 10^{-15}$ & $5.3\times 10^{-15}$ & $2.7\times 10^{-4}$ \\
\hline
Posterior step              & $\nu=\nu_q+\nu_w$                 & $9$                  & $15$                 & $197$               \\
                            & $\nu_1$                           & $6$                  & $12$                 & $125$               \\
                            & $\varepsilon$                     & $0.5$                & $0.5$                & $0.5$               \\
                            & $N_s$                             & $200$                & $200$                & $200$               \\
                            & $\varepsilon_{N_s}$               & $0.0422$             & $0.0322$             & $0.0433$            \\
                            & $f_0^\post$                       & $10^{-5}$            & $10^{-5}$            & $10^{-3}$           \\
                            & $\varepsilon_\diff^\post$         & $4.0\times 10^{3}$   & $3.5\times 10^{3}$   & $3.0\times 10^{5}$  \\
                            & $m_\post$                         & $9$                  & $11$                 & $69$                \\
                            & $n_\MC^\post$                     & $200$                & $200$                & $200$               \\
                            & $\nu_\post=n_\MC^\post\times N_s$ & $40\, 000$           & $40\, 000$           & $40\, 000$          \\
                            & $\Delta t$                        & $0.0277$             & $0.03384$            & $0.05854$           \\
                            & $M_0^\post$                       & $100$                & $100$                & $100$               \\
                            & $\ell_0^{\,\post}$                & $10\, 000$           & $10\, 000$           & $120$               \\
\hline
\end{tabular}
\end{table}

\section{Conclusion}
In this paper, we have presented a methodology for implementing the
Bayesian inference in the framework of the small-data challenge using
the probabilistic learning on manifolds under the following
hypotheses: the likelihood probability distribution is not Gaussian
and cannot be approximated by a Gaussian measure, the problem can be
in high dimension, the number of given realizations in the initial
dataset of the prior model is assumed to be small, which corresponds,
for instance, to the use of an expensive computer code for generating
the initial data set (training), the number of experimental
realizations is also small, and the number of posterior realizations
can be arbitrarily large. For solving these difficult problem, a novel
methodology has been developed. The method and the associated algorithms have
been adapted to take into account all the constraints induced by the
given framework. Three applications have been presented for validating
the approach  proposed: two are relatively simple and can easily be
reproduced, and the third one corresponds to a difficult statistical
inverse problem. The results obtained are good. The method proposed
will have to be tested for many other applications for confirming its
robustness and its capability to treat problems in high dimension and
in the framework of the small-data challenge.

\section*{Acknowledgments}
Part of this research was supported by the PIRATE project funded under
DARPA's AIRA program.
%
%
\appendix
%
%
\section{Summary of the algorithm of the probabilistic learning on manifolds}
\label{AppendixA}
In this Appendix, we summarize the algorithm of the probabilistic learning on manifolds (PLoM) that is used in Section~\ref{Section4}. This algorithm has been introduced in \cite{Soize2016}.  Complementary developments can be found in \cite{Ghanem2018a,Soize2019b}. Applications and validations can be found in \cite{Ghanem2018b,Ghanem2019,Soize2019,Soize2019a}. In addition, we give the formula for estimating the values of the two hyperparameters that control the algorithm of the PLoM.\\

\noindent Let $\{\bfx_d^j= (\bfq_d^j,\bfw_d^j) , j=1,\ldots , N_d\}$ be the set of the $N_d$ independent realizations given in $\RR^{n} = \RR^{n_q}\times \RR^{n_w}$ with $n=n_q+n_w$, which constitute the \textit{initial data set} $D_{N_d}$. Let $\bfX=(\bfQ,\bfW)$ be the random variable
with values in $\RR^{n} = \RR^{n_q}\times \RR^{n_w}$ for which $\{\bfx_d^j, j=1,\ldots , N_d\}$ constitutes $N_d$ independent realizations. The objective of the PLoM is to generate $\nu_\ar\gg N_d$  additional  realizations
$\{\bfx_\ar^{\ell},\ell = 1, \ldots ,\nu_\ar\}$  of random vector $\bfX$. As soon as the set $\{\bfx_\ar^\ell,\ell=1,\ldots,\nu_\ar\}$ has been constructed, the additional realizations for $\bfQ$ and $\bfW$ can be extracted as $(\bfq_\ar^\ell,\bfw_\ar^\ell) = \bfx_\ar^\ell$ for $\ell = 1, \ldots ,\nu_\ar$,
which constitute the \textit{learned dataset} $D_{\nu_\ar}$.\\

\noindent \textit{A.1. Normalization of the initial dataset}.
The $N_d$ independent realizations $\{\bfx_d^j , j=1,\ldots , N_d\}$ of $\bfX$ with values in $\RR^{n}$ can be represented by the matrix $[x_d]= [\bfx_d^1 \ldots \bfx_d^{N_d}]$ in $\MM_{n,N_d}$. Let $[\bfX]= [\bfX^1,\ldots ,\bfX^{N_d}]$ be the random
matrix with values in $\MM_{n,N_d}$, whose columns are $N_d$ independent copies of random vector $\bfX$. Therefore, $[x_d]$ is one realization of random matrix $[\bfX]$. The normalization of random matrix $[\bfX]$ is attained with the random matrix $[\bfH] = [\bfH^1,\ldots ,\bfH^{N_d}]$ with values in $\MM_{\nu_x,N_d}$ with $\nu_x\leq n$, obtained by using the principal component analysis of random vector $\bfX$. Consequently, random matrix $[\bfX]$ is written as,
\begin{equation}
[\bfX] = [\underline x] + [\varphi]\, [\lambda]^{1/2}\, [\bfH]\, ,                                                     \label{EQA0}
\end{equation}
in which $[\lambda]$ is the $(\nu_x\times\nu_x)$ diagonal matrix of the $\nu_x$ positive eigenvalues of the empirical estimate of the covariance matrix of $\bfX$ (computed using $\bfx_d^1, \ldots, \bfx_d^{N_d}$), where $[\varphi]$ is the $(n\times\nu_x)$ matrix of the associated eigenvectors such $[\varphi]^T\,[\varphi]= [I_{\nu_x}]$, and where  $[\underline x]$ is the matrix in $\MM_{n,N_d}$ with identical columns, each one being equal to the empirical estimate $\underline\bfx\in\RR^{n}$ of the mean value of random vector $\bfX$ (computed using $\bfx_d^1, \ldots, \bfx_d^{N_d}$).
The columns of $[\bfH]$ are $N_d$ independent copies of a random vector $\bfH$ wit values in $\RR^{\nu_x}$.
 The realization $[\eta_d] = [\bfeta^{1} \ldots \bfeta^{N_d}] \in \MM_{\nu_x,N_d}$ of $[\bfH]$ (associated with the realization $[x_d]$ of $[\bfX]$) is computed by $[\eta_d] =  [\lambda]^{-1/2} [\varphi]^T\, ([x_d] - [\underline x])$.
When  $n$ is small, $\nu_x$ can be chosen as $n$. If some eigenvalues are zero, they must be eliminated and then $\nu_x < n$. When $n$ is high, a statistical reduction can be done as usual and therefore $\nu_x < n$ in such a case.\\

\noindent \textit{A.2. Diffusion-maps basis}.
This is an algebraic basis of vector space $\RR^{N_d}$, which  is constructed using the diffusion maps proposed in \cite{Coifman2005}. Let $[\bb]$ be the positive-definite diagonal real matrix in $\MM_{N_d}$ such that $[\bb]_{ij} = \delta_{ij}\,\sum_{j'=1}^{N_d} [K]_{jj'}$ in which $[K]_{jj'} = \exp(-\frac{1}{4\,\varepsilon_\pdiff} \Vert\bfeta^{j} -\bfeta^{j'}\Vert^2)$, depending on a real smoothing parameter $\varepsilon_\diff > 0$. Let $[\PP]$ be the transition matrix in $\MM_{N_d}$ of a Markov chain such that $[\PP] = [\bb]^{-1}\, [K]$. For $1 < m \leq N_d$, let  $\bfg^1,\ldots ,\bfg^m$ be the right eigenvectors in $\RR^{N_d}$ of matrix $[\PP]$ such that $[\PP]\, \bfg^\alpha = \Lambda_\alpha\, \bfg^\alpha$, whose  eigenvalues are real and such that $1=\Lambda_1  > \Lambda_2 > \ldots > \Lambda_m$. The normalization condition of these eigenvectors is $[g]^T\, [\bb]\, [g] = [I_{m}]$, in which  $[g] = [\bfg^1 \ldots \bfg^m]\in \MM_{N_d,m}$ is the \textit{diffusion-maps basis}. The eigenvector $\bfg^1$ associated with the largest eigenvalue $\Lambda_1=1$ is a constant vector. For $m = N_d$, the diffusion-maps basis is an algebraic basis of $\RR^{N_d}$. The right-eigenvalue problem of the nonsymmetric matrix $[\PP]$ can be performed solving the eigenvalue problem $ [\bb]^{-1/2}\, [K]\, [\bb]^{-1/2}\, \bfPhi^\alpha = \Lambda_\alpha\, \bfPhi^\alpha$ related to a positive-definite symmetric real matrix, and  with the normalization $[\Phi]^T\,[\Phi] = [I_m]$, in which $[\Phi] = [\bfPhi^1 \ldots \bfPhi^m]$. Therefore, $\bfg^\alpha$ can be deduced from $\bfPhi^\alpha$ by $\bfg^\alpha =[\bb]^{-1/2}\, \bfPhi^\alpha$.
The construction introduces two hyperparameters: the dimension $m \leq N_d$ and the smoothing parameter $\varepsilon_\diff > 0$. An algorithm is proposed in \cite{Soize2019b} for estimating their values. Most of the time, $m$ and $\varepsilon_\diff$ can be chosen as follows.
Let $\varepsilon_\diff\mapsto \widehat m(\varepsilon_\diff)$ be the function from $\RR^{+*} = ]0\, , +\infty[$ into $\NN$ such that
\begin{equation}
\widehat m(\varepsilon_\diff) =
\arg \min_{\alpha \, \vert \, \alpha \geq 3}\left\{ \frac{\Lambda_{\alpha}(\varepsilon_\diff)}{\Lambda_{2}(\varepsilon_\diff)}
                                                                                                                      < 0.1\right\} \, .     \label{EQA1}
\end{equation}
If function $\widehat m$ is a decreasing function of $\varepsilon_\diff$ in the broad sense (if not, see \cite{Soize2019b}), then the optimal value  $\varepsilon_\diff^\opt$ of $\varepsilon_\diff$ can be chosen as the smallest value of the integer $\widehat m(\varepsilon_\diff^\opt)$ such that
\begin{equation}
\{\widehat m(\varepsilon_\diff^\opt) \! < \widehat m(\varepsilon_\diff)\, , \forall \varepsilon_\diff \in \, ]0 ,\varepsilon_\diff^\opt[ \,\}
\,\cap \, \{\widehat m(\varepsilon_\diff^\opt)  =
\widehat m(\varepsilon_\diff)\, , \forall \varepsilon_\diff \in \, ]\varepsilon_\diff^\opt , 1.5\,\varepsilon_\diff^\opt[ \, \} \, .        \label{EQA2}
\end{equation}
The corresponding optimal value $m^\opt$ of $m$ is then given by $m^\opt = \widehat m(\varepsilon_\diff^\opt)$.\\

\noindent \textit{A.3. Reduced-order representation of random matrices $[\bfH\,]$ and $[\bfX\,]$}.
The diffusion-maps vectors $\bfg^{1},\ldots , \bfg^{m}\in\RR^{N_d}$  span a subspace of $\RR^{N_d}$ that characterizes, for the optimal values $m^\opt$ and $\varepsilon_\diff^\opt$ of $m$ and $\varepsilon_\diff$, the local geometry structure of the  dataset $\{\bfeta^j, j=1,\ldots , N_d\}$.
The reduced-order representation is obtained by projecting each column of the $\MM_{N_d,\nu_x}$-valued random matrix $[\bfH]^T$ on the subspace
of $\RR^{N_d}$, spanned by $\{\bfg^{1}, \ldots, \bfg^m\}$. Let $[\bfZ]$ be the random matrix with values in $\MM_{\nu_x,m}$ such that
\begin{equation}
[\bfH] = [\bfZ]\, [g]^T \, .                                                                                             \label{EQA3}
\end{equation}
Since the eigenvector $\bfg^1$ is a constant vector and since random matrix $[\bfH]$ is centered, this eigenvector can be removed from the basis. As the matrix $[g]^T\, [g] \in \MM_m$ is invertible, the least-squares approximation of $[\bfZ]$ is written as
$[\bfZ] = [\bfH]\, [a]$ in which
\begin{equation}
[a]  = [g]\, ([g]^T\, [g])^{-1} \in \MM_{N_d,m}\, ,                                                   \nonumber
\end{equation}
and the realization $[z_d]\in \MM_{\nu_x,m}$ of  $[\bfZ]$ is written as
\begin{equation}
[z_d] = [\eta_d]\, [a]  \in \MM_{\nu_x,m}\, .                                                           \nonumber
\end{equation}

\noindent \textit{A.4. Generation of additional realizations $\{\bfeta^{\ell}_\ar, \ell=1,\ldots,\nu_\ar\}$ of random vector $\bfH$}.

An MCMC generator for random matrix $[\bfH]$ is constructed using the approach proposed in \cite{Soize2008b,Soize2015} belonging to the class of Hamiltonian Monte Carlo methods \cite{Soize2008b,Girolami2011,Neal2011}.
The generation of additional realizations $[z^1_\ar],\ldots ,[z^{n_\pMC}_\ar]$ of random matrix $[\bfZ]$ is carried out by using an
unusual MCMC method based on a reduced-order It\^o stochastic differential equation (ISDE)  that is constructed as the projection on the diffusion-maps basis of the ISDE related to a dissipative Hamiltonian dynamical system for which the invariant measure is the pdf of random matrix  $[\bfH]$ constructed with the Gaussian kernel-density estimation method and $[\eta_d]$. This method preserves the concentration of the probability measure and avoids the scatter phenomenon.
Let  $\{ ([\bfcurZ (t)],$ $[\bfcurY(t)]),$ $t\in \RR^+ \}$ be  the unique asymptotic (for $t\rightarrow +\infty$) stationary diffusion stochastic process with values in $\MM_{\nu_x,m}\times\MM_{\nu_x,m}$, of the following reduced-order ISDE (stochastic nonlinear second-order dissipative Hamiltonian dynamical system), for $t >0$,
\begin{equation}
 d[{\bfcurZ}(t)]  =  [{\bfcurY}(t)] \, dt \, , \nonumber
 \end{equation}
 \begin{equation}
  d[{\bfcurY}(t)] =  [\curL([{\bfcurZ(t)}])]\, dr -{1\over 2} f_0\,[{\bfcurY}(t)]\, dt + \sqrt{f_0}\, [d{\bfcurW^\wien}(t)] \, , \nonumber
\end{equation}
with the initial condition $[\bfcurZ(0)] = [z_d]$ and $[\bfcurY(0)] = [\bfcurN\,] \, [a]$ almost surely.\\

\noindent (i) The random matrix $[\curL([\bfcurZ(t)])]$ with values in $\MM_{\nu_x,m}$ is such that
$[\curL([\bfcurZ(t)])]= [L ( [\bfcurZ(t)] \, [g]^T ) ] \, [a]$.
For all $[u] = [\bfu^1 \ldots \bfu^{N_d}]$ in $\MM_{\nu_x,N_d}$ with $\bfu^j=(u^j_1,\ldots ,u^j_{\nu})$ in $\RR^{\nu_x}$, the
matrix $[L([u])]$ in $\MM_{\nu_x,N_d}$ is defined, for all $k = 1,\ldots ,\nu_x$ and for all $j=1,\ldots , N_d$, by
\begin{equation}
[L([u])]_{kj} = \frac{1}{p(\bfu^j)} \, \{{\boldsymbol{\nabla}}_{\!\!\bfu^j}\, p(\bfu^j) \}_k \, ,      \nonumber
\end{equation}
\begin{equation}
p(\bfu^j) = \frac{1}{N_d} \sum_{j'=1}^{N_d}
\exp\{ -\frac{1}{2 {\widehat s_{\nu_x}}^{\, 2}}\Vert \frac{\widehat s_{\nu_x}}{s_{\nu_x}}\bfeta^{j'}-\bfu^j\Vert^2 \}  \, ,           \nonumber
\end{equation}
\begin{equation}
{\boldsymbol{\nabla}}_{\!\!\bfu^j}\, p(\bfu^j) = \frac{1}{\widehat s_{\nu_x}^{\,2}}\frac{1}{N_d} \sum_{j'=1}^{N_d}
(\frac{\widehat s_{\nu_x}}{s_{\nu_x}}\bfeta^{j'}-\bfu^j)\, \exp\{ -\frac{1}{2 \widehat s_{\nu_x}^{\,2}}\Vert
             \frac{\widehat s_{\nu_x}}{s_{\nu_x}}\bfeta^{j'}-\bfu^j\Vert^2    \}  \, ,                                            \nonumber
\end{equation}
in which $\widehat s_{\nu_x}$  is the modified Silverman bandwidth  $s_{\nu_x}$, which has been introduced in \cite{Soize2015},
\begin{equation}
\widehat s_{\nu_x} =   \frac{s_{\nu_x}}{\sqrt{s_{\nu_x}^2 +\frac{N_d -1}{N_d}}} \quad , \quad
s_{\nu_x} = \left\{\frac{4}{N_d(2+\nu_x)} \right\}^{1/(\nu_x +4)}   \, .                                                          \nonumber
\end{equation}

\noindent (ii) $[\bfcurW^\wien(t)] = [\WW^\wien(t)] \, [a]$ where $\{[\WW^\wien(t)], t\in \RR^+\}$ is the $\MM_{\nu_x,N_d}$-valued normalized Wiener stochastic process.

\noindent (iii)  $[\bfcurN\,]$ is the $\MM_{\nu_x,N_d}$-valued normalized Gaussian random matrix that is independent of stochastic process $[\WW^\wien]$.

\noindent (iv)  The free parameter $f_0$, such that $0 < f_0 < 4$, allows the dissipation term of the nonlinear second-order dynamical system (dissipative
Hamiltonian system)  to be controlled in order to kill the transient part induced by the initial conditions. A common value is $f_0=1.5$.

\noindent (v)  We then have ${[\bfZ]} = \lim_{t\rightarrow +\infty} {[\bfcurZ(t)]}$ in probability distribution,
which allows for generating the additional realizations, $[z_\ar^1], \ldots , [z_\ar^{n_\pMC}]$, and then, generating the additional realizations
$[\eta_\ar^1],\ldots ,[\eta_\ar^{n_\pMC}]$ by using Eq.~\eqref{EQA3}, such that $[\eta_\ar^\ell] = [z_\ar^\ell]\, [g]^T$ (see Section~A.6).\\

\noindent \textit{A.5. Algorithm for solving the reduced-order ISDE}.
Let $M = n_\MC \times M_0$ be the positive integer in which $n_\MC$ and $M_0$ are integers.
The reduced-order ISDE is solved on the finite interval $\curR = [0\, , M\, \Delta t]$, in which $\Delta t$ is the sampling step
of the continuous index parameter $t$. The integration scheme is based on the use of the $M+1$ sampling points
$t_{\ell'}$ such that $t_{\ell'} = \ell'\,  \Delta t$ for $\ell'= 0,\ldots , M$ for which
$[\bfcurZ_{\ell'}] = [\bfcurZ(t_{\ell'})]$, $[\bfcurY_{\ell'}] = [\bfcurY(t_{\ell'})]$, and $[\bfcurW^\wien_{\ell'}] = [\bfcurW^\wien(t_{\ell'})]$,
 with $[\bfcurZ_0] = [z_d]$, $[\bfcurY_0] = [\bfcurN\,] \, [a]$, and $[\bfcurW^\wien_0] = [0_{\nu_x,m}]$.
For $\ell'=0,\ldots , M-1$, let $[\Delta\bfcurW^\wien_{\ell'+1}] = [\Delta\WW^\wien_{\ell'+1}]\, [a]$
be the sequence of random matrices with values in $\MM_{\nu_x,m}$, in which the
 increments $[\Delta\WW^\wien_{1}], \ldots , [\Delta\WW^\wien_{M}]$ are $M$ independent random matrices with values in $\MM_{\nu_x,N_d}$.
For all $k=1,\ldots , \nu_x$ and for all $j=1,\ldots , N_d$, the real-valued random variables $\{[\Delta\WW^\wien_{\ell'+1}]_{kj}\}_{kj}$ are independent, Gaussian, second-order, and centered random variables such that
\begin{equation}
E\{[\Delta\WW^\wien_{\ell'+1}]_{kj}[\Delta\WW^\wien_{\ell'+1}]_{k'j'}\}=\Delta t \,\delta_{kk'}\,\delta_{jj'} \, . \nonumber
\end{equation}
For $\ell'=0,\ldots, M-1$, the St\"{o}rmer-Verlet scheme applied to the reduced-order ISDE yields
\begin{equation}
[\bfcurZ_{\ell'+\frac{1}{2}}]     =    [\bfcurZ_{\ell'}] +\frac{\Delta t}{2} \, [\bfcurY_{\ell'}] \, ,    \nonumber
\end{equation}
\begin{equation}
[\bfcurY_{\ell'+1}]   =    \frac{1-b}{1+b}\, [\bfcurY_{\ell'}] + \frac{\Delta t}{1+b}\, [\bfcurL_{\ell'+\frac{1}{2}}] +
       \frac{\sqrt{f_0}}{1+b}\, [\Delta\bfcurW^\wien_{\ell'+1}]\, ,                                                                         \nonumber
\end{equation}
\begin{equation}
[\bfcurZ_{\ell'+ 1}]  =    [\bfcurZ_{\ell'+\frac{1}{2}}] +\frac{\Delta t}{2} \, [\bfcurY_{\ell'+1}]\, ,                                \nonumber
\end{equation}
with the initial condition defined before, where $b=f_0\, \Delta t\, / 4$, and where $[\bfcurL_{\ell'+\frac{1}{2}}]$ is the $\MM_{\nu_x,m}$-valued random variable such that
\begin{equation}
 [\bfcurL_{ \ell'+\frac{1}{2} }]  = [\curL( [\bfcurZ_{\ell'+\frac{1}{2}}] )] = [L( [\bfcurZ_{\ell'+\frac{1}{2}}]\, [g]^T )]\, [a]\, .    \nonumber
 \end{equation}

\noindent \textit{A.6. Additional realizations $\{\bfx^{\ell}_\ar, \ell=1,\ldots,\nu_\ar\}$ of random vector $\bfX$}.
The reduced-order ISDE is then used for generating $n_\MC$ additional realizations, $[z_\ar^1], \ldots , [z_\ar^{n_\pMC}]$ in $\MM_{\nu_x,m}$, of random matrix $[\bfZ]$, and therefore, for deducing the $n_\MC$ additional realizations,
$[\eta_\ar^1], \ldots , [\eta_\ar^{n_\pMC}]$ in  $\MM_{\nu_x,N_d}$ of random matrix $[\bfH]$, such that $[\eta_\ar^\ell] = [z_\ar^\ell]\, [g]^T$ for $\ell=1,\ldots , n_\MC$. The computation is performed as follows.
Let $\nu_\ar = n_\MC \times N_d$, in which $n_\MC$ is a any given integer.
Let $[\WW^\wien(\cdot;\theta)]$ with $\theta\in\Theta$ be a realization of the Wiener stochastic process $[\WW^\wien]$ defined in Section~A.4-(ii). Let $\{([\bfcurZ(t;\theta)],[\bfcurY(t;\theta)]),t\in\RR^+\}$ be one realization of the $(\MM_{\nu_x,m} \!\times\! \MM_{\nu_x,m})$-valued stochastic process $\{([\bfcurZ(t)],[\bfcurY(t)]),t\in\RR^+\}$, for which its time-sampling is computed using the algorithm presented in Section~A.5.
Let $\ell_0$ be the integer such that, for $t \geq \ell_0\, \Delta t$, the solution is asymptotic to the stationary solution. Therefore, the independent realizations $\{\bfeta^\ell,\ell=1,\ldots, \nu_\ar\}$ of $\bfH$ are generated as follows. Let $M_0$ be a given positive integer.  For $\kappa =1,\ldots , n_\MC$ and for $t_{\ell'} =\ell'\,\Delta t$ with $\ell' = \ell_0 + \kappa\, M_0$, we have, for $j=1,\ldots, N_d$ and for $k=1,\ldots, \nu_x$,
$\eta_k^\ell  = \{[\bfcurZ(t_{\ell'},\theta)]\,[g]^T\}_{kj}$ with $\ell = j +(\ell'-1)\, N_d$.
In this method of generation, only one realization $\theta$ is used and $M_0$ is chosen sufficiently large in order that
$[\bfcurZ(t_{\ell'})]$ and $[\bfcurZ(t_{(\ell'+M_0)})]$ be two random matrices that are approximately independent.
The realizations $\{\bfx_\ar^\ell,\ell=1,\ldots,\nu_\ar \}$ of random vector $\bfX$ are then calculated by
$\bfx_\ar^\ell = \underline \bfx + [\varphi]\, [\lambda]^{1/2}\, \bfeta^\ell$ with $\bfeta^\ell = (\eta_1^\ell,\ldots,\eta_\nu^\ell)$.
%
%
\section{Proof of the convergence of the random sequence  $\bfX^{(\nu_q,\nu_w)}$.}
\label{AppendixB}
Since $\bfX=(\bfQ,\bfW)$, $\underline\bfx_\ar=(\underline\bfq_\ar , \underline\bfw_\ar)$, and $\bfX^{(\nu_q,\nu_w)}= (\bfQ^{(\nu_q)},\bfW^{(\nu_w)})$, we have
\begin{equation}
\nonumber
E\{\Vert\bfX - \underline\bfx_\ar\Vert^2 \}= E\{\Vert\bfQ - \underline\bfq_\ar\Vert^2\}  + E\{\Vert\bfW - \underline\bfw_\ar\Vert^2\}\, ,
\end{equation}
that is equal to $\tr[C_\bfQ] + \tr[C_\bfW]$, in which $\bfX$, $\bfQ$, and $\bfW$ stand for $\bfX^{(n_q,n_w)}$, $\bfQ^{(n_q)}$, and $\bfW^{(n_w)}$.
We also have
\begin{equation}
E\{\Vert\bfX - \bfX^{(\nu_q,\nu_w)}\Vert^2 \}= E\{\Vert\bfQ - \bfQ^{(\nu_q)}\Vert^2\}  + E\{\Vert\bfW - \bfW^{(\nu_w)}\Vert^2\}\, ,
\nonumber
\end{equation}
that can be rewritten, using Eqs.~\eqref{EQ9} and \eqref{EQ15}, as
\begin{equation}
\nonumber
E\{\Vert\bfX - \bfX^{(\nu_q,\nu_w)}\Vert^2 \} =\err_{\bfQ}(\nu_q)\, E\{\Vert\bfQ - \underline\bfq_\ar\Vert^2\}   +\err_{\bfW}(\nu_w)\, E\{\Vert\bfW - \underline\bfw_\ar\Vert^2\} \, .
\end{equation}
Since $\tr[C_\bfQ] > 0$ and $\tr[C_\bfW] > 0$, it can then be deduced that
\begin{equation}
\nonumber
\err_{\bfX}(\nu_q,\nu_w) =\err_{\bfQ}(\nu_q)\,\frac{1}{1 +{\tr[C_\bfW] }/{\tr[C_\bfQ]}} +
                           \err_{\bfW}(\nu_w)\,\frac{1}{1 +{\tr[C_\bfQ] }/{\tr[C_\bfW]}}\, .
\end{equation}
Defining $\zeta = \max\{ (1 +{\tr[C_\bfW]}/{\tr[C_\bfQ]})^{-1}, (1 +{\tr[C_\bfQ]}/{\tr[C_\bfW]})^{-1}\} > 0$ yields
$\err_{\bfX}(\nu_q,\nu_w) \leq \zeta\, (\err_{\bfQ}(\nu_q) +\err_{\bfW}(\nu_w))$. Since $\zeta < 1$, we then obtain
\begin{equation}
\err_{\bfX}(\nu_q,\nu_w) \leq \err_{\bfQ}(\nu_q) +\err_{\bfW}(\nu_w)\, .
\end{equation}

%
%
\section{Proof of the range of the values of covariance matrix $[C_{\widehat X}]$ defined by Eq.~\eqref{EQ26}}
\label{AppendixC}

Since matrix $[C_{\widehat X}]$ is positive or positive definite, we have $< \!\! [C_{\widehat X}]\, \widehat\bfx ,\widehat\bfx \! > \, \geq 0$  for all $\widehat\bfx = (\widehat\bfq,\widehat\bfw)$ in $\RR^\nu = \RR^{\nu_q}\times\RR^{\nu_w}$. Using Eq.~\eqref{EQ26} yields
\begin{equation}
2 < \! \![C_{qw}]^T\, \widehat\bfq \, , \widehat\bfw  \! > + \Vert\widehat\bfq\Vert^2 + \Vert\widehat\bfw\Vert^2 \, \geq 0 \, .                  \label{EQC1}
\end{equation}
For all $\widehat\bfq$ in $\RR^{\nu_q}$, we can choose $\widehat\bfw = - [C_{qw}]^T\, \widehat\bfq$ in Eq.~\eqref{EQC1}, which yields
\begin{equation}
\Vert[C_{qw}]^T\, \widehat\bfq\Vert^2 \leq \Vert\widehat\bfq\Vert^2 \, ,                                                    \nonumber
\end{equation}
which can be rewritten as
\begin{equation}
< \!\! [C_{qw}]\,[C_{qw}]^T\, \widehat\bfq \, , \widehat\bfq \! > \,\,\, \leq \,\,\Vert\widehat\bfq\Vert^2 \quad ,
                                                 \quad \forall\, \widehat\bfq \in \RR^{\nu_q}\, .                                            \label{EQC2}
\end{equation}
Let $\Lambda_1\geq \ldots \geq \Lambda_{\nu_q}\geq 0$ be the eigenvalues  of the positive matrix $[C_{qw}]\,[C_{qw}]^T$. Equation~\eqref{EQC2} shows that
\begin{equation}
1 \geq \Lambda_1\geq \ldots \geq \Lambda_{\nu_q}\geq 0\, .                                                                                \label{EQC3}
\end{equation}
Let us consider the eigenvalue problem
$[C_{\widehat\bfX}]\, \widehat\varphi = \lambda\, \widehat\varphi$ (see Eq.~\eqref{EQ30} in which the columns of matrix $[\Phi]$ are the eigenvectors $\widehat\varphi$). Using the block decomposition defined by Eq.~\eqref{EQ26} and $\widehat\varphi = (\widehat\varphi_q,\widehat\varphi_w)$ yield
\begin{equation}
\widehat\varphi_q + [C_{qw}]\,\widehat\varphi_w  = \lambda\, \widehat\varphi_q \, ,                                            \label{EQC4}
\end{equation}
\begin{equation}
[C_{qw}]^T\,\widehat\varphi_q  + \widehat\varphi_w  = \lambda\, \widehat\varphi_w \, .                                            \label{EQC5}
\end{equation}
Eliminating $\widehat\varphi_w$ between Eqs.~\eqref{EQC4} and \eqref{EQC5} yields
\begin{equation}
[C_{qw}]\,[C_{qw}]^T\, \widehat\varphi_q = (1-\lambda)^2\, \widehat\varphi_q \, .                                            \nonumber
\end{equation}
Consequently, $(1-\lambda)^2$ appears as the eigenvalue $\Lambda$ of matrix $[C_{qw}]\,[C_{qw}]^T$. Taking into account Eq.~\eqref{EQC3}, it can be deduced that $-1\leq  1-\lambda \leq 1$, which proves that any eigenvalue $\lambda$ of matrix $[C_{\widehat X}]$ is such that $0\leq \lambda \leq 2$.
%
%
\section{Proof of Eq.~\eqref{EQ35.3} for the consistency of the estimator defined by Eq.~\eqref{EQ35.2} corresponding to the estimation defined by Eq.~\eqref{EQ35}}
\label{AppendixD}
The proof is inspired of \cite{Parzen1962}, is slightly different, is adapted to the Gaussian kernel-density, and the upper bound defined by Eq.~\eqref{EQ35.3} is not the same. The Silverman bandwidth $s_\ar$ is defined by Eq.~\eqref{EQ29} and $\widehat\bfx$ is a point fixed in $\RR^\nu$.
Let $\widehat\bfx\mapsto\kappa(\widehat\bfx)$ be the Gaussian pdf, centered, with invertible covariance matrix $[\widehat C_\varepsilon]$ defined by Eq.~\eqref{EQ30quarter}, such that $[G] = [\widehat C_\varepsilon]^{-1} \in \MM_\nu^+$, and let
$\widehat\bfx\mapsto\kappa_{\nu_\arp}(\widehat\bfx)$ be the function on $\RR^\nu$, such that,
\begin{equation}
\kappa(\widehat\bfx) = \frac{\sqrt{\det [G]}}{(2\pi)^{\nu/2}} \, \exp\{-\frac{1}{2} < \! \! [G]\,\widehat\bfx \, , \widehat\bfx\! >\}  \, ,       \label{EQD1}
\end{equation}
\begin{equation}
\kappa_{\nu_\arp}(\widehat\bfx) = \frac{1}{s_\arp^\nu} \,\kappa(\frac{\widehat\bfx}{s_\arp})\, .                    \label{EQD2}
\end{equation}
Using the change of variable $\widehat\bfx = [\Phi] \, \widehat\bfeta$ with $[G] = [\Phi]\,[\Lambda_\varepsilon]^{-1}\, [\Phi]^T $
(see Eq.~\eqref{EQ31bis}) and since $s_\ar\rightarrow 0$ when $\nu_\ar\rightarrow +\infty$, it can be seen that we have the following limit in the space of measures on $\RR^\nu$,
\begin{equation}
\lim_{\nu_\arpp \rightarrow +\infty} \kappa_{\nu_\arp}(\widehat\bfx) \, d\widehat\bfx  = \delta_0(\widehat\bfx)\, ,                    \label{EQD3}
\end{equation}
in which $d\widehat\bfx$ is the Lebesgue measure on $\RR^\nu$ and $\delta_0(\widehat\bfx)$ is the Dirac measure on $\RR^\nu$ at point $\widehat\bfx = \bfzero$.\\

\noindent \textit{(i) Sequence of estimators of} $p_{\widehat\bfX}$. Let $\widehat\bfX^1, \ldots , \widehat\bfX^{\nu_\arp}$ be $\nu_\ar$ independent copies of random variable $\widehat\bfX$ whose pdf is $p_{\widehat\bfX}$. Therefore, $\widehat\bfx^\ell$ is a realization of $\widehat\bfX^\ell$.
For $\widehat\bfx$ fixed in $\RR^\nu$, the sequence of estimators of $p_{\widehat\bfX}(\widehat\bfx)$, whose an estimation is
$p_{\widehat\bfX}^{(\nu_\arpp)}(\widehat\bfx)$ defined by Eq.~\eqref{EQ35}, is the sequence $\{P_{\nu_\arp}(\widehat\bfx)\}_{\nu_\arp}$ of positive-valued random variables defined by
\begin{equation}
P_{\nu_\arp}(\widehat\bfx) = \frac{1}{\nu_\arp} \sum_{\ell=1}^{\nu_\arp} \kappa_{\nu_\arp}(\widehat\bfX^\ell -\widehat\bfx)\, .         \label{EQD4}
\end{equation}
\\

\noindent \textit{(ii) Mean value} $\underline{P}_{\nu_\arp}(\widehat\bfx)$ \textit{of} $P_{\nu_\arp}(\widehat\bfx)$. The mean value of random variable
$P_{\nu_\arp}(\widehat\bfx)$ is written as $E\{P_{\nu_\arp}(\widehat\bfx)\} = \frac{1}{\nu_\arp} \sum_{\ell=1}^{\nu_\arp} E\{\kappa_{\nu_\arp}(\widehat\bfX^\ell -\widehat\bfx)\}$, which yields
\begin{equation}
\underline{P}_{\nu_\arp}(\widehat\bfx) = \int_{\RR^\nu} \kappa_{\nu_\arp}(\widetilde\bfx -\widehat\bfx)\,
                   p_{\widehat\bfX}(\widetilde\bfx)\,d\widetilde\bfx \, .                                                               \label{EQD5}
\end{equation}
Assuming that $p_{\widehat\bfX}$ is a continuous function in $\widehat\bfx\in\RR^\nu$, using Eq.~\eqref{EQD3} yields
\begin{equation}
\lim_{\nu_\arpp \rightarrow +\infty} \underline{P}_{\nu_\arp}(\widehat\bfx) = p_{\widehat\bfX}(\widehat\bfx)\, .                           \label{EQD6}
\end{equation}

\noindent \textit{(iii) Variance of} $P_{\nu_\arp}(\widehat\bfx)$. Since the random variables
$\widehat\bfX^1, \ldots , \widehat\bfX^{\nu_\arp}$ are independent copies of $\widehat\bfX$, and using Eq.~\eqref{EQD2}, the variance of
$P_{\nu_\arp}(\widehat\bfx)$ is such that
\begin{align}
E\{ (P_{\nu_\arp}(\widehat\bfx) - \underline{P}_{\nu_\arp}(\widehat\bfx))^2\} & =
          \frac{1}{\nu_\ar} E\{(\kappa_{\nu_\arp}(\widehat\bfX -\widehat\bfx))^2\} -\frac{1}{\nu_\ar}(\underline{P}_{\nu_\arp}(\widehat\bfx))^2 \nonumber \\
                                                  &  \leq   \frac{1}{\nu_\ar} E\{(\kappa_{\nu_\arp}(\widehat\bfX -\widehat\bfx))^2\} \nonumber \\
                                                  & =  \frac{1}{\nu_\ar} \int_{\RR^\nu} (\kappa_{\nu_\arp}(\widetilde\bfx -\widehat\bfx))^2\,
                                                        p_{\widehat\bfX}(\widetilde\bfx)\,d\widetilde\bfx \, . \nonumber
\end{align}
Since $\forall\,\widehat\bfx$, $\, \sup_{\widetilde\bfx} \, \kappa_{\nu_\arpp}(\widetilde\bfx -\widehat\bfx) = \frac{1}{s_\arpp^\nu}
\frac{\sqrt{\det [G]}}{(2\pi)^{\nu/2}}$ and using Eqs.~\eqref{EQD2} and \eqref{EQD5}, we have
\begin{equation}
E\{ (P_{\nu_\arp}(\widehat\bfx) - \underline{P}_{\nu_\arp}(\widehat\bfx))^2\} \leq
\frac{1}{\nu_\arpp s_\arpp^\nu}
\frac{\sqrt{\det [G]}}{(2\pi)^{\nu/2}}\,\underline{P}_{\nu_\arp}(\widehat\bfx) \, .                                                 \label{EQD7}
\end{equation}
Substituting $s_\ar$ given by Eq.~\eqref{EQ29} into the right-hand side of Eq.~\eqref{EQD7} yields
\begin{equation}
E\{ (P_{\nu_\arp}(\widehat\bfx) - \underline{P}_{\nu_\arp}(\widehat\bfx))^2\}  \leq
\left \{ \! \frac{1}{\nu_\ar} \!\right \}^{\! 4/(\nu+4)}\!
\left \{\! \frac{\nu \! + \! 2}{4}\! \right \}^{\! \nu/(\nu+4)} \,
\!\!\frac{\sqrt{\det [G]}}{(2\pi)^{\nu/2}} \,\underline{P}_{\nu_\arp}(\widehat\bfx)  \, .                                           \label{EQD8}
\end{equation}

\noindent \textit{(iv) Properties of the sequence of estimators}. It can be seen that
\begin{equation}
E\{ (P_{\nu_\arp}(\widehat\bfx) - p_{\widehat\bfX}(\widehat\bfx))^2\}  =
E\{ (P_{\nu_\arp}(\widehat\bfx) - \underline{P}_{\nu_\arp}(\widehat\bfx))^2\}  +
 ( \underline{P}_{\nu_\arp}(\widehat\bfx) - p_{\widehat\bfX}(\widehat\bfx) )^2 \, .                                           \label{EQD9}
\end{equation}
Using Eqs.~\eqref{EQD6}, \eqref{EQD8}, and \eqref{EQD9} for $\nu_\ar \rightarrow +\infty$, it can be seen that the estimator
$P_{\nu_\arp}(\widehat\bfx)$ is asymptotically unbiased  and is consistent because
\begin{equation}
\lim_{\nu_\arpp \rightarrow +\infty} E\{ ( P_{\nu_\arp}(\widehat\bfx) - p_{\widehat\bfX}(\widehat\bfx))^2\} = 0 \, .                   \label{EQD10}
\end{equation}
The mean-square convergence corresponding to Eq.~\eqref{EQD10} implies the convergence in probability.
%
%
\section{Construction of the diffusion-maps basis for the posterior model}
\label{AppendixE}
The construction, based on \cite{Coifman2005}, is the one presented in \cite{Soize2016} and is summarized in Appendix ~A.2, using the $N_s$ independent realizations $\{\bfs^j,j=1,\ldots,N_s\}$ defined by Eq.~\eqref{EQ90}. Let $[\PP_s] = [\bb_s]^{-1}\, [K_s]$ be the matrix in $\MM_{N_s}$ such that,
for all $i$, $j$ and $j'$ in $\{1,\ldots , N_s\}$, $[K_s]_{jj'} = \exp(-\frac{1}{4\,\varepsilon^\ppost_\pdiff} \Vert\bfs^{j} -\bfs^{j'}\Vert^2)$ and
$[\bb_s]_{ij} = \delta_{ij}\,\sum_{j'=1}^{N_s} [K_s]_{jj'}$ depending on a positive parameter $\varepsilon_\diff^\post$ whose value depends on dataset $\{\bfs^j,j=1,\ldots,N_s\}$. Therefore, $[\PP_s]$ is a transition matrix of a Markov chain. For $1 < m_\post \leq N_s$, let  $\bfg_s^1,\ldots ,\bfg_s^{m_\ppost}$ be the right eigenvectors in $\RR^{N_s}$,  of the eigenvalue problem $[\PP_s]\, \bfg_s^\alpha = \Lambda_{s,\alpha}\, \bfg_s^\alpha$ with the normalization condition $[g_s]^T\, [\bb_s]\, [g_s] = [I_{m_\ppost}]$ and where the associated $m_\post \leq N_s$ positive eigenvalues are such that
$1=\Lambda_{s,1}  > \Lambda_{s,2}  > \ldots > \Lambda_{s,m_\ppost}$. The diffusion-maps basis is represented by the
matrix $[g_s] = [\bfg_s^1 \ldots \bfg^{m_\ppost}_s]\in \MM_{N_s,m_\ppost}$. The eigenvector $\bfg_s^1$ associated with the largest eigenvalue $\Lambda_{s,1}=1$ is a constant vector that has to be kept because stochastic process $[\bfS]$ is not centered (it is $[\bfS^\linear]$ that is a centered stochastic process). For $m_\post = N_s$, the diffusion-maps basis is an algebraic basis of $\RR^{N_s}$. The right-eigenvalue problem of the nonsymmetric matrix $[\PP_s]$ can be performed solving the eigenvalue problem $ [\bb_s]^{-1/2}\, [K_s]\, [\bb_s]^{-1/2}\, \Phi_s^\alpha = \Lambda_{s,\alpha}\, \Phi_s^\alpha$ related to a positive-definite symmetric real matrix with the normalization $\Vert\Phi_s^\alpha\Vert = 1$. Therefore, $\bfg_s^\alpha$ can be deduced from $\Phi_s^\alpha$ by $\bfg_s^\alpha =[\bb_s]^{-1/2}\, \Phi_s^\alpha$.
The construction introduces two hyperparameters: the dimension $m_\post \leq N_s$ and the smoothing parameter $\varepsilon_\diff^\post > 0$. The algorithm
for estimating the optimal values of $\varepsilon_\diff^\post$ and $m_\post$ is detailed in \cite{Soize2019b}. Most of the time, these optimal values can be calculated using Eqs.~\eqref{EQA1} and \eqref{EQA2} in which $\varepsilon_\diff$ has to be replaced by $\varepsilon_\diff^\post$.
%
%
\section{St\"{o}rmer-Verlet scheme for solving the reduced-order ISDE}
\label{AppendixE}
Let $n_\MC^\post$  and $M_0^\post$ be the given integers defined in Section~\ref{Section8.5}. The reduced-order ISDE defined by Eqs.~\eqref{EQ92} to \eqref{EQ94} is solved for $t\in [0,t_\pmax]$ with $t_\pmax =(\ell_0+n_\MC^\post\, M_0^\post)\, \Delta t$ in which $\Delta t$ is the sampling step and where $\ell_0$ is chosen in order that the solution of Eqs.~\eqref{EQ92} to \eqref{EQ94} has reached the stationary regime. For $\ell=0,1,\ldots, n_\MC^\post\, M_0^\post$, we consider the sampling points $t_\ell = \ell\, \Delta t$ and the following notations:
$[\bfcurZ_\ell] = [\bfcurZ(t_\ell)]$, $[\bfcurY_\ell] = [\bfcurY(t_\ell)]$, and $[\bfcurW^\wien_\ell] = [\bfcurW^\wien(t_\ell)]$.
The St\"{o}rmer-Verlet scheme is used for solving the reduced-order ISDE, which is written, for $\ell=0,1,\ldots, n_\MC^\post\, M_0^\post$, as
\begin{equation}
[\bfcurZ_{\ell+\frac{1}{2}}]     =    [\bfcurZ_{\ell}] +\frac{\Delta t}{2} \, [\bfcurY_{\ell}] \, ,    \nonumber
\end{equation}
\begin{equation}
[\bfcurY_{\ell+1}]   =    \frac{1-\beta}{1+\beta}\, [\bfcurY_{\ell}] + \frac{\Delta t}{1+\beta}\, [\widetilde\bfcurL_{\ell+\frac{1}{2}}] +
       \frac{\sqrt{f_0^\post}}{1+\beta}\, [\Delta\bfcurW^\wien_{\ell+1}]\, ,                                                          \nonumber
\end{equation}
\begin{equation}
[\bfcurZ_{\ell+ 1}]  =    [\bfcurZ_{\ell+\frac{1}{2}}] +\frac{\Delta t}{2} \, [\bfcurY_{\ell+1}]\, ,                                \nonumber
\end{equation}
with the initial condition defined by Eq.~\eqref{EQ94}, where
$\beta=f_0^\post\, \Delta t\, / 4$, and where $[\widetilde\bfcurL_{\ell+\frac{1}{2}}]$ is the $\MM_{\nu_w,m_\ppost}$-valued random variable such that
\begin{equation}
 [\widetilde\bfcurL_{ \ell+\frac{1}{2} }]  = [\widetilde\curL( [\bfcurZ_{\ell+\frac{1}{2}}] )]
 = [\widetilde L( [\bfcurZ_{\ell+\frac{1}{2}}]\, [g_s]^T )]\, [a_s]\, .                                                                \nonumber
\end{equation}
In the above equation,
$[\Delta\bfcurW^\wien_{\ell+1}] = [\Delta\bfW^\wien_{\ell+1}]\, [a_s]$
is a random variable with values  in $\MM_{\nu_w,m_\ppost}$, in which the
 increment $[\Delta\bfW^\wien_{\ell+1}] = [\bfW^\wien_{\ell+1}] - [\bfW^\wien_{\ell}]$. The increments are statistically independent.
For all $k=1,\ldots , \nu_w$ and for all $j=1,\ldots , N_s$, the real-valued random variables $\{[\Delta\bfW^\wien_{\ell+1}]_{kj}\}_{kj}$ are independent, Gaussian, second-order, and centered random variables such that
\begin{equation}
E\{[\Delta\bfW^\wien_{\ell+1}]_{kj}[\Delta\bfW^\wien_{\ell+1}]_{k'j'}\}=\Delta t \,\delta_{kk'}\,\delta_{jj'} \, . \nonumber
\end{equation}
%



\bibliographystyle{elsarticle-num}
\bibliography{references}

\end{document}